\documentclass{article}

\usepackage{microtype}
\usepackage{graphicx}
\usepackage{subfigure}
\usepackage{booktabs} 
\usepackage{array}
\usepackage{makecell}
\usepackage{xcolor}

\usepackage{hyperref}

\usepackage{amsmath}
\usepackage{amssymb}
\usepackage{amsthm}
\usepackage{graphicx}
\newtheorem{assumption}{Assumption}

\theoremstyle{plain}
\newtheorem{theorem}{Theorem}[section]
\newtheorem{lemma}[theorem]{Lemma}
\newtheorem{corollary}[theorem]{Corollary}
\theoremstyle{remark}
\newtheorem{remark}[theorem]{Remark}

\usepackage[accepted]{mlsys2025}

\mlsystitlerunning{Multi-Turn Semantic Caching for Low-Cost LLM Serving}

\begin{document}

\twocolumn[
\mlsystitle{Multi-Turn Semantic Caching for Low-Cost LLM Serving}



\mlsyssetsymbol{equal}{*}

\begin{mlsysauthorlist}
\mlsysauthor{Baran Atalar}{cmu}
\mlsysauthor{Selly Beyene}{uwt}
\mlsysauthor{Xutong Liu}{uwt}
\mlsysauthor{Carlee Joe-Wong}{cmu}
\end{mlsysauthorlist}

\mlsysaffiliation{cmu}{Carnegie Mellon University, Pittsburgh, PA, USA}
\mlsysaffiliation{uwt}{University of Washington Tacoma, Tacoma, WA, USA}

\mlsyscorrespondingauthor{Baran Atalar}{batalar@andrew.cmu.edu}




\mlsyskeywords{Machine Learning, MLSys}

\vskip 0.3in

\begin{abstract}
Semantic caching lowers the cost of serving large language models (LLMs) by reusing stored responses for similar requests. In multi-turn conversations, whether a cached response is acceptable depends on both conversations' histories, which embedding similarity of requests and histories predicts poorly (AUC 0.62--0.74 on five workloads). We therefore learn the \emph{reuse risk}, the probability that a cached response is unacceptable, from judge feedback, and choose a budgeted set of conversation states to cache by weighing the generation cost a reuse avoids against a price for unacceptable reuse. We propose CUCB-SC-MT, which learns a cache from logged conversations with a suboptimality guarantee, and CLCB-LS-MT, which learns online and has $\tilde O(\sqrt N)$ regret over $N$ requests when conversation states recur. On math-tutoring (SocraTeach) and movie-ticketing (Taskmaster-3) conversations, both achieve lower loss than prior semantic caching approaches: ContextCache, multi-turn variants of ElastiCache and vCache, and single-turn semantic caching at every cache budget. With 1,000 cache entries, CLCB-LS-MT avoids 35--41\% of the generation cost through acceptable reuses, against at most 22--25\% for the baselines, and lowers estimated serving latency by 19--35\%.
\end{abstract}
]



\printAffiliationsAndNotice{}  
\section{Introduction}\label{sec:intro}

Serving large language models (LLMs) is expensive, and much of the work repeats: about 31\% of ChatGPT users' queries are similar to earlier ones~\cite{gill2025meancache}. \emph{Semantic caching} answers a new request with the stored response of a similar earlier one and skips token generation entirely~\cite{bang2023gptcache,gaspar2026vcache}, thus lowering serving cost and latency. Recent
work gives such caches cost guarantees, both offline and online~\cite{liu2026semanticcachinglowcostllm,atalar2026continuoussemanticcachinglowcost}. Cloud providers also offer semantic caching as a service \cite{aws_elasticache_doc}.
 
\begin{figure}[t]
  \centering
  \includegraphics[width=\columnwidth]{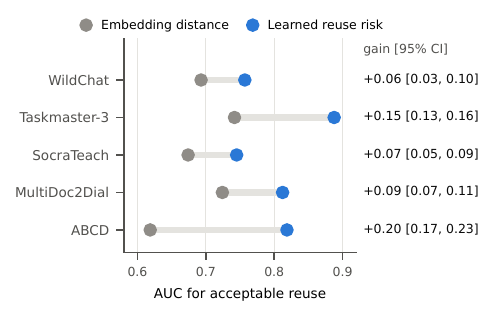}
  \caption{AUC for acceptable reuse of embedding distance and of a learned reuse risk when each follow-up turn is served by its nearest state (request and history) in another conversation and an LLM judge rates the reuse. Right: AUC gain of the learned risk (95\% CI). The risk model is fixed on WildChat (Appendix~\ref{app:screening}).
  }
  \label{fig:motivation}
\end{figure}

These guarantees treat each request as a standalone query, but LLM traffic is conversational: 73\% of ShareGPT conversations have several turns~\cite{gao2024cachedattention}. The same follow-up, such as ``Yes, that's right'' in a ticket booking, calls for different replies under different histories, so whether a cached response is acceptable depends on a \emph{pair} of conversation states, the current one and the one whose response is cached, and the probability that it is not, the \emph{reuse risk}, is unknown. Yet multi-turn semantic caches still threshold an embedding similarity, computed on the recent messages or on the request followed by a context check~\cite{gill2025meancache,contextcache,yu2025smartcache}, and single-turn analyses take the cost of a mismatch to be a known function of embedding distance~\cite{atalar2026continuoussemanticcachinglowcost}.

Figure~\ref{fig:motivation} shows that distance is a poor proxy in conversations. We took five workloads: open-domain chat (WildChat~\cite{zhao2024wildchat1mchatgptinteraction}), movie-ticket booking
(Taskmaster-3~\cite{byrne2021tickettalk}), math tutoring (SocraTeach~\cite{socrateach}), document-grounded assistance (MultiDoc2Dial~\cite{feng2021multidoc2dial}) and customer service (ABCD~\cite{chen2021abcd}). We matched every follow-up turn to its nearest state (request and history) in another conversation and asked an LLM judge whether the cached response would be acceptable. Distance ranks acceptable reuse poorly (AUC 0.62--0.74). A reuse-risk model learned from judge labels with kernel regression on features that compare the two requests, their histories, the cached response and the turn positions, ranks it better on every workload (AUC $+0.06$ to $+0.20$), with one configuration fixed on open-domain WildChat. At 80\% judge precision it serves 2.2$\times$ as many Taskmaster-3 follow-ups as a distance threshold (38.9\% vs.\ 17.3\%).

Learning this risk while serving is hard: it lives on pairs of states whose histories grow with every turn and is revealed only by judge labels on reuses, while generation costs are revealed only by LLM calls; turns of a conversation are correlated; and choosing what to cache is NP-hard. Our contributions are:
\begin{itemize}
\item \textbf{Model} (Sections~\ref{sec:model}--\ref{sec:oracle}): multi-turn semantic caching over conversation states with a learned reuse risk and a price for unacceptable reuse. Given the costs and risks, Reverse Greedy chooses the cache and keeps its approximation guarantee for learned, masked and asymmetric reuse losses. 
\item \textbf{Offline learning} (Section~\ref{sec:offline}): The learning algorithm CUCB-SC-MT, whose suboptimality bound holds for arbitrarily dependent turns and adds a reuse-risk term evaluated only along the comparator's decisions.
\item \textbf{Online learning} (Section~\ref{sec:online}): Our proposed algorithm CLCB-LS-MT, which freezes its policy within each conversation, labels a random fraction of reuses and rarely recomputes its cache. Its regret is $\tilde O(\sqrt N)$ over $N$ requests when conversation states recur, although its net is built from correlated turns.
\item \textbf{Evaluation} (Section~\ref{sec:experiments}): on math tutoring (SocraTeach) and movie-ticket booking (Taskmaster-3), our algorithms have the lowest loss at every cache budget, offline and online, against ElastiCache, ContextCache, vCache, a joint-state threshold and single-turn caching. 
\end{itemize}

\paragraph{Related work.}
GPTCache~\cite{bang2023gptcache} serves the nearest cached response above a similarity threshold. For conversations, MeanCache~\cite{gill2025meancache} also compares each entry's context chain, ContextCache~\cite{contextcache} refines the match with self-attention over the dialogue history, and SmartCache~\cite{yu2025smartcache} detects context shifts from the LLM's attention scores; vCache~\cite{gaspar2026vcache} learns a threshold per entry to meet an error-rate target. None weighs a reuse against the generation cost it saves. Learning
formulations~\cite{liu2026semanticcachinglowcostllm,atalar2026continuoussemanticcachinglowcost}, similarity caching~\cite{garetto2020similarity} and online caching with regret guarantees~\cite{paschos2019learning} take the mismatch cost as given while we learn the reuse risk itself. KV and prefix
caching~\cite{kwon2023vllm,gao2024cachedattention,gim2024promptcache,zheng2024sglang} reuses computation within a request and complements response reuse, and cascades and routers~\cite{chen2024frugalgpt,ong2025routellm} choose which
model answers.  Unlike generic combinatorial bandits~\cite{chen2013combinatorial}, CUCB-SC-MT and CLCB-LS-MT choose a cache whose loss couples a learned cost with a learned risk on pairs of states, and learn from correlated turns that reveal the risk only on sampled reuses.



\section{System Model}\label{sec:model}
 




\begin{figure}[t]
  \centering
  \includegraphics[width=\linewidth]{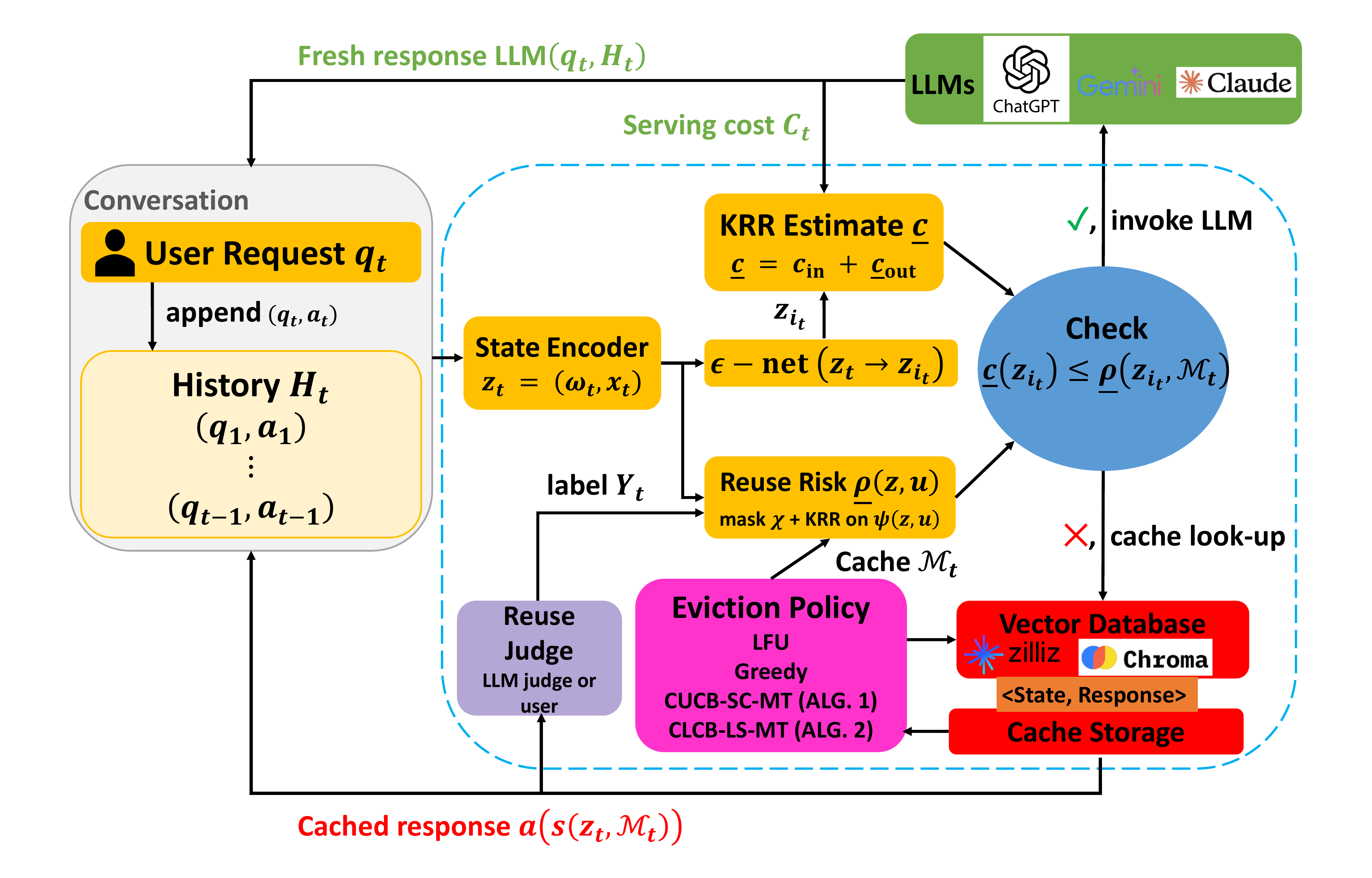}
  \caption{Multi-turn semantic caching system.} 
  \label{fig:system}
\end{figure}

We extend the continuous single-turn model
of~\citet{atalar2026continuoussemanticcachinglowcost} to multi-turn conversations. Figure~\ref{fig:system} shows the serving path. For a request $q_t$ with history $H_t$, the \emph{state encoder} computes a fixed-size representation $z_t=(\omega_t,x_t)$, on which the cache lookup operates instead of on $q_t$: the history must enter the lookup, since the same message calls for different replies in different conversations, but in compressed form, since histories grow with every turn. 
The \emph{$\epsilon$-net} maps $z_t$ to a \emph{representative} $z_{i_t}$, so that costs and risks are learned per representative rather than per request. The \emph{check} compares the estimated cost of a fresh response, $c=c_{\mathrm{in}}+c_{\mathrm{out}}$ with $c_{\mathrm{out}}$ estimated by kernel ridge regression (KRR), with the estimated loss $\rho$ of the best cached response, which combines a compatibility mask $\chi$ with a reuse risk learned on pair features $\psi(z,u)$. Either the LLM is invoked and its serving cost $C_t$ observed, or the cached response $a(s(z_t,\mathcal M_t))$ is returned from the vector database, at a fraction of the latency. A \emph{reuse judge} (an LLM judge or the user) labels a sample of reuses for the risk model, and an \emph{eviction policy} keeps at most $k$ items within the memory and lookup budget. Our algorithms (Algorithms~\ref{alg:offline} and~\ref{alg:online}) are such policies; least frequently used (LFU) and greedy thresholding are common alternatives. We now formalize these components.

\subsection{Multi-Turn LLM Serving System}\label{sec:model-serving} 
 

\paragraph{Conversations and states.}
Conversation $j$ has $1\le T_j\le T_{\max}$ turns. At turn $t$ the user
submits a request $q_{j,t}$, a natural-language message such as a question
or an instruction, and receives a response $a_{j,t}$, which is either
generated by the LLM or served from the cache. When invoked, the LLM receives $q_{j,t}$ together with the conversation history $H_{j,t}:=\big((q_{j,1},a_{j,1}),\dots,(q_{j,t-1},a_{j,t-1})\big)$, with
$H_{j,1}:=\emptyset$. The same request calls for different responses under
different histories (contexts), so we cache over
\emph{conversation states} $(q_{j,t},H_{j,t})$ rather than requests. A
cache item stores the fixed-size representation of its state, defined next,
and its response, so its size does not grow with the history, and the cache
capacity $k$ counts items, as in single-turn caches.

\paragraph{State representation.}

The state encoder represents the conversation state of each turn by $z=(\omega,x)$, which we also call a state. The \emph{exact-match key} $\omega\in\Omega$ holds the discrete facts that decide whether a response is admissible at all: the entities the request refers to after resolving the history, and explicit output constraints, e.g., $\omega=\{\text{city: Paris},\ \text{language: French}\}$. Every state has exactly one key, which is empty when the request names no entity or constraint, and $\Omega$ is the finite set of keys that occur with positive probability. The \emph{embedding} $x\in\mathcal X_\omega\subset\mathbb R^{d_e}$ concatenates unit-norm sentence embeddings of the request and the history
with a fixed weight $w\in(0,1)$: $x=[\sqrt{w}\,e(q);\sqrt{1-w}\,e(H)]$. We write $z_{j,t}=(\omega_{j,t},x_{j,t})$ and
$\mathcal Z:=\bigcup_{\omega\in\Omega}\mathcal Z_\omega$ with strata $\mathcal Z_\omega:=\{\omega\}\times\mathcal X_\omega$, and measure distance only within a stratum: $d(z,z'):=\|x-x'\|_2$ if $\omega=\omega'$ and $d(z,z'):=+\infty$ otherwise. Replacing ``Paris'' by ``Lyon''  barely changes the embedding but can make a cached response wrong, so keeping such facts in $\omega$ lets the cost and reuse risk vary smoothly with $x$ inside each stratum, so nearby states with the same key can share estimates.  

\paragraph{Conversation Arrivals.}

Conversations are i.i.d.\ draws of a random sequence
$(T,Z_1,\dots,Z_T)$ of $T\le T_{\max}$ states whose distribution is unknown. Turns within a conversation
may be arbitrarily dependent. We evaluate the system under the
\emph{per-request law}
\begin{equation}\label{eq:perrequest}
  P(B):=\frac{\mathbb E\big[\sum_{t=1}^{T}\mathbf 1\{Z_t\in B\}\big]}
             {\mathbb E[T]},\qquad B\subseteq\mathcal Z,
\end{equation}
which, by the strong law of large numbers, is the long-run fraction of requests falling in $B$. For $T\equiv1$, it is the query distribution $f$ of~\citet{atalar2026continuoussemanticcachinglowcost}. With $Z=(\omega_Z,X)\sim P$ and $P_\omega$ the law of $X$ given $\omega_Z=\omega$, we assume intrinsic dimension $p\le d_e$: there are $c_0,\bar r>0$ such that
$P(\|X-x\|_2\le r\mid\omega_Z=\omega)\ge c_0r^{p}$ for all $\omega\in\Omega$, $x\in\operatorname{supp}P_\omega$ and $r\le\bar r$.


We assume that serving a request from the cache does not change how the conversation continues. Single-turn queries satisfy this by construction, as they are drawn independently from $f$~\cite{atalar2026continuoussemanticcachinglowcost}; in conversations the response at turn $t$ enters $H_{j,t+1}$, but the assumption is accurate when reused responses are acceptable substitutes for fresh ones, so that users continue as they would have, and unacceptable reuses are rare: online, they affect 8.2\% and 13.1\% of the requests served by CLCB-LS-MT (Table~\ref{tab:reuse-key}). It also matches how semantic caches are evaluated, by replaying logged queries or conversations~\cite{gill2025meancache,contextcache,gaspar2026vcache}. Relaxing it turns caching into a reinforcement-learning problem, which we leave to future work.


\begin{assumption}[Exogenous conversation states]\label{ass:exogenous}
The law of the conversation process does not depend on how earlier turns
were served, i.e., on whether, and with which cached responses, previous
requests were answered.
\end{assumption}

\paragraph{Serving costs and partial observations.}

Invoking the LLM at $z$ has expected cost
$c(z)=c_{\mathrm{in}}(z)+c_{\mathrm{out}}(z)\in(0,1]$. The input cost
$c_{\mathrm{in}}$ of processing $q$ and $H$ grows with the conversation and
is known before the call while the generation cost $c_{\mathrm{out}}$ is unknown.
Invoking the LLM at turn $(j,t)$ reveals $C_{j,t}=c(z_{j,t})+\eta_{j,t}$; otherwise no
cost is observed. The noise $\eta_{j,t}$ is conditionally $R_c$-sub-Gaussian
given the $\sigma$-field $\mathcal F^-_{j,t}$ of all states, decisions and
feedback at earlier turns (ordered by conversation) together
with $z_{j,t}$.

\paragraph{Cache items, compatibility and reuse risk.}

The cache $\mathcal M$ holds at most $k$ items $(u,a(u))$: a \emph{source state} $u\in\mathcal Z$ at which the LLM was invoked and the response $a(u)$
generated there. We write $z$ for the state being served and $u$ for the state whose response is reused. Whether $a(u)$ may be returned at $z$ cannot be read off the embedding distance alone. A known \emph{compatibility mask} $\chi(z,u)=\bar\chi(\omega_z,\omega_u)\in\{0,1\}$ excludes reuse that is never admissible, e.g., $\bar\chi(\omega,\omega')=\mathbf 1\{\omega=\omega'\}$,
and the unknown \emph{reuse risk} $r(z,u)\in[0,1]$, the probability that $a(u)$ is unacceptable at $z$ for an LLM judge or the user, is learned from pair features $\psi(z,u)\in\mathbb R^{d_\psi}$. The \emph{reuse loss} is
\begin{equation}\label{eq:reuseloss}
  \rho(z,u):=
  \begin{cases}
    \min\{1,\gamma\,r(z,u)\}, & \chi(z,u)=1,\\
    1, & \chi(z,u)=0,
  \end{cases}
\end{equation}
where the known price $\gamma>0$ converts risk into units of normalized cost.
As with $\phi\le1$ in~\citet{atalar2026continuoussemanticcachinglowcost}, the
cap never changes $\min\{c,\rho\}$ since $c\le1$. We write
$\rho(z,\mathcal M):=\min_{u\in\mathcal M}\rho(z,u)$, with
$\rho(z,\emptyset):=1$, and
$s(z,\mathcal M):=\operatorname*{arg\,min}_{u\in\mathcal M}\rho(z,u)$.
Reusing $u$ at $z$ may reveal a label $Y\in[0,1]$ with conditional mean
$r(z,u)$, e.g., a judge's verdict on a sampled reuse. The reverse feedback
of~\cite{atalar2026continuoussemanticcachinglowcost} thus applies to both
unknowns: costs are observed only on LLM calls, and risk labels only on
reused or judged pairs.

\subsection{Multi-Turn Semantic Caching}\label{sec:model-caching}

At each turn the system either returns $a(s(z,\mathcal M))$ and incurs
$\rho(z,\mathcal M)$, or invokes the LLM on $(q,H)$ and incurs and observes
its cost. The optimal
rule is
\begin{equation}\label{eq:decision}
  a_{j,t}=
  \begin{cases}
    \mathrm{LLM}(q_{j,t},H_{j,t}), & \text{if } c(z_{j,t})\le\rho(z_{j,t},\mathcal M),\\
    a\big(s(z_{j,t},\mathcal M)\big), & \text{otherwise}.
  \end{cases}
\end{equation}
A cache therefore incurs the per-state loss
$g_{\mathcal M}(z):=\min\{c(z),\rho(z,\mathcal M)\}$ and the expected loss
\begin{equation}\label{eq:loss}
  \mathcal L(\mathcal M):=\mathbb E_{Z\sim P}\big[g_{\mathcal M}(Z)\big],
\end{equation}
which, by~\eqref{eq:perrequest} and Assumption~\ref{ass:exogenous}, is the
long-run average cost per request of a fixed cache. The cache design problem
is
\begin{equation}\label{eq:design}
  \mathcal M^*\in\operatorname*{arg\,min}_{\mathcal M\subseteq\mathcal U,\;|\mathcal M|\le k}\mathcal L(\mathcal M),
\end{equation}
where $\mathcal U\subseteq\operatorname{supp}P$ is the set of cacheable
source states.

\paragraph{Discretization via a stratified $\epsilon$-net.}

A \emph{stratified $\epsilon$-net} $\mathcal S=\{z_1,\dots,z_m\}\subseteq\mathcal Z$ contains, for every $z\in\operatorname{supp}P$, a \emph{representative} $z_i$ with $d(z,z_i)\le\epsilon$. Since $d=\infty$ across strata, each stratum has its own representatives $m\ge|\Omega|$. The Voronoi cells $V_i:=\{z:\,i=\min\operatorname*{arg\,min}_{i'}d(z,z_{i'})\}$ partition
$\mathcal Z$ without crossing strata and have weights $w_i:=P(V_i)$. With $c_i:=c(z_i)$ and
$R_{ij}:=\rho(z_i,z_j)$, the discretized loss of $\mathcal M\subseteq[m]$ is

\begin{equation}\label{eq:discloss}
  \widehat{\mathcal L}(\mathcal M;\mathcal S,\boldsymbol w,\boldsymbol c,\boldsymbol R)
  :=\sum_{i=1}^{m}w_i\min\Big\{c_i,\ \min_{j\in\mathcal M}R_{ij}\Big\}.
\end{equation}
Unlike in~\citet{atalar2026continuoussemanticcachinglowcost},
$\boldsymbol R$ is not a function of distance: it is learned, equals $1$
wherever $\bar\chi(\omega_i,\omega_j)=0$, and need not be symmetric. Caches are chosen among the representatives, whose states and responses they store, and a request in cell $V_i$ is routed by the costs and risks estimated at $z_i$.



\section{Oracle Caching with Known Parameters}\label{sec:oracle}

We first assume that $P$, $c$ and $r$ are known. Problem~\eqref{eq:design} is NP-hard, as it contains the single-turn problem of~\citet{atalar2026continuoussemanticcachinglowcost} (Lemma~\ref{lem:nphard}). We therefore minimize the discretized loss~\eqref{eq:discloss} with Reverse Greedy~\cite{liu2026semanticcachinglowcostllm} (Algorithm~\ref{alg:rg} in Appendix~\ref{app:oracle}), which starts from all representatives and repeatedly removes the one whose removal increases the loss the least. Its guarantee uses no metric property of the reuse costs, so it holds for learned, masked and asymmetric reuse losses.
 
\begin{lemma}[Approximation guarantee~\cite{liu2026semanticcachinglowcostllm}]\label{lem:supermod}
For any $\boldsymbol w\in\mathbb R_{\ge0}^m$, $\boldsymbol c\in[0,1]^m$ and
$\boldsymbol R\in[0,1]^{m\times m}$, the discretized loss
$\widehat{\mathcal L}$ is non-increasing and supermodular on $2^{[m]}$, and
Reverse Greedy returns $\mathcal M$ with
$\widehat{\mathcal L}(\mathcal M)\le\alpha\,\widehat{\mathcal L}(\mathcal M')$
for every $|\mathcal M'|\le k$, where $\alpha=(e^{\Gamma}-1)/\Gamma$, $\Gamma=s/(1-s)$, and $s\in[0,1)$ is
the curvature of the instance~\cite{il2001approximation}.
\end{lemma}
 
Relating the discretized loss to the true loss needs regularity, but only within a key.

\begin{assumption}[Regularity within keys]\label{ass:lipschitz}
There are constants $L_c,L_r,L_{\mathrm{src}}\ge0$ such that for all
states $z,z'$ with the same key and all source states $u,u'$ with the same
key,
\[
\begin{aligned}
|c(z)-c(z')|&\le L_c\|x-x'\|_2,\\
|r(z,u)-r(z',u)|&\le L_r\|x-x'\|_2,\\
|r(z,u)-r(z,u')|&\le L_{\mathrm{src}}\|x_u-x_{u'}\|_2.
\end{aligned}
\]
\end{assumption}
 

The first two conditions are Assumption~1
of~\citet{atalar2026continuoussemanticcachinglowcost} within a key; across keys $\rho$ jumps to $1$. The third is new: the cache stores responses generated at representatives, not at the items of $\mathcal M^*$, and without it $\mathcal M^*$ could rely on a response that no representative reproduces. It holds automatically when the risk depends on the source only through its distance to the target, as in the single-turn model. Under Assumption~\ref{ass:lipschitz}, $g_{\mathcal M}(\omega,\cdot)$ is $L_g$-Lipschitz with $L_g:=\max\{L_c,\gamma L_r\}$, so the discretized loss is within $L_g\epsilon$ of the true loss (Lemma~\ref{lem:disc}).

 

\begin{theorem}[Approximation guarantee]\label{thm:oracle}
Let $\mathcal M$ be returned by Reverse Greedy on
$(\mathcal S,\boldsymbol w,\boldsymbol c,\boldsymbol R)$ and let
$\mathcal M^*$ solve~\eqref{eq:design}. Under
Assumption~\ref{ass:lipschitz},
\[
\mathcal L(\mathcal M)\le\alpha\,\mathcal L(\mathcal M^*)+(1+\alpha)L_g\epsilon+\alpha\gamma L_{\mathrm{src}}\epsilon .
\]
\end{theorem}
 

The last term is the price of replacing each item of $\mathcal M^*$ by its nearest representative with the same key. In the single-turn model the bound becomes the $(1+2\alpha)L_g\epsilon$ of Theorem~3.3 in~\citet{atalar2026continuoussemanticcachinglowcost}.
 
\begin{lemma}[Size of a stratified net]\label{lem:netsize}
If the representatives lie in $\operatorname{supp}P$, representatives with
the same key are more than $\epsilon$ apart, and $\epsilon\le2\bar r$, then
$m\le|\Omega|\,2^{p}/(c_0\epsilon^{p})$.
\end{lemma}
 

This replaces the $O(\epsilon^{-d_e})$ of the single-turn analysis by the intrinsic dimension $p$, at the price of a factor $|\Omega|$, which favors coarse keys that contain only facts that change the answer. Proofs for this section are in Appendix~\ref{app:oracle}.

\section{Offline Learning with Unknown Parameters}\label{sec:offline}

None of $P$, $c$ and $r$ is known when a cache is deployed: they depend on the application's traffic, on the length of responses not yet generated, and on what its users accept. We therefore learn them from $J$ logged conversations. For every turn the logs record the state $z_{j,t}$, the cache in use and the realized cost $C_{j,t}$ if the LLM was invoked. They are complemented by $n_r$ labeled pairs $\{(z_s,u_s,Y_s)\}_{s=1}^{n_r}$ from an LLM judge run on logged reuses or on pairs of logged states. As in~\citet{atalar2026continuoussemanticcachinglowcost}, we compete with the $\alpha$-approximate oracle and evaluate the serving rule actually deployed, whose expected loss is
$\mathcal L^{\mathrm{dep}}(\widehat{\mathcal M})\ge\mathcal L(\widehat{\mathcal M})$:
\begin{equation}\label{eq:subopt}
\mathrm{SubOpt}(\widehat{\mathcal M}):=\mathcal L^{\mathrm{dep}}(\widehat{\mathcal M})-\alpha\,\mathcal L(\mathcal M^*).
\end{equation}

\begin{algorithm}[t]
\caption{CUCB-SC-MT: Offline Multi-Turn Semantic Caching}
\label{alg:offline}
\begin{algorithmic}[1]
\REQUIRE conversations split into $\mathcal D_1,\mathcal D_2$; cost
  observations $\{(z_s,C_s)\}_{s=1}^{n_c}$; labeled pairs
  $\{(z_s,u_s,Y_s)\}_{s=1}^{n_r}$; $\epsilon$, kernels $k_c,k_r$, ridge
  $\lambda$, $\delta$, $\gamma$, mask $\bar\chi$, $k$
\STATE Build $\mathcal S=\{z_1,\dots,z_m\}$ on $\mathcal D_1$: add a state
  as a representative if it is farther than $\epsilon$ from all
  representatives with the same key
\STATE $\hat w_i\gets\frac{1}{N_2}\sum_{j\in\mathcal D_2}\sum_{t=1}^{T_j}\mathbf 1\{z_{j,t}\in V_i\}$ for all $i\in[m]$
\STATE Fit KRR with $k_c$ on $\{(x_s,C_s-c_{\mathrm{in}}(z_s))\}$, giving
  $\hat c_{\mathrm{out}},\sigma_c$; set
  $\bar c_i\gets\min\{1,\,c_{\mathrm{in}}(z_i)+\hat c_{\mathrm{out}}(z_i)+\beta_c\sigma_c(z_i)\}$
\STATE Fit KRR with $k_r$ on the compatible labeled pairs
  $\{(\psi(z_s,u_s),Y_s)\}$, giving $\hat r,\sigma_r$
\FOR{$i,j\in[m]$}
\STATE $\bar R_{ij}\gets\min\{1,\gamma(\hat r+\beta_r\sigma_r+\varepsilon_{\mathrm{app}})(z_i,z_j)\}$
  if $\bar\chi(\omega_i,\omega_j)=1$, else $\bar R_{ij}\gets1$
\ENDFOR
\STATE $\widehat{\mathcal M}\gets$ Reverse Greedy$(\mathcal S,\hat{\boldsymbol w},\bar{\boldsymbol c},\bar{\boldsymbol R},k)$
\STATE Serve $z\in V_i$ from
  $j^\circ=\arg\min_{j\in\widehat{\mathcal M}}\bar R_{ij}$ if
  $\bar R_{ij^\circ}<\bar c_i$, else invoke the LLM
\end{algorithmic}
\end{algorithm}

Algorithm~\ref{alg:offline} extends CUCB-SC-Cont
of~\citet{atalar2026continuoussemanticcachinglowcost}, which learns one function, the cost, from independent queries and takes the mismatch cost as known. Here the data differ in three ways, and each changes the algorithm. (i) Turns of a conversation are dependent, so it splits conversations, not turns, into $\mathcal D_1$, which builds the net, and $\mathcal D_2$ ($J_2$ conversations, $N_2$ turns), which estimates the weights. (ii) The reuse risk is a second unknown function, defined on pairs of states and observed only through judge labels; it is learned by KRR on pair features $\psi$, and a pessimistic reuse-loss matrix $\bar{\boldsymbol R}$ is formed for compatible pairs only ($\chi=1$, e.g., the same key; the others have $\bar R_{ij}=1$). (iii) Later states contain earlier responses, so regression inputs depend on earlier noise, and the confidence widths must be self-normalized~\cite{chowdhury2017kernelizedmultiarmedbandits} (Appendix~\ref{app:offline}). Below, $\alpha$ refers to the pessimistic instance $(\hat{\boldsymbol w},\bar{\boldsymbol c},\bar{\boldsymbol R})$.

\begin{assumption}[Logged conversations]\label{ass:logs}
Conversations in $\mathcal D_2$ are independent of each other and of $\mathcal D_1$, and
$\mathbb E\big[\sum_{t=1}^{T_j}\mathbf 1\{Z_{j,t}\in B\}\mid T_j\big]=T_j\,P(B)$ for every $B\subseteq\mathcal Z$. Turns within a conversation, and the logging decisions, may be arbitrarily dependent.
\end{assumption}
 
\begin{assumption}[Kernel regularity]\label{ass:kernel}
(a) $c_{\mathrm{out}}$ depends on $z$ through $x$ and lies in the RKHS of
a kernel $k_c\le1$ with norm at most $B_c$. (b) $r=r_0+e$, where $r_0$ lies
in the RKHS of a kernel $k_r\le1$ on $\psi(z,u)$ with norm at most $B_r$,
and $\sup|e|\le\varepsilon_0$. (c) The labeled pairs can be ordered so that
each pair is fixed before its label is drawn and
$\mathbb E[Y_s\mid\text{past}]=r(z_s,u_s)$.
\end{assumption}

Part (b) allows misspecification of size $\varepsilon_0$, with
$\varepsilon_0=0$ the realizable case. Let $\mathcal I_c$ and
$\mathcal I_r$ be the maximum information gains of $k_c$ and $k_r$ after
$n_c$ and $n_r$ observations~\cite{chowdhury2017kernelizedmultiarmedbandits}, and set
\[
\begin{aligned}
\beta_c&:=B_c+R_c\sqrt{2(\mathcal I_c+1+\ln(3/\delta))},\\
\beta_r&:=B_r+\tfrac12\sqrt{2(\mathcal I_r+1+\ln(3/\delta))},\\
\varepsilon_{\mathrm{app}}(\cdot)&:=\varepsilon_0\big(1+\sqrt{n_r/\lambda}\,\sigma_r(\cdot)\big).
\end{aligned}
\]



With probability at least $1-\delta$, the estimated weights satisfy $\|\hat{\boldsymbol w}-\boldsymbol w\|_1\le\sqrt{2(m\ln2+\ln(3/\delta))/N_{\mathrm{eff}}}$, where $N_{\mathrm{eff}}:=\big(\sum_{j\in\mathcal D_2}T_j\big)^2/\sum_{j\in\mathcal D_2}T_j^2\le J_2$, and $\bar c$ and $\bar{\boldsymbol R}$ upper-bound $c$ and $\boldsymbol R$ with gaps at most $2\beta_c\sigma_c$ and $2\gamma(\beta_r\sigma_r+\varepsilon_{\mathrm{app}})$ (Lemma~\ref{lem:conc}, Appendix~\ref{app:offline}). The weights count
conversations, not turns: $N_{\mathrm{eff}}=J_2$ when all conversations have the same length, which cannot be improved under arbitrary within-conversation dependence.

To state the main result, let the \emph{comparator}
$\mathcal M^\circ\subseteq\mathcal S$ replace each item of $\mathcal M^*$ by its nearest representative with the same key. For every $i$, let $a_i\in\{\mathrm{LLM}\}\cup\mathcal M^\circ$ attain
$g_{\mathcal M^\circ}(z_i)$ (ties go to the LLM), and define $I_L:=\{i:a_i=\mathrm{LLM}\}$, $I_R:=[m]\setminus I_L$, and $j^*(i):=a_i$ for $i\in I_R$. These are the cells where the comparator invokes the LLM and where it reuses, and the item it reuses.
 

\begin{theorem}[Suboptimality of CUCB-SC-MT]\label{thm:offline}
Suppose Assumptions~\ref{ass:exogenous}--\ref{ass:kernel} hold and the net
built on $\mathcal D_1$ covers $\operatorname{supp}P$. With probability at
least $1-\delta$,
\begin{align*}
&\mathrm{SubOpt}(\widehat{\mathcal M})\le(1+\alpha)L_g\epsilon+\alpha\gamma L_{\mathrm{src}}\epsilon\\
&\quad+\tfrac{1+\alpha}{2}\sqrt{\tfrac{2(m\ln2+\ln(3/\delta))}{N_{\mathrm{eff}}}}
+2\alpha\beta_c\textstyle\sum_{i\in I_L}w_i\sigma_c(z_i)\\
&\quad+2\alpha\gamma\textstyle\sum_{i\in I_R}w_i\big(\beta_r\sigma_r+\varepsilon_{\mathrm{app}}\big)(z_i,z_{j^*(i)}).
\end{align*}
\end{theorem}


The four terms are discretization, arrival, cost and reuse risk. The first three have the form of Theorem~4.1 in~\citet{atalar2026continuoussemanticcachinglowcost}, but each needs a new step: the arrival term concentrates over conversations rather than turns (Lemma~\ref{lem:chernoff}), the cost widths must hold although regression inputs contain earlier responses, and the discretization term pays $\alpha\gamma L_{\mathrm{src}}\epsilon$ because cached responses move with their source states. The reuse-risk term is new. Both learning terms are evaluated only along the comparator's decisions: costs where $\mathcal M^\circ$ invokes the LLM and risks on the pairs it reuses. The logs therefore need to cover what a good cache does, not every state and pair, which is the usual benefit of pessimism ~\cite{jin2021pessimism}.

\begin{corollary}[Rate]\label{cor:offline}
Use RBF kernels with length scales $\ell_c$ on $x$ and $\ell_r$ on $\psi$,
and suppose the logs cover the comparator's decisions: the LLM is invoked
with propensity at least $\nu_{\min}$ near its LLM cells, and the labeled
pairs near its reused pairs have density constant $C_R$ and intrinsic
dimension $p_r$ (Appendix~\ref{app:explicit}). For $N_2\ge N_0$ and
$\epsilon^*\asymp(|\Omega|/(c_0N_{\mathrm{eff}}))^{1/(p+2)}$, with
probability at least $1-\delta$,
\[
\begin{aligned}
\mathrm{SubOpt}=\tilde O\Big(&\big(\tfrac{|\Omega|}{N_{\mathrm{eff}}}\big)^{\frac{1}{p+2}}
+\big(\tfrac{|\Omega|}{\nu_{\min}N_2}\big)^{\frac{1}{p+2}}\\
&+\gamma\big(\tfrac{C_R}{n_r}\big)^{\frac{1}{p_r+2}}+\gamma\bar\varepsilon_{\mathrm{app}}W_R\Big),
\end{aligned}
\]
where $W_R:=\sum_{i\in I_R}w_i$ is the comparator's reuse mass and
$\bar\varepsilon_{\mathrm{app}}:=\max_{i\in I_R}\varepsilon_{\mathrm{app}}(z_i,z_{j^*(i)})$.
Appendix~\ref{app:explicit} gives the explicit bound.
\end{corollary}

 

The new term converges more slowly because it lives in pair space, with $p_r\le2p$ for concatenated target and source features. It is weighted by the comparator's reuse mass $W_R$ and depends on label coverage only through $C_R$, so labels should target plausible pairs, such as nearest compatible candidates, rather than all pairs. The term $\gamma\bar\varepsilon_{\mathrm{app}}W_R$ is the price of misspecification
and vanishes in the realizable case. With $T\equiv1$ and a known risk, Theorem~\ref{thm:offline} recovers Theorem~4.1 of~\citet{atalar2026continuoussemanticcachinglowcost} with a smaller arrival constant and an optimal radius set by $p$ rather than $d_e$ (Appendix~\ref{app:reduction}). 
 

\section{Online Learning with Unknown Parameters}\label{sec:online}

Online, the system starts without data and must build the net, learn costs and risks, and choose caches while serving. This is the usual case in deployment: a new application, or one that switches LLMs, has no judged logs to learn from. Conversations make this harder in two ways. Turns of a conversation are correlated, so a request is not a fresh draw from $P$ given the past, and the per-request concentration arguments of the single-turn analysis fail. Risk labels arrive only for a sample of the reuses that the algorithm itself chose, so it must reuse uncertain pairs to learn their risk.

\subsection{Online Learning Setup}\label{sec:online-setup}
 
\paragraph{Stream and regret.}
Conversations arrive one after another as the independent copies of
Section~\ref{sec:model}, and each is served to completion before the next
one starts. We index requests globally by rounds $n=1,\dots,N$, write $z_n$
for the state at round $n$, and let $\mathcal G_j$ be the $\sigma$-field of
the first $j$ conversations and of all feedback they generated. Let
$\bar T:=\mathbb E[T^2]/\mathbb E[T]\le T_{\max}$, the expected length of
the conversation that a random request belongs to. At round $n$ the system
either invokes the LLM, with loss $\ell_n:=c(z_n)$, or reuses a cached item
$u$, with loss $\ell_n:=\rho(z_n,u)$. 
\begin{equation}\label{eq:regret}
\begin{aligned}
\mathrm{Reg}(N):=\;&\mathbb E\Big[\sum_{n=1}^N\ell_n+\sum_{n=1}^N\sum_{u\in\mathcal M_n\setminus\mathcal M_{n-1}}c(u)\Big]\\
&-\alpha N\mathcal L(\mathcal M^*),
\end{aligned}
\end{equation}


with $\mathcal M_0=\emptyset$ and $\mathcal M^*$ from~\eqref{eq:design}.
As in~\citet{atalar2026continuoussemanticcachinglowcost}, the second sum charges $c(u)$ for every inserted item, whose response must be generated (in Algorithm~\ref{alg:online}, when the representative is created). \eqref{eq:regret} compares with the best cache over all cacheable states and, in the single-turn case, upper-bounds the regret of~\citet{atalar2026continuoussemanticcachinglowcost}.

 
\begin{assumption}[Online feedback]\label{ass:online}
(a) On each reuse round, a label $Y_n$ is observed with probability
$\zeta\in(0,1]$, independently of the past given the served pair, and
$\mathbb E[Y_n\mid\text{past},z_n,u]=r(z_n,u)$. (b) For every source state
$u$, $x_z\mapsto\psi(z,u)$ is $L_\psi$-Lipschitz within a key.
\end{assumption}
 
Part (a) models an LLM judge run on a random sample of reuses, and part (b) holds with $L_\psi=1$ when $\psi$ concatenates the embeddings.

\subsection{Algorithm Design of CLCB-LS-MT}\label{sec:online-alg}
 
\begin{algorithm}[t]
\caption{CLCB-LS-MT: Online Multi-Turn Semantic Caching with Low Switching}
\label{alg:online}
\begin{algorithmic}[1]
\REQUIRE horizon $N$, radius $\epsilon$, kernels $k_c,k_r$, ridge
  $\lambda$, $\delta$, $\gamma$, mask $\bar\chi$, $k$, label rate $\zeta$
\STATE $\mathcal S,\mathcal M\gets\emptyset$; cost data
  $\mathcal D^c$ and label data $\mathcal D^r\gets\emptyset$
\FOR{conversations $j=1,2,\dots$}
\FOR{each round $n$ of conversation $j$}

\STATE If $z_n$ is covered: $i\gets$ its cell,
  $u^\star\gets\arg\min_{u\in\mathcal M}\underline R_{iu}$
  ($\underline R_{iu^\star}:=1$ if $\mathcal M=\emptyset$)
\IF{$z_n$ is uncovered (no representative with key $\omega_n$ within
  $\epsilon$) or $\underline c_i\le\underline R_{iu^\star}$}
\STATE Invoke the LLM; add $(x_n,C_n-c_{\mathrm{in}}(z_n))$ to $\mathcal D^c$
\ELSE
\STATE Return $a(u^\star)$; with probability $\zeta$ observe $Y_n$ and
  add $(\psi(z_n,u^\star),Y_n)$ to $\mathcal D^r$
\ENDIF
\ENDFOR

\STATE Add the states of conversation $j$ to $\mathcal S$ greedily, as in
  Algorithm~\ref{alg:offline}
\IF{a representative was added, an arrival stage ends, or
  $\ln\det(I+K/\lambda)$ of $\mathcal D^c$ or $\mathcal D^r$ grew by
  $\ge\ln2$ since the last switch}
\STATE $\hat w_i\gets$ fraction of all past rounds in $V_i$, with the
  uncovered rounds as an extra cell
\STATE $\underline c_i\gets[c_{\mathrm{in}}(z_i)+\hat c_{\mathrm{out}}(z_i)-\beta_c\sigma_c(z_i)]_{[0,1]}$
\STATE $\underline R_{ij}\gets\min\{1,\gamma[\hat r-\beta_r\sigma_r-\varepsilon_{\mathrm{app}}]_+(z_i,z_j)\}$
  if $\bar\chi(\omega_i,\omega_j)=1$, else $\underline R_{ij}\gets1$
\STATE $\mathcal M\gets$ Reverse Greedy$(\mathcal S,\hat{\boldsymbol w},\underline{\boldsymbol c},\underline{\boldsymbol R},k)$
\ENDIF
\ENDFOR
\end{algorithmic}
\end{algorithm}

Algorithm~\ref{alg:online} extends CLCB-SC-LS-Cont
of~\citet{atalar2026continuoussemanticcachinglowcost}, keeping its optimism
and low switching, with four changes. (i)~\emph{Per-conversation
snapshots.} Each conversation $j$ is served with a frozen snapshot
$\Pi_j=(\mathcal S_j,\hat{\boldsymbol w}_j,\underline{\boldsymbol c},\underline{\boldsymbol R},\mathcal M_j)$,
and all updates happen at conversation boundaries. A request is correlated
with the earlier turns of its conversation, but a whole conversation visits
every region $B$ $\mathbb E[T]P(B)$ times in expectation
by~\eqref{eq:perrequest}, so freezing the policy within a conversation lets
the analysis average its loss against $P$. (ii)~\emph{Routing at the
representative.} A covered request takes the option that the lower
confidence bounds choose for its cell representative; an uncovered request
goes to the LLM and may become a representative. Optimism keeps uncertain cells receiving LLM calls and uncertain pairs being reused and labeled. (iii)~\emph{Learned risk with
sampled labels.} The risk is learned by KRR from the sampled labels, with
$\beta_c$, $\beta_r$ and $\varepsilon_{\mathrm{app}}$ as in
Section~\ref{sec:offline}, the information gains taken after $N$
observations and $n_r$ the number of labels at the snapshot.
(iv)~\emph{Low switching.} The cache changes only at a conversation boundary
where a representative was added, an arrival stage ended, or $\ln\det(I+K/\lambda)$ of the cost or label data grew by at least $\ln2$ since the last switch. Unlike the per-representative counters of~\citet{atalar2026continuoussemanticcachinglowcost}, this \emph{information trigger} accounts for how one KRR observation shrinks the widths of all nearby representatives and pairs (Appendix~\ref{app:online}).


\subsection{Theoretical Results}\label{sec:online-results}
 
 

\begin{figure*}[t]
    \centering
    \includegraphics[width=\textwidth,trim=0 380.9 0 0,clip]{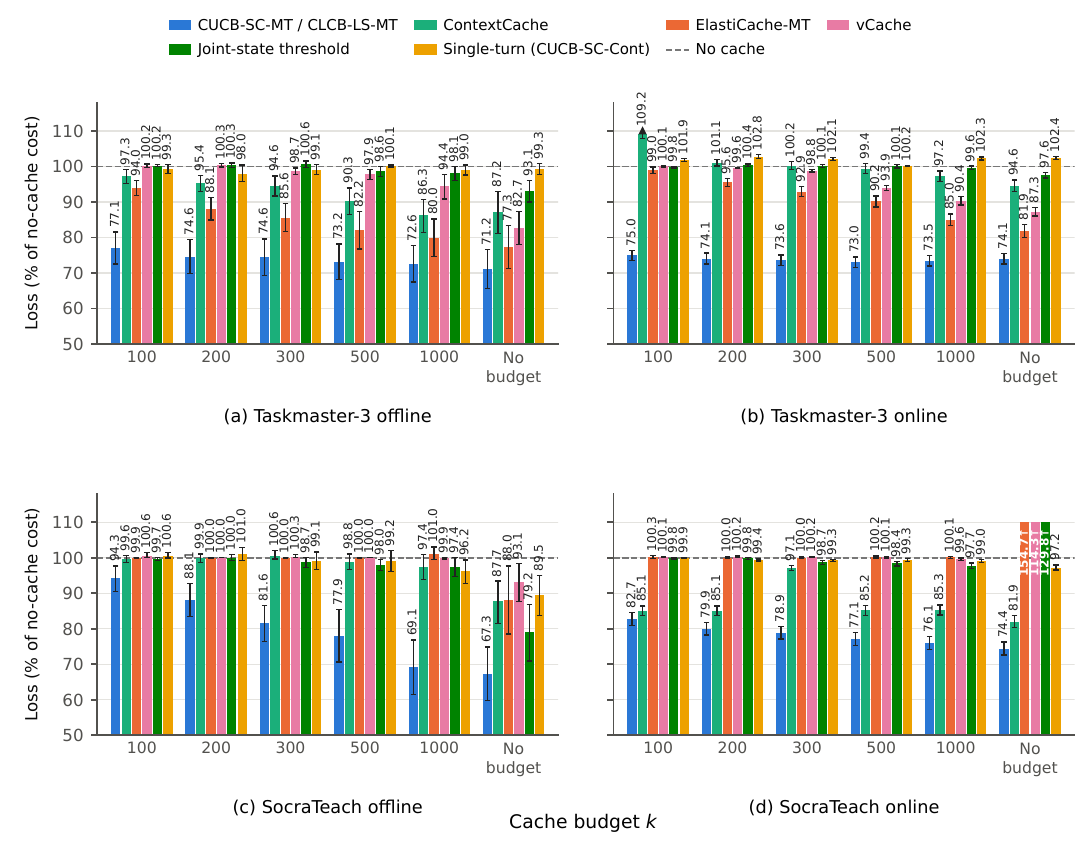}\par\nointerlineskip
    \includegraphics[width=\textwidth,trim=0 203.4 0 42.5,clip]{figures/loss_by_budget_four_panels.pdf}\par\nointerlineskip
    \includegraphics[width=\textwidth,trim=0 0 0 233,clip]{figures/loss_by_budget_four_panels.pdf}
    \caption{Loss at $\gamma=1$ as a percentage of the no-cache loss (lower is better)
    by cache budget $k$, offline on the test conversations (a, c) and online over
    the whole stream (b, d). Blue bars are CUCB-SC-MT offline and CLCB-LS-MT online.
    Error bars are 95\% bootstrap CIs over conversations (2,000 resamples); the dashed line is no
    caching; bars above 110\% are clipped and labeled.}
    \label{fig:loss_by_budget}
\end{figure*}

\begin{figure*}[t]
    \centering
    \includegraphics[width=\textwidth,trim=0 369.5 0 0,clip]{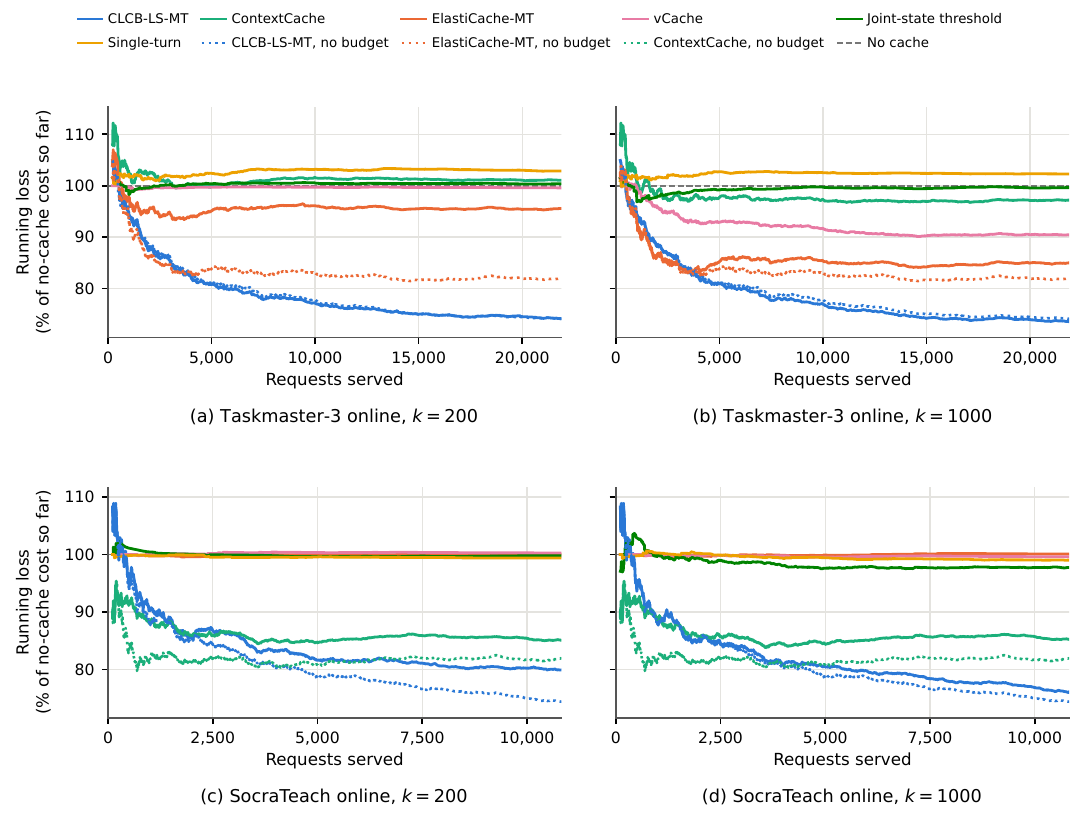}\par\nointerlineskip
    \includegraphics[width=\textwidth,trim=0 189 0 48.5,clip]{figures/online_running_loss_four_panels.pdf}\par\nointerlineskip
    \includegraphics[width=\textwidth,trim=0 0 0 230.5,clip]{figures/online_running_loss_four_panels.pdf}
    \caption{Running online loss at $\gamma=1$: the cumulative loss divided by the cumulative no-cache cost after each request, for budgets $k=200$ (a, c) and $k=1000$ (b, d). Dotted curves show CLCB-LS-MT and the strongest unbounded baseline (ElastiCache-MT on Taskmaster-3, ContextCache on SocraTeach) without a budget. The dashed line is no caching.}
    \label{fig:online_running_loss}
\end{figure*}


\begin{theorem}[Regret of CLCB-LS-MT]\label{thm:online}
Suppose Assumptions~\ref{ass:exogenous}, \ref{ass:lipschitz},
\ref{ass:kernel}(a)--(b) and~\ref{ass:online} hold, $\epsilon\le2\bar r$
and $\delta=1/N$, and let $m_{\max}:=|\Omega|2^p/(c_0\epsilon^p)$ bound the net size (Lemma~\ref{lem:netsize}). Let $Q_N:=\mathbb E\sum_n\|x_n-x_{i_n}\|_2\mathbf 1\{z_n\text{ covered}\}\le\epsilon N$
be the quantization of the served requests, and let
$\epsilon^{\mathrm{proj}}_n$ be the largest distance from an item of
$\mathcal M^*$ to the net at round $n$,
set to $0$ where it exceeds $3\epsilon/2$. Then
\begin{align*}
\mathrm{Reg}(N)\le\;&4\beta_c\sqrt{\tfrac{N\mathcal I_c}{\ln(1+1/\lambda)}}
+4\gamma\beta_r\sqrt{\tfrac{N\mathcal I_r}{\zeta\ln(1+1/\lambda)}}+E_{\mathrm{mis}}\\
&+(1+\alpha)(N_{\mathrm{arr}}+2)\big(\sqrt{24\bar T\Lambda}+\sqrt{t_1}\big)\sqrt N\\
&+L_{\mathrm{disc}}Q_N+\alpha\gamma L_{\mathrm{src}}\textstyle\sum_n\mathbb E\,\epsilon^{\mathrm{proj}}_n\\
&+k\,(m_{\max}+N_{\mathrm{arr}}+N_c+N_r)+R_{\mathrm{low}},
\end{align*}
where $L_{\mathrm{disc}}:=(1+\alpha)L_g+2\max\{\beta_c/\ell_c,\allowbreak\gamma\beta_rL_\psi/\ell_r\}$,
$\Lambda:=m_{\max}\ln(2NT_{\max})+\ln(32N/\delta)$,
$t_0:=\max\{\mathbb E[T],\allowbreak T_{\max}^2\Lambda/(2\mathbb E[T])\}$ and
$t_1:=\max\{2T_{\max}(m_{\max}+1+\Lambda),\allowbreak t_0\}$,
$E_{\mathrm{mis}}:=2\gamma\,\mathbb E\sum_{n:\text{reuse}}\varepsilon_{\mathrm{app}}$
vanishes when $\varepsilon_0=0$, and the remainder $R_{\mathrm{low}}$
(Appendix~\ref{app:thm-online}) grows with $N$ only logarithmically for
fixed $\epsilon$.
\end{theorem}

\begin{table}[t]
\centering
\caption{Online reuse quality at $k=1000$ (\%). Prec.: share of hits the
judge accepts; False: unacceptable reuses as a share of all requests; SCA: safe
cost avoided (Section~\ref{sec:exp-reuse}). Appendix~\ref{app:reuse} gives every
budget and no budget.}
\label{tab:reuse-key}
\vspace{4pt}
\small
\setlength{\tabcolsep}{4pt}
\begin{tabular}{l rrr rrr}
\toprule
& \multicolumn{3}{c}{Taskmaster-3} & \multicolumn{3}{c}{SocraTeach} \\
\cmidrule(lr){2-4}\cmidrule(lr){5-7}
Method & Prec. & False & SCA & Prec. & False & SCA \\
\midrule
CLCB-LS-MT       & 74.6 & 8.2 & \textbf{35.0} & 67.5 & 13.1 & \textbf{41.0} \\
ContextCache     & 62.0 & 9.5 & 17.9 & 76.9 & 6.9 & 25.1 \\
ElastiCache-MT   & 82.5 & 3.6 & 22.3 & 95.6 & 0.3 &  0.8 \\
vCache           & 81.1 & 2.8 & 14.7 & 78.4 & 0.1 &  0.7 \\
Joint-state thr. & 70.7 & 1.0 &  2.3 & 83.5 & 2.0 &  6.0 \\
Single-turn      & 29.1 & 2.0 &  1.1 & 84.7 & 0.5 &  2.0 \\
\bottomrule
\end{tabular}
\end{table}



The cost and risk terms are the elliptical-potential terms of kernelized
bandits, and the arrival term has the $\sqrt{m_{\max}N}$ rate of the
single-turn analysis, inflated by the conversation length and by the
$N_{\mathrm{arr}}=O(\log\log N)$ arrival stages; $N_c\le2\mathcal I_c/\ln2$
and $N_r\le2\mathcal I_r/\ln2$ count the information triggers. The
discretization term covers the quantization of the served requests, the
transfer of widths from representatives to requests, and the projection of
$\mathcal M^*$ onto the net. For deployment, a smaller $\epsilon$ lowers the
discretization term but enlarges the net ($m_{\max}\propto\epsilon^{-p}$),
and labeling a fraction $\zeta$
of reuses saves judge calls at a factor $1/\sqrt\zeta$ in the risk term.
 
Unlike the single-turn proof (Appendix~\ref{app:online}), the error
$\|\hat{\boldsymbol w}-\boldsymbol w\|_1$ of the estimated cell weights must
concentrate over conversations, as the net is built from correlated turns
(Lemma~\ref{lem:arr-online}), and the confidence widths
$\sigma_c,\sigma_r$ frozen within a conversation must stay within a factor
$\sqrt2$ of current ones, which the information triggers ensure except on
$T_{\max}(N_c+N_r)$ rounds (Lemmas~\ref{lem:frozen}--\ref{lem:switch-online}).

\begin{corollary}[Regret rates]\label{cor:online}
Under the conditions of Theorem~\ref{thm:online}:
(a)~\emph{No structural assumption.} With $\epsilon$ tuned to $N$
(Appendix~\ref{app:cor-online}),
$\mathrm{Reg}(N)=\tilde O\big(N^{(p+1)/(p+2)}+\beta_c\sqrt{N\mathcal I_c}
+\gamma\beta_r\sqrt{N\mathcal I_r/\zeta}\big)+E_{\mathrm{mis}}$.
(b)~\emph{Separated support.} If, for every key $\omega$, $P_\omega$
is supported on finitely many points that are pairwise more than
$\epsilon$ apart, and $m_{\max}$ denotes their number, then $Q_N=0$,
$\mathcal I_c=O(m_{\max}\log N)$, $\mathcal I_r=O(m_{\max}^2\log N)$, and
$\mathrm{Reg}(N)=\tilde O(\sqrt N)+E_{\mathrm{mis}}$.
\end{corollary}

In the realizable case $\varepsilon_0=0$, part (b) gives $\tilde O(\sqrt N)$ regret when conversation states, including their histories, recur, and the same rate holds whenever
$Q_N+\sum_{n}\mathbb E[\epsilon_n^{\mathrm{proj}}]=\tilde O(\sqrt N)$. For general support, part (a) tunes $\epsilon$ to $N$ and gives the per-request
rate $\tilde O(N^{-1/(p+2)})$ of Corollary~\ref{cor:offline}. In the single-turn case, part (b) matches the horizon dependence of Theorem~5.1 in~\citet{atalar2026continuoussemanticcachinglowcost} (Appendix~\ref{app:online}, which also contains the proofs).

\begin{table}[t]
\centering
\caption{Online estimated latency relative to $t_{\mathrm{LLM}}$ (858\,ms on
Taskmaster-3, 1,100\,ms on SocraTeach; lower is better), including background
work. Appendix~\ref{app:latency} gives every budget, the hit rates and the
offline estimates.}
\label{tab:online-latency-main}
\vspace{4pt}
\small
\setlength{\tabcolsep}{4pt}
\begin{tabular}{l cc cc}
\toprule
& \multicolumn{2}{c}{Taskmaster-3} & \multicolumn{2}{c}{SocraTeach} \\
\cmidrule(lr){2-3}\cmidrule(lr){4-5}
Method & \makecell{$k=100$\\--$1000$} & \makecell{No\\budget}
       & \makecell{$k=100$\\--$1000$} & \makecell{No\\budget} \\
\midrule
CLCB-LS-MT       & 0.80--0.85 & 0.81 & 0.65--0.70 & 0.59 \\
ContextCache     & 0.76--0.87 & 0.75 & 0.72--0.94 & 0.53 \\
ElastiCache-MT   & 0.83--0.94 & 0.77 & 0.95--1.00 & 0.10 \\
vCache           & 0.90--1.04 & 0.92 & 1.01       & 0.65 \\
Joint-state thr. & 1.05--1.08 & 0.99 & 0.91--1.02 & 0.41 \\
Single-turn      & 1.01--1.05 & 1.03 & 0.99--1.02 & 0.89 \\
\bottomrule
\end{tabular}
\end{table}

\section{Experiments}
\label{sec:experiments}

\paragraph{Workloads.}

Multi-turn caching needs responses that are reusable across conversations,
which is rare in open domain chat: in WildChat 1M~\cite{zhao2024wildchat1mchatgptinteraction}, every tested cache could safely serve only 1--2\% of follow-up turns from another conversation (Appendix~\ref{app:screening}). We therefore use two structured workloads in which history matters. For identical messages with different histories, a response from another conversation is acceptable only 22\% of the time on SocraTeach and 45\% on Taskmaster-3.

\paragraph{SocraTeach.}
SocraTeach~\cite{socrateach} pairs a GPT-4 tutor with a GPT-4-simulated student on GSM8K and MAWPS problems. We sample five dialogues for each of 440 problems (2,200 dialogues, 10,821 requests), as for a class working through a shared assignment. Shared openings carry only 2\% of the cost; savings must come from follow-ups such as ``12 cookies'', whose reply depends on the problem and step.

\paragraph{Taskmaster-3.}
Taskmaster-3~\cite{byrne2021tickettalk} contains crowd-written movie-ticketing dialogues that follow a common booking flow. Follow-ups such as ``Yes, that's right'' recur across conversations, but the correct reply depends on the movie, theater and time chosen earlier. We sample 2,200 dialogues with at least two user turns (21,892 requests); responses are the logged, human-written agent turns.

\paragraph{Costs, acceptability and loss.}
The cost $c(z_t)$ of request $q_t$ with history $H_t$ counts the input and output tokens of its LLM call (GPT-4o-mini tokenizer), normalized to $[0,1]$ by the 1st and 99th percentiles of the calibration set. States are embedded with
Granite~\cite{awasthy2025graniteembeddingmodels} as $x_t=[\sqrt{0.7}\,e(q_t);\sqrt{0.3}\,e(H_t)]$. GPT-4o-mini (temperature 0) judges whether a cached response is acceptable for a request and its history \cite{zheng2023judging}. All methods share the judge and its labels (on SocraTeach it accepts 98\% of
exact-context duplicates and 2.7\% of random pairs). A miss costs $c(z_t)$ and a hit costs $\gamma Y_t$, where $Y_t=1$ if the judge rejects the reuse. We report the total loss at $\gamma=1$ as a percentage of the no-cache loss.

\paragraph{Baselines.}
ElastiCache-MT~\cite{aws_elasticache_doc} thresholds the cosine similarity of an embedding of the recent messages; ContextCache~\cite{contextcache} retrieves by the current message and then matches the recent user messages; vCache~\cite{gaspar2026vcache} learns per-entry thresholds from correctness feedback to keep its error rate below~$\delta$. A joint-state threshold on our $x_t$ isolates the state representation from CUCB-SC-MT's cache design and risk bounds, and single-turn
CUCB-SC-Cont~\cite{atalar2026continuoussemanticcachinglowcost} uses the current message alone (Appendix~\ref{app:baselines}).

\paragraph{Offline protocol.}
Each workload is split by conversation into 1,800 calibration, 200 validation and 200 test conversations. All hyperparameters (net radii, similarity thresholds, vCache's $\delta$, risk multipliers) minimize the validation loss at
$\gamma=1$ with caches built from calibration data, on which Algorithm~\ref{alg:offline} also fits its risk model from judge-labeled pairs. The selected configurations are rebuilt on calibration and validation data (19,932 source states on Taskmaster-3, 9,822 on SocraTeach) and tested on 1,960 and 999 requests at $k\in\{100,200,300,500,1000\}$ and without a budget, where the baselines cache every source state and CUCB-SC-MT and single-turn every $\epsilon$-net center.
Validation also selects the risk multiplier $\hat\gamma$ that replaces $\gamma$ in $\bar R$: the analyzed $\hat\gamma=1$, except on SocraTeach at $k\ge500$ and without a budget ($\hat\gamma=0.75$; $\hat\gamma=1$ would change the test loss by $+0.7$, $+4.9$ and $-1.9$\,pp).


\paragraph{Online protocol.}
All 2,200 conversations arrive in one random order, turns in order, and every method sees the same stream, starting from an empty cache with no labels. Methods observe the cost of each LLM call and the judge label of each reuse and
keep their offline-selected hyperparameters (CLCB-LS-MT uses
$\hat\gamma=\gamma=1$). CLCB-LS-MT also reuses from a \emph{session buffer} of the current conversation's responses, which Theorem~\ref{thm:online} does not cover (Appendix~\ref{app:reuse} compares the methods without it).
Losses include the cold start and CIs bootstrap conversations with the decisions fixed and so omit arrival-order variation.

\subsection{Loss across cache budgets}
\label{sec:exp-loss}
Figure~\ref{fig:loss_by_budget} reports each method's loss at its
validation-selected setting. Differences are in percentage points (pp) of the
no-cache loss.

\paragraph{Offline.}
CUCB-SC-MT has the lowest loss at every budget. On Taskmaster-3 it is 16.9 to 7.4\,pp below the strongest baseline, ElastiCache-MT, from $k=100$ (77.1\%) to
$k=1000$ (72.6\%); with 100 entries it already matches unbounded ElastiCache-MT (77.3\%), which caches all 19,932 source states. On SocraTeach, the budgeted baselines stay near no caching (96.2--101.0\%), while CUCB-SC-MT falls from 94.3\% to 69.1\%. Without a budget it is 6.2\,pp (CI 2.5--9.8) and 11.9\,pp (CI 2.7--20.9) below the best unbounded baseline, but within 2\,pp of its own loss at $k=1000$. Single-turn, which ignores history, stays at 96--101\%.


\paragraph{Online.}
CLCB-LS-MT has the lowest loss at every budget on both workloads. On Taskmaster-3 its loss barely depends on $k$ (73.0--75.0\%; 74.1\% without a budget), while ElastiCache-MT needs a large cache (99.0\% at $k=100$, 81.9\% unbounded): the gap is 24.0 to 11.4\,pp at matched budgets and 7.8\,pp (CI 6.5--9.1) without a budget. On SocraTeach only ContextCache is competitive; CLCB-LS-MT is 2.4\,pp (CI 0.5--4.1) below it at $k=100$, 9.2\,pp (CI 7.1--11.2)
at $k=1000$ and 7.5\,pp (CI 5.4--9.6) without a budget. There, unbounded ElastiCache-MT, the joint-state threshold and vCache are worse than no caching.

\subsection{Learning over the stream}
\label{sec:exp-running}


Figure~\ref{fig:online_running_loss} plots the running loss $100\sum_{s\le t}\ell_s/\sum_{s\le t}c(z_s)$. Fixed-threshold baselines flatten once their caches fill, while CLCB-LS-MT keeps learning from feedback: on Taskmaster-3 it falls from 80.3\% in the first quarter of the stream to 72.0\% ($k=200$) and 71.7\% ($k=1000$) in the last, while the budgeted baselines' quarterly losses move by at most 2.6\,pp. On SocraTeach, it briefly exceeds 100\% while its risk model has few labels, falls below ContextCache within 2,500--4,000 requests, and improves by 7.5--15.8\,pp from the first to the last quarter (ContextCache: at most 4.1\,pp; Appendix~\ref{app:diagnostics}). As Theorem~\ref{thm:online} predicts, the loss per request falls as the confidence widths shrink, and the early excess on SocraTeach is the price of optimism, which reuses uncertain pairs to learn their risk.

\subsection{Reuse quality and generation cost saved}
\label{sec:exp-reuse}
Table~\ref{tab:reuse-key} separates the avoided cost from the penalties via the \emph{safe cost avoided} (SCA), $\sum_{t:\,\text{hit},\,Y_t=0}c(z_t)/\sum_t c(z_t)$.
At $k=1000$, CLCB-LS-MT safely avoids 35.0\% (Taskmaster-3) and 41.0\% (SocraTeach) of the no-cache cost, against at most 22.3\% and 25.1\% for the
baselines.
Since CLCB-LS-MT weighs cost against risk as in ~\eqref{eq:decision}, it reuses on requests that cost about 1.5 times the average, against 1.1--1.3 times for the strongest baselines. Its precision is lower and its false reuse higher (8.2\% and 13.1\% of requests), each charged as much as the most expensive request, yet its loss is lowest. 

\subsection{Latency}
\label{sec:latency}

We estimate the mean latency of method $a$ at budget $k$ as
$t_{\mathrm{emb},a}+t_{\mathrm{dec},a}(k)+(1-h_{a,k})\,t_{\mathrm{LLM}}+t_{\mathrm{bg},a}(k)$: embedding time (GPU) and decision time (CPU), averaged over 300 requests, plus the LLM time $t_{\mathrm{LLM}}$ (mean of 40 sequential GPT-4o-mini calls) on misses, where $h_{a,k}$ is the hit rate. Online, $t_{\mathrm{bg},a}(k)$ amortizes the updates made after each response; judge calls for learning are off the request path and excluded (Appendix~\ref{app:latency}).



Online (Table~\ref{tab:online-latency-main}), CLCB-LS-MT lowers the estimated latency by 15--20\% on Taskmaster-3 and 30--35\% on SocraTeach at $k=100$--$1000$ (19\% and 41\% without a budget), including 23--28\,ms of background work per request. Its hit path takes 39--83\,ms, under a tenth of an LLM call. The hit rate dominates: vCache, the joint-state threshold and single-turn rarely reuse and
stay at 0.90--1.08, while aggressive reuse can be fast but costly (unbounded ElastiCache-MT on SocraTeach: 0.10, at a loss of 154.7\%). Offline, CUCB-SC-MT reaches 0.81--0.84 on Taskmaster-3 and 0.61--0.99 on SocraTeach. These
estimates ignore queueing and count every reuse as a fast hit, so they should be read with the loss.


\section{Conclusion}

In multi-turn conversations, whether a cached response can be reused depends on a pair of conversation states, which embedding similarity predicts poorly. We learn this reuse risk from judge feedback and weigh it against the generation cost a reuse saves. CUCB-SC-MT and CLCB-LS-MT have suboptimality and regret guarantees despite correlated turns, and online with 1,000 entries CLCB-LS-MT avoids 35--41\% of the generation cost through acceptable reuses, against at most 22--25\% for the baselines. Natural next steps are to let reuses change how conversations continue (Assumption~\ref{ass:exogenous}) and to extend the guarantees to reuse within a conversation. The appendix contains all proofs (Appendices~\ref{app:oracle}--\ref{app:online}) and further results (Appendix~\ref{app:experiments}): the motivation study, per-budget latency and reuse quality, the full loss tables, the online loss over the stream and a sweep over the price $\gamma$.

\bibliography{example_paper}
\bibliographystyle{mlsys2025}

\clearpage
\appendix

\begin{table*}[!t]
\centering
\caption{Notation.}
\label{tab:notation}
\footnotesize
\renewcommand{\arraystretch}{1.1}
\begin{minipage}[t]{0.485\textwidth}
\begin{tabular}[t]{@{}l >{\raggedright\arraybackslash}p{0.60\linewidth}@{}}
\toprule
Symbol & Meaning \\
\midrule
\multicolumn{2}{@{}l}{\textit{System model}} \\
$j,\ t,\ T_j\le T_{\max}$ & conversation, turn, conversation length \\
$q_{j,t},\ H_{j,t},\ a_{j,t}$ & request, history, returned response \\
$z=(\omega,x)\in\mathcal Z$ & conversation state: exact-match key $\omega\in\Omega$, embedding $x\in\mathcal X_\omega\subset\mathbb R^{d_e}$ \\
$\mathcal Z_\omega,\ d(z,z')$ & stratum of key $\omega$; distance within a stratum, $+\infty$ across strata \\
$P,\ P_\omega$ & per-request law~\eqref{eq:perrequest}, law of $x$ given $\omega$ \\
$p,\ c_0,\ \bar r$ & intrinsic dimension, lower-mass constant, radius up to which the lower-mass bound holds \\
$c=c_{\mathrm{in}}+c_{\mathrm{out}}$ & expected serving cost: known input part, unknown generation part \\
$C_{j,t},\ \eta_{j,t},\ R_c$ & observed cost, its noise, sub-Gaussian parameter \\
$(u,a(u))$ & cache item: source state $u\in\mathcal Z$ and the response stored with it \\
$\mathcal M,\ k$ & cache, cache size \\
$\chi=\bar\chi(\omega_z,\omega_u)$ & known compatibility mask \\
$r(z,u),\ \psi(z,u),\ Y$ & reuse risk, pair features, risk label \\
$\gamma,\ \rho(z,u)$ & risk price, reuse loss~\eqref{eq:reuseloss} \\
$\rho(z,\mathcal M),\ s(z,\mathcal M)$ & smallest reuse loss over the cache, item attaining it \\
$g_{\mathcal M},\ \mathcal L(\mathcal M)$ & per-state loss, expected loss~\eqref{eq:loss} \\
$\mathcal U,\ \mathcal M^*$ & cacheable source states, optimal cache~\eqref{eq:design} \\
$\mathcal S,\ z_i,\ V_i,\ w_i,\ m,\ \epsilon$ & stratified $\epsilon$-net, representatives, cells, weights, size, radius \\
$c_i,\ R_{ij},\ \widehat{\mathcal L}$ & costs and reuse losses at the representatives, discretized loss~\eqref{eq:discloss} \\
\midrule
\multicolumn{2}{@{}l}{\textit{Oracle}} \\
$\alpha,\ s$ & Reverse Greedy approximation factor, curvature \\
$L_c,\ L_r,\ L_{\mathrm{src}}$ & Lipschitz constants of $c$, of $r$ in the target and in the source \\
$L_g$ & $\max\{L_c,\gamma L_r\}$, Lipschitz constant of $g_{\mathcal M}$ within a key \\
\bottomrule
\end{tabular}
\end{minipage}\hfill
\begin{minipage}[t]{0.485\textwidth}
\begin{tabular}[t]{@{}l >{\raggedright\arraybackslash}p{0.60\linewidth}@{}}
\toprule
Symbol & Meaning \\
\midrule
\multicolumn{2}{@{}l}{\textit{Offline learning}} \\
$\mathcal D_1,\ \mathcal D_2$ & conversations that build the net and that estimate the weights \\
$J_2,\ N_2,\ N_{\mathrm{eff}}$ & conversations and turns in $\mathcal D_2$, effective sample size $(\sum_jT_j)^2/\sum_jT_j^2$ \\
$n_c,\ n_r$ & numbers of cost observations and labeled pairs \\
$k_c,\ k_r,\ \ell_c,\ \ell_r,\ \lambda$ & kernels on $x$ and on $\psi$, RBF length scales, ridge \\
$B_c,\ B_r,\ \varepsilon_0$ & RKHS norm bounds, risk misspecification \\
$\hat c_{\mathrm{out}},\ \hat r,\ \sigma_c,\ \sigma_r$ & KRR estimates and widths for generation cost and reuse risk \\
$\mathcal I_c,\ \mathcal I_r$ & maximum information gains of $k_c$, $k_r$ \\
$\beta_c,\ \beta_r,\ \varepsilon_{\mathrm{app}}$ & confidence multipliers, misspecification width \\
$\bar c_i,\ \bar R_{ij}$ & pessimistic (upper) cost and reuse-loss estimates \\
$\mathcal L^{\mathrm{dep}},\ \mathrm{SubOpt}$ & loss of the deployed rule, suboptimality~\eqref{eq:subopt} \\
$\mathcal M^\circ,\ I_L,\ I_R,\ j^*(i)$ & $\mathcal M^*$ projected onto the net, cells where it invokes the LLM or reuses, item it reuses \\
$\nu_{\min},\ C_R,\ p_r$ & LLM propensity, label-density constant, intrinsic dimension of $\psi$ (coverage) \\
$W_R,\ \bar\varepsilon_{\mathrm{app}}$ & comparator's reuse mass, largest misspecification width on its pairs \\
\midrule
\multicolumn{2}{@{}l}{\textit{Online learning}} \\
$n,\ N,\ z_n$ & round, horizon, state at round $n$ \\
$\bar T,\ \mathcal G_j$ & $\mathbb E[T^2]/\mathbb E[T]$; $\sigma$-field of the first $j$ conversations \\
$\ell_n,\ \mathrm{Reg}(N)$ & loss served at round $n$, regret~\eqref{eq:regret} \\
$\zeta,\ L_\psi$ & label rate, Lipschitz constant of $\psi$ in $x_z$ \\
$\underline c_i,\ \underline R_{ij},\ \Pi_j$ & optimistic (lower) estimates, snapshot used in conversation $j$ \\
$m_{\max},\ G_s$ & largest net size; $\ln\det(I+K_s/\lambda)$ of the data before round $s$ \\
$\Lambda,\ t_0,\ t_1$ & logarithmic factor and burn-in thresholds (Lemma~\ref{lem:arr-online}, Theorem~\ref{thm:online}) \\
$N_{\mathrm{arr}},\ N_c,\ N_r,\ N_{\mathrm{sw}}$ & arrival stages, cost and label information triggers, all switches \\
$Q_N,\ \epsilon^{\mathrm{proj}}_n$ & quantization of served requests, projection distance of $\mathcal M^*$ \\
$L_{\mathrm{disc}},\ E_{\mathrm{mis}},\ R_{\mathrm{low}}$ & discretization constant, misspecification term, lower-order remainder \\
$\tau_{\mathrm{cov}},\ \vartheta_u$ & coverage burn-in, mass of the $\epsilon/2$-ball around $u$ \\
\bottomrule
\end{tabular}
\end{minipage}
\end{table*}

\section{Proofs for Section~\ref{sec:oracle}}\label{app:oracle}

Table~\ref{tab:notation} collects the notation used in the proofs. For
$\mathcal M\subseteq[m]$ we write
$h_i(\mathcal M;\boldsymbol c,\boldsymbol R):=\min\{c_i,\min_{j\in\mathcal M}R_{ij}\}$,
so that
$\widehat{\mathcal L}(\mathcal M;\mathcal S,\boldsymbol w,\boldsymbol c,\boldsymbol R)=\sum_i w_ih_i(\mathcal M;\boldsymbol c,\boldsymbol R)$. 

Setting $T\equiv1$, $|\Omega|=1$, $\chi\equiv1$, $\gamma=1$, $c_{\mathrm{in}}\equiv0$ and $r(z,u)=\phi(d(z,u))$ for the known mismatch
function $\phi$ of~\citet{atalar2026continuoussemanticcachinglowcost} gives $\rho=\phi\circ d$, makes $s(z,\mathcal M)$ the nearest cached query, and reduces~\eqref{eq:decision}--\eqref{eq:discloss} to Eqs.~(1)--(4) of~\citet{atalar2026continuoussemanticcachinglowcost}.
 
\begin{lemma}[NP-hardness]\label{lem:nphard}
The multi-turn semantic caching problem~\eqref{eq:design} is NP-hard.
\end{lemma}
\begin{proof}
Problem~\eqref{eq:design} contains the single-turn problem
of~\cite{atalar2026continuoussemanticcachinglowcost} as the special case
described above, which is NP-hard by
Lemma~3.1 there.
\end{proof}
 
\begin{algorithm}[h]
\caption{Reverse Greedy~\cite{liu2026semanticcachinglowcostllm}}
\label{alg:rg}
\begin{algorithmic}[1]
\REQUIRE representatives $\mathcal S$ with $|\mathcal S|=m$, weights
  $\boldsymbol w$, costs $\boldsymbol c$, reuse losses $\boldsymbol R$,
  cache size $k$
\STATE $\mathcal M\gets[m]$
\WHILE{$|\mathcal M|>k$}
\STATE $q\gets\arg\min_{q'\in\mathcal M}\widehat{\mathcal L}(\mathcal M\setminus\{q'\};\mathcal S,\boldsymbol w,\boldsymbol c,\boldsymbol R)$
\STATE $\mathcal M\gets\mathcal M\setminus\{q\}$
\ENDWHILE
\ENSURE cache $\mathcal M$
\end{algorithmic}
\end{algorithm}
 
Lemma~\ref{lem:supermod} applies Lemma~2 and Theorem~1
of~\cite{liu2026semanticcachinglowcostllm} with reuse costs $R_{ij}$; their
proofs treat the reuse cost of each pair as an arbitrary non-negative
number. The continuous loss and its discretization are related as follows.
 
\begin{lemma}[Continuous-to-discrete approximation]\label{lem:disc}
Let $\mathcal S$ be a stratified $\epsilon$-net and suppose
Assumption~\ref{ass:lipschitz} holds. For every cache $\mathcal M$ and key
$\omega$, $x\mapsto g_{\mathcal M}(\omega,x)$ is $L_g$-Lipschitz with
$L_g:=\max\{L_c,\gamma L_r\}$, and
\[
\Big|\mathcal L(\mathcal M)-\sum_{i=1}^m w_i\,g_{\mathcal M}(z_i)\Big|\le L_g\epsilon .
\]
For $\mathcal M\subseteq[m]$, the sum equals
$\widehat{\mathcal L}(\mathcal M;\mathcal S,\boldsymbol w,\boldsymbol c,\boldsymbol R)$.
\end{lemma}
 
The third condition of Assumption~\ref{ass:lipschitz} holds automatically
in the single-turn model, where $\phi(d(x,u))$ is Lipschitz in $u$. In that
model $\gamma=1$ and $L_{\mathrm{src}}=L_\phi\le L_g$, so
Theorem~\ref{thm:oracle} gives $(1+2\alpha)L_g\epsilon$. The single-turn
analysis uses $m=O(\epsilon^{-d_e})$, which grows with the embedding
dimension; Lemma~\ref{lem:netsize} replaces $d_e$ by $p$ because each key
is covered separately.

\subsection{Proof of Lemma~\ref{lem:disc}}
The main difference relative to Lemma~3.2
of~\cite{atalar2026continuoussemanticcachinglowcost} is that
$g_{\mathcal M}$ is Lipschitz only within a key. Fix a key $\omega$ and a
cache $\mathcal M$. The map $x\mapsto c(\omega,x)$ is $L_c$-Lipschitz by
Assumption~\ref{ass:lipschitz}. For $u\in\mathcal M$, the mask
$\chi((\omega,x),u)=\bar\chi(\omega,\omega_u)$ does not depend on $x$: if it
is $0$, then $\rho((\omega,\cdot),u)\equiv1$; if it is $1$, then
$\rho((\omega,\cdot),u)=\min\{1,\gamma r((\omega,\cdot),u)\}$ is
$\gamma L_r$-Lipschitz, since $t\mapsto\min\{1,\gamma t\}$ is
$\gamma$-Lipschitz. A minimum of finitely many $L$-Lipschitz functions is
$L$-Lipschitz, so $g_{\mathcal M}(\omega,\cdot)$ is $L_g$-Lipschitz.
 
The approximation bound then follows from the proof of Lemma~3.2
of~\cite{atalar2026continuoussemanticcachinglowcost}, applied within each
key: since the net covers $\operatorname{supp}P$ and $d=\infty$ across
keys, every $z\in V_i\cap\operatorname{supp}P$ has the key of $z_i$ and
$\|x-x_i\|_2\le\epsilon$. Finally, for $\mathcal M\subseteq[m]$,
$g_{\mathcal M}(z_i)=h_i(\mathcal M;\boldsymbol c,\boldsymbol R)$.
\hfill$\square$
 
\subsection{Proof of Theorem~\ref{thm:oracle}}\label{app:proj}
 
The only new step relative to Theorem~3.3
of~\cite{atalar2026continuoussemanticcachinglowcost} is the projection of
the optimum, where the mismatch function is replaced by the masked reuse
risk. Every $u\in\mathcal M^*\subseteq\mathcal U\subseteq\operatorname{supp}P$
has a representative $\pi(u)\in\mathcal S$ with the same key and
$\|x_u-x_{\pi(u)}\|_2\le\epsilon$; let
$\mathcal M^\circ:=\{\pi(u):u\in\mathcal M^*\}$, so $|\mathcal M^\circ|\le k$.
For every state $z$, $\chi(z,u)=\chi(z,\pi(u))$ because the keys of $u$ and
$\pi(u)$ coincide. If both masks are $0$, both reuse losses are $1$;
otherwise the third condition of Assumption~\ref{ass:lipschitz} gives
$\rho(z,\pi(u))\le\rho(z,u)+\gamma L_{\mathrm{src}}\epsilon$. Taking the
minimum over $u\in\mathcal M^*$ yields
$g_{\mathcal M^\circ}\le g_{\mathcal M^*}+\gamma L_{\mathrm{src}}\epsilon$
pointwise, hence
\begin{equation}\label{eq:proj}
\mathcal L(\mathcal M^\circ)\le\mathcal L(\mathcal M^*)+\gamma L_{\mathrm{src}}\epsilon .
\end{equation}
The remainder is the chain in the proof of Theorem~3.3
of~\cite{atalar2026continuoussemanticcachinglowcost}, with
Lemma~\ref{lem:disc} in place of Lemma~3.2 there and~\eqref{eq:proj} in
place of its projection step:
$\mathcal L(\mathcal M)\le\alpha\widehat{\mathcal L}(\mathcal M^\circ)+L_g\epsilon\le\alpha\mathcal L(\mathcal M^\circ)+(1+\alpha)L_g\epsilon$.
\hfill$\square$
 
\subsection{Proof of Lemma~\ref{lem:netsize}}
 
Fix a key $\omega$ and let $x_1,\dots,x_{m_\omega}$ be the embeddings of
the representatives with key $\omega$. They are pairwise more than
$\epsilon$ apart, so the balls of radius $\epsilon/2$ around them are
disjoint. Each $x_i$ lies in $\operatorname{supp}P_\omega$ and
$\epsilon/2\le\bar r$, so by the lower-mass condition of
Section~\ref{sec:model} each ball has conditional mass at least
$c_0(\epsilon/2)^p$. The masses of disjoint sets sum to at most $1$, so
$m_\omega\le2^p/(c_0\epsilon^p)$. Representatives lie in
$\operatorname{supp}P$, so their keys belong to $\Omega$, and summing over
$\omega\in\Omega$ gives the claim. The representatives built by
Algorithm~\ref{alg:offline} satisfy the separation condition by
construction.
\hfill$\square$

\section{Proofs for Section~\ref{sec:offline}}\label{app:offline}
 
We write
$F(\mathcal M;\boldsymbol w,\boldsymbol c,\boldsymbol R):=\widehat{\mathcal L}(\mathcal M;\mathcal S,\boldsymbol w,\boldsymbol c,\boldsymbol R)$,
and $c_i=c(z_i)$, $R_{ij}=\rho(z_i,z_j)$ for the true values at the
representatives. We use Lemma~A.2
of~\cite{atalar2026continuoussemanticcachinglowcost} (regularized
posterior width via local counts) and the variational characterization
in its proof: the regularized variance at $v$ equals
$\min_{\boldsymbol a}\{\|k(v,\cdot)-\sum_sa_sk(v_s,\cdot)\|_{\mathcal H}^2+\lambda\|\boldsymbol a\|_2^2\}$,
attained at $\boldsymbol a^*=(K+\lambda I)^{-1}\kappa_v$. Both hold for an
RBF kernel on any feature map, which we apply to $x$ for the cost and to
$\psi$ for the risk.

\paragraph{Design choices of Algorithm~\ref{alg:offline}.}
Splitting turns rather than conversations between $\mathcal D_1$ and
$\mathcal D_2$ would leave dependent samples on both sides. The input cost
$c_{\mathrm{in}}$, which grows with the history, enters $\bar c$ exactly
(Line~3). The confidence widths are self-normalized with a fixed ridge
$\lambda$ because later states contain earlier responses, so regression
inputs depend on earlier noise and fixed-design bounds do not apply.
 
\begin{lemma}[Concentration of estimates]\label{lem:conc}
Let $\mathcal S$ be built on $\mathcal D_1$ and cover
$\operatorname{supp}P$, and let
$N_{\mathrm{eff}}:=\big(\sum_{j\in\mathcal D_2}T_j\big)^2/\sum_{j\in\mathcal D_2}T_j^2$.
Under Assumptions~\ref{ass:logs}--\ref{ass:kernel}, each of the following
events holds with probability at least $1-\delta/3$:
\[
\begin{aligned}
\mathcal E_{\mathrm{arr}}&:=\Big\{\|\hat{\boldsymbol w}-\boldsymbol w\|_1\le\sqrt{\tfrac{2(m\ln2+\ln(3/\delta))}{N_{\mathrm{eff}}}}\Big\},\\
\mathcal E_{c}&:=\big\{0\le\bar c(z)-c(z)\le2\beta_c\sigma_c(z)\big\},\\
\mathcal E_{r}&:=\big\{0\le\bar R_{ij}-R_{ij}\le2\gamma(\beta_r\sigma_r+\varepsilon_{\mathrm{app}})(z_i,z_j)\big\},
\end{aligned}
\]
where $\mathcal E_c$ holds for all $z\in\mathcal Z$, $\mathcal E_r$ for all
$i,j\in[m]$, and $\bar R_{ij}=R_{ij}=1$ for incompatible pairs.
\end{lemma}
 
$\mathcal E_{\mathrm{arr}}$ counts conversations, not turns: by
Cauchy--Schwarz, $N_{\mathrm{eff}}\le J_2$, with equality iff all
conversations have the same length. This cannot be improved under
arbitrary within-conversation dependence, and for $T_j\equiv1$ it recovers
the arrival event of~\cite{atalar2026continuoussemanticcachinglowcost}.
$\mathcal E_c$ and $\mathcal E_r$ hold uniformly over the space, so they
can be evaluated at the data-dependent representatives.

\subsection{A Conversation-Level Chernoff Bound}
 
Local counts in the single-turn analysis rely on a binomial Chernoff
bound for independent samples. Turns within a conversation are
dependent and logging may be adaptive, so we use the following martingale
version over conversations.
 
\begin{lemma}[Conversation-level Chernoff bound]\label{lem:chernoff}
Let $\mathcal G_0\subseteq\mathcal G_1\subseteq\cdots$ be a filtration and
$0\le B_j\le T_j\le T_{\max}$ with $B_j$ $\mathcal G_j$-measurable. If
$\sum_j\mathbb E[B_j\mid\mathcal G_{j-1}]\ge\mu$ almost surely, then
\[
\Pr\Big(\sum_jB_j\le\mu/2\Big)\le\exp\Big(-\frac{\mu}{8T_{\max}}\Big).
\]
\end{lemma}
 
\begin{proof}
Let $Y_j=B_j/T_{\max}\in[0,1]$ and $q_j=\mathbb E[Y_j\mid\mathcal G_{j-1}]$.
By convexity, $e^{-\theta y}\le1-(1-e^{-\theta})y$ on $[0,1]$, so
$\mathbb E[e^{-\theta Y_j}\mid\mathcal G_{j-1}]\le\exp(-(1-e^{-\theta})q_j)$,
and $M:=\exp\big(-\theta\sum_jY_j+(1-e^{-\theta})\sum_jq_j\big)$ satisfies
$\mathbb EM\le1$. Take $\theta=\ln2$. On $\{\sum_jB_j\le\mu/2\}$,
$\sum_jY_j\le\frac12\sum_jq_j$ and $\sum_jq_j\ge\mu/T_{\max}$, so
$M\ge\exp\big(\frac{1-\ln2}{2}\mu/T_{\max}\big)$. Markov's inequality and
$\frac{1-\ln2}{2}\ge\frac18$ give the claim.
\end{proof}

With $T_{\max}=1$ this reduces to the Chernoff step behind the local-count
event of~\cite{atalar2026continuoussemanticcachinglowcost}.
 
\subsection{Proof of Lemma~\ref{lem:conc}}
 
\emph{Arrival event.} Condition on $\mathcal D_1$, so the cells are fixed,
and follow the proof of the arrival event in Lemma~4.2
of~\cite{atalar2026continuoussemanticcachinglowcost}: with
$W(A):=\sum_{i\in A}w_i$ and $\widehat W(A):=\sum_{i\in A}\hat w_i$,
$\|\hat{\boldsymbol w}-\boldsymbol w\|_1\ge\eta$ iff
$\widehat W(A)-W(A)\ge\eta/2$ for some $A\subseteq[m]$. The only change is
that turns are not independent. Write
$N_2\widehat W(A)=\sum_{j\in\mathcal D_2}S_j(A)$ with
$S_j(A):=\sum_{t=1}^{T_j}\mathbf 1\{Z_{j,t}\in\cup_{i\in A}V_i\}\in[0,T_j]$.
By Assumption~\ref{ass:logs}, conditionally on the lengths, the $S_j(A)$
are independent with mean $T_jW(A)$, so Hoeffding's inequality gives
$\Pr(\widehat W(A)-W(A)\ge\eta/2)\le\exp(-N_2^2\eta^2/(2\sum_jT_j^2))=\exp(-N_{\mathrm{eff}}\eta^2/2)$.
A union bound over the $2^m$ subsets and
$\eta=\sqrt{2(m\ln2+\ln(3/\delta))/N_{\mathrm{eff}}}$ give
$\mathcal E_{\mathrm{arr}}$.

\begin{remark}[Random lengths]\label{rem:randomlen}
If length-invariance is dropped and conversations are i.i.d.\ with random
lengths, as in Section~\ref{sec:model}, then
$\mathbb E[S_j(A)-T_jW(A)]=0$ by~\eqref{eq:perrequest}, and the summands
are independent with ranges $2T_{\max}$. The same argument gives
$\mathcal E_{\mathrm{arr}}$ with $N_{\mathrm{eff}}$ replaced by
$N_2^2/(4J_2T_{\max}^2)$.
\end{remark}
 
\emph{Cost event.} As in the proof of Lemma~5.2
of~\cite{atalar2026continuoussemanticcachinglowcost}, this is Theorem~2
of~\cite{chowdhury2017kernelizedmultiarmedbandits}, applied to
$y_s:=C_s-c_{\mathrm{in}}(z_s)=c_{\mathrm{out}}(x_s)+\eta_s$ with the cost
observations ordered conversation by conversation and turn by turn. Its
hypotheses hold because the decision to invoke the LLM is made before
the cost is revealed, so each input is predictable, and by the noise
model of Section~\ref{sec:model} each $\eta_s$ is conditionally
$R_c$-sub-Gaussian given the past, even though later states contain
earlier responses. The bound is uniform over the space, so it holds at the
data-dependent representatives.
Since $c_{\mathrm{in}}$ is exact and $c\le1$,
$c\le\bar c$ and $\bar c-c\le2\beta_c\sigma_c$.

\emph{Risk event.} Let $q=(z,u)$ be a compatible pair and
$\boldsymbol h_q:=(K_r+\lambda I)^{-1}\kappa_q$, so
$\hat r(q)=\boldsymbol h_q^\top\boldsymbol Y$. Write
$Y_s=r_0(q_s)+e(q_s)+\eta^r_s$ with $\eta^r_s:=Y_s-r(q_s)$, a martingale
difference by Assumption~\ref{ass:kernel}(c) that is conditionally
$\frac12$-sub-Gaussian because $Y_s\in[0,1]$. Then
$\hat r(q)-r(q)=[\boldsymbol h_q^\top(\boldsymbol r_0+\boldsymbol\eta^r)-r_0(q)]+[\boldsymbol h_q^\top\boldsymbol e-e(q)]$.
The first bracket is at most $\beta_r\sigma_r(q)$ by Theorem~2
of~\cite{chowdhury2017kernelizedmultiarmedbandits}, uniformly over $q$. For the second,
$|\boldsymbol h_q^\top\boldsymbol e|\le\sqrt{n_r}\,\|\boldsymbol h_q\|_2\,\varepsilon_0$,
and the variational characterization at its minimizer $\boldsymbol h_q$
gives $\sigma_r^2(q)\ge\lambda\|\boldsymbol h_q\|_2^2$; with
$|e(q)|\le\varepsilon_0$ this yields
$|\hat r(q)-r(q)|\le\beta_r\sigma_r(q)+\varepsilon_{\mathrm{app}}(q)$.
Since $t\mapsto\min\{1,\gamma t\}$ is monotone and $\gamma$-Lipschitz,
$0\le\bar R_{ij}-R_{ij}\le2\gamma(\beta_r\sigma_r+\varepsilon_{\mathrm{app}})$
at $q=(z_i,z_j)$, and incompatible pairs have $\bar R_{ij}=R_{ij}=1$.
\hfill$\square$
 
\subsection{Proof of Theorem~\ref{thm:offline}}\label{app:thm-offline}
 
The deployed rule serves $z\in V_i$ from
$j^\circ=\arg\min_{j\in\widehat{\mathcal M}}\bar R_{ij}$ if
$\bar R_{ij^\circ}<\bar c_i$ and invokes the LLM otherwise.
Work on $\mathcal E:=\mathcal E_{\mathrm{arr}}\cap\mathcal E_c\cap\mathcal E_r$,
which has probability at least $1-\delta$; on $\mathcal E$, all entries of
$\bar{\boldsymbol c}$ and $\bar{\boldsymbol R}$ lie in $[0,1]$.
 
(i) $\mathcal L(\widehat{\mathcal M})\le\mathcal L^{\mathrm{dep}}(\widehat{\mathcal M})$,
because oracle routing is pointwise optimal among rules that use
$\widehat{\mathcal M}$.
 
(ii) $\mathcal L^{\mathrm{dep}}(\widehat{\mathcal M})\le F(\widehat{\mathcal M};\boldsymbol w,\bar{\boldsymbol c},\bar{\boldsymbol R})+L_g\epsilon$.
Let $z\in V_i\cap\operatorname{supp}P$, so $z$ has the key of $z_i$ and
$\|x-x_i\|_2\le\epsilon$. If the rule invokes the LLM, its loss is
$c(z)\le c_i+L_c\epsilon\le\bar c_i+L_g\epsilon$; if it reuses $z_j$, the
mask is the same at $z$ and $z_i$, so
$\rho(z,z_j)\le R_{ij}+\gamma L_r\epsilon\le\bar R_{ij}+L_g\epsilon$. In
both cases the pessimistic value of the chosen option is
$h_i(\widehat{\mathcal M};\bar{\boldsymbol c},\bar{\boldsymbol R})$.
Integrate over the cells.
 
(iii) By Lemma~4 of~\cite{liu2026semanticcachinglowcostllm},
$F(\widehat{\mathcal M};\boldsymbol w,\bar{\boldsymbol c},\bar{\boldsymbol R})\le F(\widehat{\mathcal M};\hat{\boldsymbol w},\bar{\boldsymbol c},\bar{\boldsymbol R})+\frac12\|\hat{\boldsymbol w}-\boldsymbol w\|_1$,
where the constant $\frac12$ instead of $1$ follows from
$\sum_i(w_i-\hat w_i)h_i=\sum_i(w_i-\hat w_i)(h_i-\frac12)$ and
$h_i\in[0,1]$.
 
(iv) By Lemma~\ref{lem:supermod} on the instance
$(\hat{\boldsymbol w},\bar{\boldsymbol c},\bar{\boldsymbol R})$, with
$\mathcal M^\circ$ feasible,
$F(\widehat{\mathcal M};\hat{\boldsymbol w},\bar{\boldsymbol c},\bar{\boldsymbol R})\le\alpha F(\mathcal M^\circ;\hat{\boldsymbol w},\bar{\boldsymbol c},\bar{\boldsymbol R})$.
 
(v) By (iii) again, and because
$h_i(\mathcal M^\circ;\bar{\boldsymbol c},\bar{\boldsymbol R})$ is at most
the pessimistic value of the option $a_i$ that attains
$h_i(\mathcal M^\circ;\boldsymbol c,\boldsymbol R)$ (as in Lemma~6
of~\cite{liu2026semanticcachinglowcostllm}),
\[
\begin{aligned}
F(\mathcal M^\circ;\hat{\boldsymbol w},\bar{\boldsymbol c},\bar{\boldsymbol R})
\le\;&F(\mathcal M^\circ;\boldsymbol w,\boldsymbol c,\boldsymbol R)+\tfrac12\|\hat{\boldsymbol w}-\boldsymbol w\|_1\\
&+\sum_{i\in I_L}w_i(\bar c_i-c_i)\\
&+\sum_{i\in I_R}w_i\big(\bar R_{ij^*(i)}-R_{ij^*(i)}\big).
\end{aligned}
\]
 
(vi) By Lemma~\ref{lem:disc} and~\eqref{eq:proj},
$F(\mathcal M^\circ;\boldsymbol w,\boldsymbol c,\boldsymbol R)\le\mathcal L(\mathcal M^*)+\gamma L_{\mathrm{src}}\epsilon+L_g\epsilon$.
 
Chaining (ii)--(vi), subtracting $\alpha\mathcal L(\mathcal M^*)$ and
substituting $\mathcal E_{\mathrm{arr}}$, $\mathcal E_c$ and $\mathcal E_r$
gives the theorem. Each $j^*(i)$ is compatible, since an incompatible
item has reuse loss $1\ge c_i$ and ties go to the LLM.
\hfill$\square$

\subsection{Coverage Conditions and Proof of Corollary~\ref{cor:offline}}\label{app:explicit}
 
Order the conversations of $\mathcal D_2$ arbitrarily, let $\mathcal G_j$
be generated by $\mathcal D_1$ and the first $j$ of them, and condition on
the lengths. The quantities $\mathcal S$, $I_L$, $I_R$ and $j^*$ depend
only on $\mathcal D_1$ and the true parameters, so they are
$\mathcal G_0$-measurable. For $i\in I_L$ and $r>0$, let $B_{j,i}(r)$ be
the number of turns of conversation $j$ at which the LLM was invoked, the
key equals $\omega_i$ and $\|x_{j,t}-x_i\|_2\le r$. Write
$P(\omega):=P(\omega_Z=\omega)$.
 
\textbf{(C1) Cost coverage.} For every $i\in I_L$ and $r\le\bar r$,
\[
\begin{aligned}
&\textstyle\sum_{j\in\mathcal D_2}\mathbb E\big[B_{j,i}(r)\mid\mathcal G_{j-1}\big]\\
&\qquad\ge\nu_iN_2\,P\big(\omega_Z=\omega_i,\ \|X-x_i\|_2\le r\big),
\end{aligned}
\]
with $\nu_i\ge\nu_{\min}>0$: the logging policy invoked the LLM on at
least a $\nu_i$ fraction of the $\mathcal D_2$ turns near $z_i$.
 
\textbf{(C2) Risk coverage.} For every $i\in I_R$, $j=j^*(i)$ and
$r\in(0,r_0]$, at least $\nu_{ij}n_r(r/r_0)^{p_r}$ labeled pairs satisfy
$\|\psi(z_s,u_s)-\psi(z_i,z_j)\|_2\le r$, and
$\nu_{ij}n_r\ge\lambda\ell_r^2/r_0^2$. Set
$C_R:=\sum_{i\in I_R}w_i/\nu_{ij^*(i)}$.
 
\textbf{Burn-in.}
\[
N_0:=\max_{i\in I_L}\frac{\max\{A,B\}}{c_0\nu_iP(\omega_i)},
\]
where $A:=\big(8T_{\max}\ln\tfrac{4m}{\delta}\big)^{(p+2)/2}(2\lambda\ell_c^2)^{-p/2}$
and $B:=2\lambda\ell_c^2\,\bar r^{\,-(p+2)}$.

Under (C1), (C2) and $N_2\ge N_0$, the explicit form of
Corollary~\ref{cor:offline} is: with probability at least $1-\delta$,
\[
\begin{aligned}
\mathrm{SubOpt}(\widehat{\mathcal M})\le\;&(1+\alpha)L_g\epsilon+\alpha\gamma L_{\mathrm{src}}\epsilon\\
&+\tfrac{1+\alpha}{2}\sqrt{\tfrac{2(m\ln2+\ln(4/\delta))}{N_{\mathrm{eff}}}}\\
&+2\sqrt2\,\alpha\beta_c\Big(\tfrac{2\lambda|\Omega|}{c_0\ell_c^{p}\nu_{\min}N_2}\Big)^{\frac{1}{p+2}}\\
&+2\sqrt2\alpha\gamma\beta_rW_R^{\frac{p_r+1}{p_r+2}}\big(\tfrac{\lambda(r_0/\ell_r)^{p_r}C_R}{n_r}\big)^{\frac{1}{p_r+2}}\\
&+2\alpha\gamma\,\bar\varepsilon_{\mathrm{app}}W_R.
\end{aligned}
\]
With Lemma~\ref{lem:netsize} and
$\epsilon^*\asymp(|\Omega|/(c_0N_{\mathrm{eff}}))^{1/(p+2)}$ this gives the
rate in Corollary~\ref{cor:offline}.

\begin{proof}[Proof of Corollary~\ref{cor:offline}]
Run Theorem~\ref{thm:offline} with each event at level $\delta/4$ and add
a count event at level $\delta/4$.
 
\emph{Local counts.} For $i\in I_L$, let
$r_i:=\big(2\lambda\ell_c^2/(c_0\nu_iP(\omega_i)N_2)\big)^{1/(p+2)}$;
$N_2\ge B/(c_0\nu_iP(\omega_i))$ gives $r_i\le\bar r$. By the lower-mass
condition and (C1), the expected count within $r_i$ is at least
$\mu_i:=c_0\nu_iP(\omega_i)N_2r_i^p$, and $N_2\ge A/(c_0\nu_iP(\omega_i))$
is exactly $\mu_i\ge8T_{\max}\ln(4m/\delta)$. Lemma~\ref{lem:chernoff} and
a union bound over $I_L$ give, with probability at least $1-\delta/4$,
$\sum_jB_{j,i}(r_i)\ge\mu_i/2$ for all $i\in I_L$.

\emph{Cost widths.} By Lemma~A.2
of~\cite{atalar2026continuoussemanticcachinglowcost} at radius $r_i$,
$\sigma_c^2(z_i)\le r_i^2/\ell_c^2+2\lambda/\mu_i=2r_i^2/\ell_c^2$, so
$\sigma_c(z_i)\le\sqrt2\,\big(2\lambda/(c_0\ell_c^p\nu_iP(\omega_i)N_2)\big)^{1/(p+2)}$.
Cells of a key partition that key, so $\sum_{i:\omega_i=\omega}w_i\le P(\omega)$,
and by concavity of $t\mapsto t^{(p+1)/(p+2)}$,
$\sum_{i\in I_L}w_iP(\omega_i)^{-1/(p+2)}\le\sum_{\omega}P(\omega)^{(p+1)/(p+2)}\le|\Omega|^{1/(p+2)}$.
With $\nu_i\ge\nu_{\min}$, this gives the cost term.
 
\emph{Risk widths.} By Lemma~A.2
of~\cite{atalar2026continuoussemanticcachinglowcost} with features
$\psi$ and (C2), for $r\le r_0$,
$\sigma_r^2(z_i,z_j)\le r^2/\ell_r^2+\lambda r_0^{p_r}/(\nu_{ij}n_rr^{p_r})$.
Choosing $r^{p_r+2}=\lambda\ell_r^2r_0^{p_r}/(\nu_{ij}n_r)\le r_0^{p_r+2}$
gives
$\sigma_r(z_i,z_j)\le\sqrt2\big(\lambda(r_0/\ell_r)^{p_r}/(\nu_{ij}n_r)\big)^{1/(p_r+2)}$.
H\"older's inequality with $q=p_r+2$ gives
$\sum_{i\in I_R}w_i\nu_{ij^*(i)}^{-1/q}\le W_R^{(q-1)/q}C_R^{1/q}$, and
$\sum_{i\in I_R}w_i\varepsilon_{\mathrm{app}}(z_i,z_{j^*(i)})\le\bar\varepsilon_{\mathrm{app}}W_R$.
 
\emph{Rate.} By Lemma~\ref{lem:netsize},
$m\le|\Omega|2^p/(c_0\epsilon^p)$, so the arrival term is
$\tilde O\big(\sqrt{|\Omega|/(c_0\epsilon^pN_{\mathrm{eff}})}\big)$.
Balancing it against the $O(\epsilon)$ discretization term gives
$\epsilon^*\asymp(|\Omega|/(c_0N_{\mathrm{eff}}))^{1/(p+2)}$, and the cost
term does not depend on $\epsilon$.
\end{proof}

\subsection{Reduction to the Single-Turn Bound}\label{app:reduction}
 
Set $T\equiv1$, $|\Omega|=1$, $\chi\equiv1$, $\gamma=1$,
$c_{\mathrm{in}}\equiv0$ and $r(z,u)=\phi(d(z,u))$ with $\phi$ known and
$L_\phi$-Lipschitz. Then $N_{\mathrm{eff}}=N_2=n_2$, $T_{\max}=1$, the risk
is known so $\bar{\boldsymbol R}=\boldsymbol R$ and the reuse-risk term
vanishes, and $L_r=L_{\mathrm{src}}=L_\phi\le L_g$. With the two remaining
events at level $\delta/2$, Theorem~\ref{thm:offline} becomes
\[
\begin{aligned}
\mathcal L(\widehat{\mathcal M})-\alpha\mathcal L(\mathcal M^*)\le\;&(1+2\alpha)L_g\epsilon\\
&+\tfrac{1+\alpha}{2}\sqrt{\tfrac{2(m\ln2+\ln(2/\delta))}{n_2}}\\
&+2\alpha\beta_c\sum_{i\in I_L}w_i\sigma_c(z_i).
\end{aligned}
\]
Compared with Theorem~4.1
of~\cite{atalar2026continuoussemanticcachinglowcost}, the discretization
term is the same, the arrival term is smaller by a factor of $2$ (step
(iii) above), and the cost term is restricted to the cells $I_L$ where
the comparator invokes the LLM (step (v)). The self-normalized $\beta_c$
also makes the bound valid for adaptively logged data, and
Lemma~\ref{lem:netsize} improves the optimal radius from
$\Theta(n_2^{-1/(d_e+2)})$ to $\Theta(n_2^{-1/(p+2)})$.

\paragraph{What a turn is worth.}
Multi-turn logs count differently for different terms. For the arrival
weights a conversation counts once, because
its turns may be arbitrarily correlated. For the cost, every turn counts: the self-normalized bound tolerates the dependence, and only the
burn-in $N_0$ grows with $T_{\max}$, since one conversation can contribute
$T_{\max}$ correlated observations. Multi-turn data are thus worth between
$J_2$ and $N_2$ single-turn samples, depending on the term.

\section{Proofs for Section~\ref{sec:online}}\label{app:online}
 
Throughout, a superscript $(j)$ marks a quantity computed at the snapshot
used in conversation $j$, and $\sigma_{c,n}$, $\sigma_{r,n}$ denote widths
computed from all data available before round $n$. For a covered round we
write $i_n$ for its cell and $d_n:=\|x_n-x_{i_n}\|_2\le\epsilon$, and
$u_n$ for the reused item on a reuse round. 
The arrival stages of Algorithm~\ref{alg:online} use the stage rule
of~\citet{atalar2026continuoussemanticcachinglowcost}: a stage ends once it
holds at least $1+\sqrt{N\cdot(\text{rounds before it})}$ rounds. The
following four lemmas are used in the proof of Theorem~\ref{thm:online}.
 
\begin{lemma}[Anytime confidence]\label{lem:anytime}
With probability at least $1-\delta/2$, at every snapshot and for all
$i,j$, $0\le c_i-\underline c_i\le2\beta_c\sigma_c(z_i)$ and
$0\le R_{ij}-\underline R_{ij}\le2\gamma(\beta_r\sigma_r+\varepsilon_{\mathrm{app}})(z_i,z_j)$,
with the widths computed from the data at the snapshot.
\end{lemma}
 
Lemma~\ref{lem:anytime} follows from the cost and risk events of
Lemma~\ref{lem:conc}: Theorem~2
of~\cite{chowdhury2017kernelizedmultiarmedbandits} holds uniformly over
time, and online, whether a round yields a cost or a label is decided
before that feedback is revealed.
 
\begin{lemma}[Arrival with correlated turns and a data-built net]\label{lem:arr-online}
Let $\Lambda:=m_{\max}\ln(2NT_{\max})+\ln(32N/\delta)$ and
$t_0:=\max\{\mathbb E[T],\,T_{\max}^2\Lambda/(2\mathbb E[T])\}$. With
probability at least $1-\delta/4$, for all $n\ge t_0$ and every
representative set $\mathcal S$ the algorithm can form from the first $n$
rounds,
\[
\|\hat{\boldsymbol w}_n(\mathcal S)-\boldsymbol w(\mathcal S)\|_1\le\sqrt{\frac{24\bar T\Lambda}{n}}+\frac{2T_{\max}(m_{\max}+1+\Lambda)}{n},
\]
where the uncovered set is included as an extra cell.
\end{lemma}
 
For $T\equiv1$, Lemma~\ref{lem:arr-online} reduces to the arrival event of
Lemma~5.2 in~\cite{atalar2026continuoussemanticcachinglowcost}. The leading
term pays $\bar T$ because a request shares its conversation with $\bar T$
correlated turns on average. The second term pays for the conversations
that host representatives, which cannot be treated as fresh samples.
 
\begin{lemma}[Frozen widths]\label{lem:frozen}
Let $\sigma_s$ be a width computed from the data available before round
$s$, and $G_s:=\ln\det(I+K_s/\lambda)$ for the same data.
(a) For $s\le n$ and every input $v$,
$\sigma_s(v)\le e^{(G_n-G_s)/2}\sigma_n(v)$.
(b) For an RBF kernel with length scale $\ell$,
$\sigma_s(v)\le\sigma_s(v')+\|v-v'\|_2/\ell$.
(c) Let $N_c$ ($N_r$) count the conversation boundaries at which $G$ of the
cost (label) data has grown by at least $\ln2$ since the last switch.
Then $G_n-G_{s(n)}<\ln2$ at every round $n$, where $s(n)$ is the time of
the snapshot in use, except on at most $T_{\max}N_c$ rounds for the cost
data and $T_{\max}N_r$ rounds for the label data.
\end{lemma}
 
Part (a) bounds how stale a frozen width can be, part (b) transfers a
width from a representative to the request it serves, and part (c) counts
the rounds on which the frozen and current widths differ by more than a
factor $\sqrt2$.
 
\begin{lemma}[Number of switches]\label{lem:switch-online}
The number of switches satisfies
$N_{\mathrm{sw}}\le m_{\max}+N_{\mathrm{arr}}+N_c+N_r$, with
$N_{\mathrm{arr}}=O(\log\log N)$ arrival stages,
$N_c\le2\mathcal I_c/\ln2$ and $N_r\le2\mathcal I_r/\ln2$.
\end{lemma}

\subsection{Proof of Lemma~\ref{lem:arr-online}}
 
Representatives are observed rounds. Index every round by its
conversation $j\le N$ and its position $t\le T_{\max}$. A representative
set is determined by the tuple $I$ of positions of its at most $m_{\max}$
representatives, the conversations containing them are its \emph{hosts}.
 
\emph{Step 1: fix the hosts.} Fix $I$ and condition on its host
conversations. This fixes the cells and the uncovered set. We do not
condition on the algorithm selecting $I$, so the non-host conversations
remain independent copies of the random conversation and independent of
the hosts. This lifts the argument of Lemma~5.2
of~\cite{atalar2026continuoussemanticcachinglowcost} from requests to
conversations, which is necessary because conditioning on one turn
conditions on its whole conversation.
 
\emph{Step 2: concentration over non-host conversations.} Fix a sign
vector $\boldsymbol o$ over the $m+1\le m_{\max}+1$ cells, let
$h_{\boldsymbol o}(z):=o_i$ for $z$ in cell $i$ and
$\theta_{\boldsymbol o}:=\boldsymbol o^\top\boldsymbol w$. For a non-host
conversation $j$, let $X_j:=S_j-T_j\theta_{\boldsymbol o}$ with
$S_j:=\sum_th_{\boldsymbol o}(Z_{j,t})$. By~\eqref{eq:perrequest},
$\mathbb E[S_j]=\mathbb E[T]\theta_{\boldsymbol o}=\mathbb E[T_j\theta_{\boldsymbol o}]$,
so $\mathbb E[X_j]=0$; moreover $|X_j|\le2T_{\max}$ and
$\mathrm{Var}(X_j)\le\mathbb E[(|S_j|+T_j)^2]\le4\mathbb E[T^2]$. By
Bernstein's inequality, for each fixed $k$, with probability at least
$1-e^{-\Lambda}$ the sum of $X_j$ over the first $k$ non-host conversations
is at most $\sqrt{8k\mathbb E[T^2]\Lambda}+\frac43T_{\max}\Lambda$.
 
\emph{Step 3: number of conversations.} By Hoeffding's inequality and a
union bound over $k\le N$, with probability at least $1-Ne^{-\Lambda}$,
$\sum_{j\le k}T_j\ge k\mathbb E[T]-T_{\max}\sqrt{k\Lambda/2}$ for all
$k\le N$. If $k$ conversations are complete before round $n$, then
$\sum_{j\le k}T_j\le n$. With
$T_{\max}\sqrt{k\Lambda/2}\le k\mathbb E[T]/2+T_{\max}^2\Lambda/(4\mathbb E[T])$,
this gives $k\le(2n+T_{\max}^2\Lambda/(2\mathbb E[T]))/\mathbb E[T]\le3n/\mathbb E[T]$
for $n\ge t_0$, hence $k\,\mathbb E[T^2]\le3n\bar T$.
 
\emph{Step 4: assemble.} $n\cdot\boldsymbol o^\top(\hat{\boldsymbol w}_n-\boldsymbol w)$
is the sum of $X_j$ over the completed non-host conversations, plus the
contribution of the host rounds (at most $2m_{\max}T_{\max}$) and of the
current partial conversation (at most $2T_{\max}$). With Steps 2 and 3,
\[
\boldsymbol o^\top(\hat{\boldsymbol w}_n-\boldsymbol w)\le\sqrt{\tfrac{24\bar T\Lambda}{n}}+\tfrac{2T_{\max}(m_{\max}+1)+\frac43T_{\max}\Lambda}{n},
\]
and $\|\boldsymbol v\|_1=\max_{\boldsymbol o}\boldsymbol o^\top\boldsymbol v$
gives the claim.
 
\emph{Step 5: union bound.} There are at most $2(NT_{\max})^{m_{\max}}$
tuples, $2^{m_{\max}+1}$ sign vectors and $N$ values of $k$. The total
failure probability is at most
$4N(2NT_{\max})^{m_{\max}}e^{-\Lambda}+Ne^{-\Lambda}\le\delta/8+\delta/32<\delta/4$
by the choice of $\Lambda$.
\hfill$\square$
 
\subsection{Proof of Lemma~\ref{lem:frozen}}
 
(a) Write $k(v,v')=\langle\varphi(v),\varphi(v')\rangle$ and, on the
finite-dimensional span of the features of the data and of $v$, let
$V_s:=\lambda I+\sum\varphi(v_t)\varphi(v_t)^\top$ over the data before
round $s$. Then $\sigma_s^2(v)=\lambda\varphi(v)^\top V_s^{-1}\varphi(v)$.
Since $V_s\preceq V_n$, $V_s^{-1}=V_n^{-1}+C$ with $C\succeq0$, and
Lemma~12 of~\cite{abbasi2011improved} gives
$\varphi^\top V_s^{-1}\varphi\le(\det V_n/\det V_s)\,\varphi^\top V_n^{-1}\varphi$.
By Sylvester's identity,
$\det V_n/\det V_s=\det(I+K_n/\lambda)/\det(I+K_s/\lambda)=e^{G_n-G_s}$.
 
(b) By the variational characterization of Appendix~\ref{app:offline},
$\sigma_s(v)$ is the distance from $(k(v,\cdot),0)$ to the subspace
$\{(\sum_ta_tk(v_t,\cdot),-\sqrt\lambda\boldsymbol a)\}$. A distance to a
fixed subspace is $1$-Lipschitz, so
$|\sigma_s(v)-\sigma_s(v')|\le\|k(v,\cdot)-k(v',\cdot)\|_{\mathcal H}\le\|v-v'\|_2/\ell$,
where the last step is the RBF bound in Lemma~A.2
of~\cite{atalar2026continuoussemanticcachinglowcost}.
 
(c) The condition is checked at every conversation boundary, and when it
holds a switch resets the snapshot. Hence $G_n-G_{s(n)}<\ln2$ at the first
round of every conversation. If the difference reaches $\ln2$ during a
conversation, the boundary at its end satisfies the condition and is
counted in $N_c$ (or $N_r$). Each such conversation has at most $T_{\max}$
rounds, and different conversations end at different boundaries.
\hfill$\square$
 
\subsection{Proof of Lemma~\ref{lem:switch-online}}
 
\emph{New representatives.} A representative can be added at most
$m_{\max}$ times, so at most $m_{\max}$ boundaries switch for this reason.
 
\emph{Information triggers.} Let $b_1<\dots<b_{N_c}$ be the boundaries at
which the cost condition holds, and $s_k$ the last switch before $b_k$.
Then $G_{b_k}-G_{s_k}\ge\ln2$, and the intervals $(s_k,b_k]$ are disjoint
because a switch occurs at every $b_k$. Hence
$N_c\ln2\le G_N-G_0\le2\mathcal I_c$. The same argument gives
$N_r\le2\mathcal I_r/\ln2$.
 
\emph{Arrival stages.} The stage rule is that of Algorithm~3
in~\cite{atalar2026continuoussemanticcachinglowcost}, and the global-stage
part of its Lemma~5.3 (Lemma~7 of~\cite{liu2026semanticcachinglowcostllm})
uses only that each stage holds at least
$1+\sqrt{N\cdot(\text{rounds before it})}$ rounds. Checking the rule only
at conversation boundaries preserves this lower bound, so
$N_{\mathrm{arr}}=O(\log\log N)$.
\hfill$\square$
 
\subsection{Width Sums and Coverage}
 
\begin{lemma}[Width sums with sampled labels]\label{lem:widthsum}
$\sum_{n:\text{LLM}}\sigma_{c,n}(x_n)\le\sqrt{2N\mathcal I_c/\ln(1+1/\lambda)}$
and
$\mathbb E\sum_{n:\text{reuse}}\sigma_{r,n}(\psi(z_n,u_n))\le\sqrt{2N\mathcal I_r/(\zeta\ln(1+1/\lambda))}$.
\end{lemma}
 
\begin{proof}
Every LLM round adds its input to the cost data, so the LLM rounds form
the sequence of cost inputs, and $\sigma_{c,n}(x_n)$ is the width before
the input is added. The first claim is then the Cauchy--Schwarz and
elliptical-potential step~\cite{srinivas2010gaussian} used in the proof
of Theorem~5.1 of~\cite{atalar2026continuoussemanticcachinglowcost}, with
at most $N$ LLM rounds. For the second, let $s_n:=\sigma_{r,n}(\psi(z_n,u_n))$
and let $B_n$ indicate that the label of reuse round $n$ is observed.
Given the past and the reuse decision, $s_n$ is fixed and $B_n$ is
Bernoulli$(\zeta)$ by Assumption~\ref{ass:online}(a), so
$\mathbb E\sum_{\text{reuse}}s_n=\zeta^{-1}\mathbb E\sum_{\text{labeled}}s_n$.
The labeled rounds form the sequence of label inputs, so the same step
gives $\sum_{\text{labeled}}s_n\le\sqrt{N^r\cdot2\mathcal I_r/\ln(1+1/\lambda)}$
with $N^r$ the number of labels. Jensen's inequality and
$\mathbb E N^r\le\zeta N$ complete the proof.
\end{proof}
 
\begin{lemma}[Coverage]\label{lem:coverage}
(a) At most $T_{\max}m_{\max}$ rounds are uncovered. (b) For
$u\in\mathcal U$, let
$\vartheta_u:=P(\omega_Z=\omega_u,\|X-x_u\|_2\le\epsilon/2)\ge P(\omega_Z=\omega_u)c_0(\epsilon/2)^p$.
The expected number of rounds before the net contains a representative
with key $\omega_u$ within $3\epsilon/2$ of $x_u$ is at most
$T_{\max}^2/(\mathbb E[T]\vartheta_u)$. Hence the expected number of rounds
at which some item of $\mathcal M^*$ has no such representative is at
most $\tau_{\mathrm{cov}}:=\sum_{u\in\mathcal M^*}T_{\max}^2/(\mathbb E[T]\vartheta_u)$.
\end{lemma}
 
\begin{proof}
(a) A conversation with an uncovered round adds at least one
representative at its end, so at most $m_{\max}$ conversations contain
uncovered rounds, each with at most $T_{\max}$ of them.
 
(b) Let $B_u:=\{z:\omega_z=\omega_u,\|x_z-x_u\|_2\le\epsilon/2\}$, let $S_j$
be the number of rounds of conversation $j$ in $B_u$, and let
$H_j:=\mathbf 1\{S_j\ge1\}$. By~\eqref{eq:perrequest},
$\mathbb E[S_j]=\mathbb E[T]\vartheta_u$, and $S_j\le T_{\max}H_j$, so
$\Pr(H_j=1)\ge\mathbb E[T]\vartheta_u/T_{\max}$. If conversation $j$
visits $B_u$ at a state $z$, then after its net update some
representative with key $\omega_u$ lies within $\epsilon$ of $z$, and
hence within $3\epsilon/2$ of $x_u$; representatives are never removed.
The index of the first visiting conversation is dominated by a geometric
variable with mean at most $T_{\max}/(\mathbb E[T]\vartheta_u)$, and each
conversation has at most $T_{\max}$ rounds. A union bound over
$u\in\mathcal M^*$ completes the proof.
\end{proof}
 
\subsection{Proof of Theorem~\ref{thm:online}}\label{app:thm-online}
 
With $L^*:=\mathcal L(\mathcal M^*)$, let $X_j:=\sum_{n\in j}\ell_n-\alpha T_jL^*$.
 
\emph{Step 0: events and conversations.} Let $E_j$ be the event that
Lemmas~\ref{lem:anytime} and~\ref{lem:arr-online} hold for the snapshot of
conversation $j$. Then $E_j\in\mathcal G_{j-1}$ and
$\Pr(\cup_jE_j^c)\le3\delta/4$; since $\ell_n\le1$, failures contribute at
most $\delta N=1$. Let $J_N$ be the conversation containing round $N$, a
stopping time for $(\mathcal G_j)$. Since $\ell_n\ge0$ and
$N\le\sum_{j\le J_N}T_j\le N+T_{\max}$,
$\sum_{n\le N}\ell_n-\alpha NL^*\le\sum_{j\le J_N}X_j+\alpha T_{\max}$, and
by optional stopping it suffices to bound $\mathbb E[X_j\mid\mathcal G_{j-1}]$
on $E_j$.
 
\emph{Step 1: per round.} For a covered round $n$ of conversation $j$ with
cell $i$, let
$\hat g_j(i):=\min\{\underline c_i,\min_{u\in\mathcal M_j}\underline R_{iu}\}$,
and $\Delta_n:=c_i-\underline c_i$ on LLM rounds and
$\Delta_n:=R_{iu_n}-\underline R_{iu_n}$ on reuse rounds. The routing rule
picks the option attaining $\hat g_j(i)$. Since $z_n$ and $z_i$ share a
key, the mask is the same at both, and Assumption~\ref{ass:lipschitz}
gives $\ell_n\le\hat g_j(i)+\Delta_n+L_gd_n$. Uncovered rounds have
$\ell_n\le1$.
 
\emph{Step 2: average over a conversation.} The snapshot $\Pi_j$ is
$\mathcal G_{j-1}$-measurable, and conversation $j$ is independent of
$\mathcal G_{j-1}$ with a law that does not depend on how it is served
(Assumption~\ref{ass:exogenous}). By~\eqref{eq:perrequest}, the expected
number of its rounds in $V_i$ is $\mathbb E[T]w_i(\mathcal S_j)$, so
\[
\mathbb E\Big[\textstyle\sum_{n\in j}\hat g_j(i_n)\mathbf 1\{\text{cov}\}\,\Big|\,\mathcal G_{j-1}\Big]=\mathbb E[T]\,F(\mathcal M_j;\boldsymbol w,\underline{\boldsymbol c},\underline{\boldsymbol R}),
\]
with $\boldsymbol w=\boldsymbol w(\mathcal S_j)$ restricted to the covered
cells. Likewise
$\mathbb E[\sum_{n\in j}d_n\mathbf 1\{\text{cov}\}\mid\mathcal G_{j-1}]=\mathbb E[T]\bar\epsilon(\mathcal S_j)$
with $\bar\epsilon(\mathcal S):=\mathbb E_P[\|X-x_{i(X)}\|_2\mathbf 1\{\text{cov}\}]$.
 
\emph{Step 3: chain at the snapshot.} Let $\mathcal M^\circ_j$ replace each
item of $\mathcal M^*$ by its nearest representative in $\mathcal S_j$ with
the same key, and let $U_j$ be the event that some item has none within
$3\epsilon/2$. On $E_j$, with $\|\cdot\|_1:=\|\boldsymbol w-\hat{\boldsymbol w}_j\|_1$,
\[
\begin{aligned}
F(\mathcal M_j;\boldsymbol w,\underline{\boldsymbol c},\underline{\boldsymbol R})
&\le F(\mathcal M_j;\hat{\boldsymbol w}_j,\underline{\boldsymbol c},\underline{\boldsymbol R})+\|\cdot\|_1\\
&\le\alpha F(\mathcal M^\circ_j;\hat{\boldsymbol w}_j,\underline{\boldsymbol c},\underline{\boldsymbol R})+\|\cdot\|_1\\
&\le\alpha F(\mathcal M^\circ_j;\hat{\boldsymbol w}_j,\boldsymbol c,\boldsymbol R)+\|\cdot\|_1\\
&\le\alpha F(\mathcal M^\circ_j;\boldsymbol w,\boldsymbol c,\boldsymbol R)+(1+\alpha)\|\cdot\|_1\\
&\le\alpha L^*+\alpha L_g\bar\epsilon(\mathcal S_j)+\alpha\gamma L_{\mathrm{src}}\epsilon^{\mathrm{proj}}_j\\
&\quad+\alpha\mathbf 1\{U_j\}+(1+\alpha)\|\cdot\|_1.
\end{aligned}
\]
The first and fourth steps are Lemma~4 of~\cite{liu2026semanticcachinglowcostllm};
the weights on the covered cells form a sub-probability vector, so the
constant is $1$. The second is Lemma~\ref{lem:supermod}, since
$\mathcal M_j$ is Reverse Greedy's output on exactly this instance; this
is why the snapshot is frozen. The third holds because $F$ is monotone in
$\boldsymbol c$ and $\boldsymbol R$, and
$\underline{\boldsymbol c}\le\boldsymbol c$,
$\underline{\boldsymbol R}\le\boldsymbol R$ by Lemma~\ref{lem:anytime}. The
fifth uses $F\le1$ on $U_j$. Off $U_j$, it uses the proof of
Lemma~\ref{lem:disc}, which bounds $|g(z)-g(z_i)|$ pointwise and therefore
gives $F\le\mathcal L+L_g\bar\epsilon$ on the covered cells, together
with~\eqref{eq:proj} with $\epsilon$ replaced by
$\epsilon^{\mathrm{proj}}_j$.
 
\emph{Step 4: widths.} On a covered LLM round, Lemma~\ref{lem:anytime} and
Lemma~\ref{lem:frozen}(b) give
$\Delta_n\le2\beta_c\sigma^{(j)}_c(x_{i_n})\le2\beta_c(\sigma^{(j)}_c(x_n)+d_n/\ell_c)$,
and Lemma~\ref{lem:frozen}(a,c) gives
$\sigma^{(j)}_c(x_n)\le\sqrt2\,\sigma_{c,n}(x_n)$ except on at most
$T_{\max}N_c$ rounds, where we use $\Delta_n\le1$. On a covered reuse round,
in the same way and with Assumption~\ref{ass:online}(b),
$\Delta_n\le2\gamma(\beta_r\sigma^{(j)}_r+\varepsilon^{(j)}_{\mathrm{app}})(\psi(z_{i_n},u_n))$,
$\sigma^{(j)}_r(\psi(z_{i_n},u_n))\le\sigma^{(j)}_r(\psi(z_n,u_n))+L_\psi d_n/\ell_r$,
and $\sigma^{(j)}_r\le\sqrt2\,\sigma_{r,n}$ except on at most $T_{\max}N_r$
rounds. Summing with Lemma~\ref{lem:widthsum},
\[
\begin{aligned}
\mathbb E\textstyle\sum_{\text{cov}}\Delta_n\le\;&2\sqrt2\,\beta_c\sqrt{\tfrac{2N\mathcal I_c}{\ln(1+1/\lambda)}}\\
&+2\sqrt2\,\gamma\beta_r\sqrt{\tfrac{2N\mathcal I_r}{\zeta\ln(1+1/\lambda)}}+E_{\mathrm{mis}}\\
&+2\max\{\tfrac{\beta_c}{\ell_c},\tfrac{\gamma\beta_rL_\psi}{\ell_r}\}Q_N\\
&+T_{\max}(N_c+N_r).
\end{aligned}
\]
Since $\varepsilon^{(j)}_{\mathrm{app}}\le\varepsilon_0(1+\sqrt{N/\lambda}\,\sigma^{(j)}_r)$,
the misspecification term satisfies
$E_{\mathrm{mis}}\le2\gamma\varepsilon_0(N+\sqrt{N/\lambda}\,\mathbb ES_r)$
with
$\mathbb ES_r\le\sqrt2\sqrt{2N\mathcal I_r/(\zeta\ln(1+1/\lambda))}+T_{\max}N_r+L_\psi Q_N/\ell_r$.
 
\emph{Step 5: sum over conversations.} Steps 1--4 give, on $E_j$,
$\mathbb E[X_j\mid\mathcal G_{j-1}]\le\mathbb E[T](F(\mathcal M_j;\boldsymbol w,\underline{\boldsymbol c},\underline{\boldsymbol R})-\alpha L^*)+\mathbb E[\sum_{n\in j}((\Delta_n+L_gd_n)\mathbf 1\{\text{cov}\}+\mathbf 1\{\text{unc}\})\mid\mathcal G_{j-1}]$.
Insert Step 3 and sum over $j\le J_N$, using
$\sum_j\mathbb E[T]Y_j=\mathbb E\sum_jT_jY_j$ for $\mathcal G_{j-1}$-measurable
$Y_j$. The terms $(1+\alpha)L_gQ_N$, the transfer terms of Step 4 and
$\alpha\gamma L_{\mathrm{src}}\sum_n\mathbb E\epsilon^{\mathrm{proj}}_n$ form the
discretization line. Lemma~\ref{lem:coverage} bounds the uncovered rounds
by $T_{\max}m_{\max}$ and the rounds in $U$ by $\tau_{\mathrm{cov}}$, and
Lemma~\ref{lem:switch-online} bounds the switching cost by
$kN_{\mathrm{sw}}$. It remains to bound the arrival sum
$(1+\alpha)\mathbb E\sum_n\|\boldsymbol w-\hat{\boldsymbol w}_{j(n)}\|_1$.
 
Let $A_\tau$ be the number of rounds before stage $\tau+1$ and $L_\tau$ the
length of stage $\tau$, with $A_0=0$. A stage ends at the first boundary
at which it holds at least $1+\sqrt{NA_{\tau-1}}$ rounds, so
$L_\tau<1+T_{\max}+\sqrt{NA_{\tau-1}}$, and every stage starts with a
switch, so the snapshot used in stage $\tau$ is taken after at least
$A_{\tau-1}$ rounds. Let $\tau^*$ be the last stage with
$A_{\tau^*-1}<t_1$. On the rounds up to $A_{\tau^*}$ we use
$\|\cdot\|_1\le2$, which contributes at most
$2A_{\tau^*}<2(t_1+1+T_{\max}+\sqrt{Nt_1})$. On a later stage,
Lemma~\ref{lem:arr-online} applies because $A_{\tau-1}\ge t_1\ge t_0$,
and with $b:=2T_{\max}(m_{\max}+1+\Lambda)\le t_1\le A_{\tau-1}$,
\[
\begin{aligned}
L_\tau\Big(\sqrt{\tfrac{24\bar T\Lambda}{A_{\tau-1}}}+\tfrac{b}{A_{\tau-1}}\Big)\le\;&\sqrt{24\bar T\Lambda}\,(\sqrt N+1+T_{\max})\\
&+\sqrt{Nb}+1+T_{\max}.
\end{aligned}
\]
Summing over at most $N_{\mathrm{arr}}$ stages, and using $b\le t_1$, gives
the arrival term of the theorem plus
$R_{\mathrm{arr}}:=N_{\mathrm{arr}}(1+T_{\max})(1+\sqrt{24\bar T\Lambda})+2(t_1+1+T_{\max})$.
Collecting the terms,
\[
\begin{aligned}
R_{\mathrm{low}}:=\;&(1+\alpha)R_{\mathrm{arr}}+T_{\max}(m_{\max}+N_c+N_r)\\
&+\alpha\tau_{\mathrm{cov}}+\alpha T_{\max}+1.
\end{aligned}
\]
For fixed $\epsilon$, $\Lambda$, $t_1$, $N_c$, $N_r$ and $N_{\mathrm{arr}}$
grow at most polylogarithmically in $N$, and $\tau_{\mathrm{cov}}$ does not
depend on $N$.
\hfill$\square$
 
\subsection{Proof of Corollary~\ref{cor:online}}\label{app:cor-online}
Let $L'_{\mathrm{disc}}:=L_{\mathrm{disc}}+\tfrac32\alpha\gamma L_{\mathrm{src}}$.
The explicit form of Corollary~\ref{cor:online}(a) is: choosing
\[
\epsilon\asymp\Big(\tfrac{T_{\max}^2|\Omega|}{\mathbb E[T]c_0{L'_{\mathrm{disc}}}^2N}\Big)^{\frac1{p+2}},
\]
\[
\begin{aligned}
\mathrm{Reg}(N)=\;&\tilde O\Big({L'_{\mathrm{disc}}}^{\frac p{p+2}}\big(\tfrac{T_{\max}^2|\Omega|}{\mathbb E[T]c_0}\big)^{\frac1{p+2}}N^{\frac{p+1}{p+2}}\Big)\\
&+\tilde O\big(\beta_c\sqrt{N\mathcal I_c}+\gamma\beta_r\sqrt{N\mathcal I_r/\zeta}\big)\\
&+E_{\mathrm{mis}}.
\end{aligned}
\]
 
(a) By definition $Q_N\le\epsilon N$ and $\epsilon^{\mathrm{proj}}_n\le3\epsilon/2$,
so the discretization line is at most $L'_{\mathrm{disc}}\epsilon N$. By
Lemma~\ref{lem:netsize}, $m_{\max}\le|\Omega|2^p/(c_0\epsilon^p)$, so
$\Lambda=O(m_{\max}\log(NT_{\max}))$ and
$t_1=O(T_{\max}^2\Lambda/\mathbb E[T])$, using
$T_{\max}\le T_{\max}^2/\mathbb E[T]$. Since also
$\bar T\le T_{\max}^2/\mathbb E[T]$, the arrival term is
$\tilde O\big(\sqrt{T_{\max}^2|\Omega|N/(\mathbb E[T]c_0\epsilon^p)}\big)$.
Balancing it against $L'_{\mathrm{disc}}\epsilon N$ gives the stated
$\epsilon$ and rate. Switching and $R_{\mathrm{low}}$ are
$\tilde O(\epsilon^{-p})=\tilde O(N^{p/(p+2)})$, since
$\vartheta_u\ge P(\omega_Z=\omega_u)c_0(\epsilon/2)^p$, and are of lower
order.
 
(b) Support points of the same key are more than $\epsilon$ apart, so the
first request at a support point is uncovered and becomes a
representative located exactly there. Every later request at that point
is served from it, so $Q_N=0$. Items of $\mathcal M^*$ are support points,
so $\epsilon^{\mathrm{proj}}_n=0$ once each has been seen, and earlier
rounds are counted by $\tau_{\mathrm{cov}}$ with $\vartheta_u$ the mass of
$u$. The cost inputs take at most $m_{\max}$ values and the label inputs
at most $m_{\max}^2$, so the kernel matrices have rank at most $m_{\max}$
and $m_{\max}^2$, which gives
$\mathcal I_c\le\frac{m_{\max}}2\ln(1+N/\lambda)$ and
$\mathcal I_r\le\frac{m_{\max}^2}2\ln(1+N/\lambda)$. All remaining terms
of Theorem~\ref{thm:online} are then $\tilde O(\sqrt N)$ or lower order.
\hfill$\square$

\paragraph{Single-turn special case.}
Setting $T\equiv1$, $|\Omega|=1$, $\chi\equiv1$, $\gamma=1$,
$c_{\mathrm{in}}\equiv0$ and $r=\phi\circ d$ with known $\phi$ gives $\bar T=T_{\max}=1$, removes the reuse-risk term and the risk switches, and yields $\tilde O(\beta_c\sqrt{N\mathcal I_c}+\sqrt{m_{\max}N}\log\log N)$
plus $L_{\mathrm{disc}}Q_N+\alpha L_\phi\sum_n\mathbb E\epsilon^{\mathrm{proj}}_n$. For fixed problem parameters, the separated-support condition of
Corollary~\ref{cor:online}(b) yields $\tilde O(\sqrt N)$ regret, matching the horizon dependence of Theorem~5.1 in~\citet{atalar2026continuoussemanticcachinglowcost}. That result likewise relies on a structural concentration condition: Assumption~2 there requires a bounded effective number of centers and approximation distances
decaying as $O(t^{-1/2})$, and our separated-support condition serves the same purpose.

\paragraph{What conversations cost online.}
Compared with a stream of independent requests, conversations change only
constants and lower-order terms. The arrival term carries
$T_{\max}^2/\mathbb E[T]\ge\bar T$, which equals $1$ for $T\equiv1$, the
burn-in grows with $T_{\max}m_{\max}$, and snapshots add the
$N_{\mathrm{arr}}$ factor. The learned risk adds the reuse-risk term,
which vanishes when the risk is known.

\section{Additional Experimental Results}
\label{app:experiments}

\subsection{Workload Screening and the Motivation Study}
\label{app:screening}
 
\paragraph{Protocol.}
Figure~\ref{fig:motivation} measures how well each score separates acceptable
from unacceptable reuse when every follow-up turn (turn $\ge2$) is served by its
nearest state in another conversation. States are
$x=[\sqrt{0.7}\,e(q);\sqrt{0.3}\,e(H)]$ with Granite embeddings. The pools are
a 52k-conversation WildChat sample (59,648 follow-ups of 20k target
conversations), screening samples of at most 6,000 conversations for
Taskmaster-3, MultiDoc2Dial and ABCD (53,781, 26,772 and 32,456 follow-ups),
and the 2,200 SocraTeach dialogues of Section~\ref{sec:experiments} (8,621
follow-ups). For each workload, 2,109--2,897 (follow-up, neighbour) pairs are
drawn stratified by distance band (width 0.05 up to 0.30, at most two per target
conversation and band), labeled by the GPT-4o-mini judge of
Section~\ref{sec:experiments}, and reweighted to the population of follow-ups.
The learned risk is a Nystr\"om RBF kernel ridge regression (rank 512) on a
40-dimensional pair feature $\psi$: similarities of the two states, requests,
standalone rewrites and histories, and of the target request to the cached
response; same-intent, same-task and exact-match indicators; turn positions and
history lengths; and 8-dimensional projections of both states and the cached
response. Its bandwidth and ridge are chosen by nested grouped cross-validation
on the Brier score, and all scores are out of fold (5 folds grouped by
conversation). The configuration was fixed on WildChat and not tuned on the
other workloads. CIs use 1,000 cluster-bootstrap resamples.
 
\paragraph{Results.}
Table~\ref{tab:motivation} also reports safe reuse at 80\% judge precision: the
share of all follow-ups that a score can serve acceptably while at least 80\% of
its served pairs are accepted. Among the five screened task-oriented and
document-grounded corpora (Taskmaster-3, MultiDoc2Dial, ABCD, SGD, MultiWOZ~2.2),
Taskmaster-3 had the most follow-up reuse; in WildChat, both scores serve under
2\% of follow-ups at this precision. On SocraTeach, the two coincide at 80\%
precision and the learned risk serves more below it (53.0\% vs.\ 38.3\% at
70\%): beyond distance 0.10, neighbours from the same problem are accepted
55--64\% of the time and those from other problems 9--15\%. The learned risk
finds the former, but they never reach 80\%.
 

\begin{table}[t]
\centering
\caption{The motivation study (Figure~\ref{fig:motivation}). AUC for acceptable
reuse and safe reuse at 80\% judge precision (\% of follow-ups, 95\% CI) for
embedding distance (Dist.) and the learned risk (Learned).}
\label{tab:motivation}
\vspace{4pt}
\footnotesize
\setlength{\tabcolsep}{3pt}
\resizebox{\columnwidth}{!}{%
\begin{tabular}{l cc cc}
\toprule
& \multicolumn{2}{c}{AUC} & \multicolumn{2}{c}{Safe reuse at 80\%} \\
\cmidrule(lr){2-3}\cmidrule(lr){4-5}
Workload & Dist. & Learned & Dist. & Learned \\
\midrule
WildChat      & 0.69 & 0.76 & 1.1               & 1.8 [0.5, 3.6] \\
Taskmaster-3  & 0.74 & 0.89 & 17.3 [14.2, 22.7] & 38.9 [32.2, 46.4] \\
SocraTeach    & 0.67 & 0.75 & 20.5 [10.8, 24.0] & 15.5 [8.1, 33.4] \\
MultiDoc2Dial & 0.72 & 0.81 & 0.0 [0.0, 6.3]    & 22.3 [16.3, 27.1] \\
ABCD          & 0.62 & 0.82 & 0.0 [0.0, 0.2]    & 13.9 [10.9, 18.8] \\
\bottomrule
\end{tabular}}
\end{table}

\subsection{Baselines}
\label{app:baselines}
 
ElastiCache-MT follows the semantic-caching recipe of Amazon ElastiCache~\cite{aws_elasticache_doc}: it embeds the most recent messages of the conversation as one string and serves the nearest cached entry if their cosine similarity exceeds a threshold. ContextCache~\cite{contextcache} retrieves candidates by the similarity of the current message and serves one only if the mean embedding of the last five user messages also matches; it also serves exact context matches. vCache~\cite{gaspar2026vcache} learns a similarity threshold for each cached entry from correctness feedback, keeping its error rate below a target $\delta$; it uses ElastiCache-MT's embedding. The joint-state threshold applies a single cosine threshold to our state embedding $x_t$. Single-turn is CUCB-SC-Cont~\cite{atalar2026continuoussemanticcachinglowcost} applied to the current message alone, without history.

Offline, ElastiCache-MT, ContextCache and the joint-state threshold cache the $k$ source states that are most often the nearest neighbor of a source state from another conversation, and vCache builds its cache by replaying the source
conversations with judge feedback. Online, the threshold baselines insert every LLM-generated response and evict the least recently used (LRU) entry when the cache is full; vCache inserts a response when its exploration finds the cached one unacceptable, also with LRU eviction. Without a budget, nothing is evicted. Online, CLCB-LS-MT updates its cost and risk models and its $\epsilon$-net at the end of each conversation and recomputes its cache when needed; the same lower-confidence rule decides reuses from its session buffer, which is cleared when the conversation ends.

\textbf{Workload details}. SocraTeach has 35,151 dialogues on 11,147 problems, with students drawn from six cognitive profiles; its responses are LLM-generated. Taskmaster-3 has 23,757 dialogues across 43 scenarios, each written by one crowd worker playing both customer and agent; it had the most follow-up reuse
among five screened task-oriented and document-grounded corpora (Appendix \ref{app:screening}). Mean test costs are 0.36 on Taskmaster-3 and 0.33 on SocraTeach. Without a budget, CUCB-SC-MT and single-turn cache every center of their $\epsilon$-nets offline (3,168 and 4,108 centers on Taskmaster-3; 2,278 and 4,836 on SocraTeach), and CLCB-LS-MT ends the stream with 5,799 net centers on Taskmaster-3 and 5,346 on SocraTeach. Its session buffer is cleared when the conversation ends.

\subsection{Latency}
\label{app:latency}
 
Tables~\ref{tab:cache-latency-taskmaster3-offline}--\ref{tab:online-latency-socrateach}
report the per-budget latency estimates of Section~\ref{sec:latency}. Each table
was timed in a single session with its own LLM measurement: $t_{\mathrm{LLM}}$ is
891 and 1,471\,ms offline and 858 and 1,100\,ms online on Taskmaster-3 and
SocraTeach. The LLM calls generate the assistant's reply to sampled requests given
the conversation so far. Offline, the hit rate is that of the selected cache on the
test requests. Online, it covers the whole stream (including the cold start),
decisions are timed at each method's final cache (an upper bound for caches that
grow), and methods are timed round-robin, request by request, so that they share the
machine's state. For CLCB-LS-MT, the decision also includes the net-coverage check
and the session candidates. Without a budget, CUCB-SC-MT's offline lookup over
3,168 net centers on Taskmaster-3 takes 132\,ms, which raises its estimate to 0.90.
 
Online methods also work after each response: CLCB-LS-MT and single-turn update their models and nets and recompute their caches at the end of conversations
(0.11--0.28\,s per conversation for CLCB-LS-MT), the threshold baselines insert and
evict entries, and vCache records observations. The online estimates add this
work, amortized per request, to the request path, which is conservative because it
can run after the response is returned. CLCB-LS-MT needs a judge label for each reuse
(0.28--0.32 labels per request on Taskmaster-3 and 0.36--0.40 on SocraTeach at
$k=100$--$1000$), and vCache one for each exploration (0.85--0.99 and
0.99--1.00); these are extra LLM calls made off the request path.
Methods that reuse aggressively can have lower estimated latency than CLCB-LS-MT
while losing more, e.g., ContextCache at $k=100$ on Taskmaster-3 (0.76; loss
109.2\%) and the unbounded ElastiCache-MT, joint-state threshold and ContextCache on
SocraTeach (0.10, 0.41 and 0.53; losses 154.7\%, 129.8\% and 81.9\%).

\begin{table*}[t]
    \centering
    \renewcommand{\arraystretch}{1.25}
    \caption{Offline latency estimates on Taskmaster-3. Budget columns report the
    estimated mean per-request latency relative to the no-cache LLM-call latency
    ($t_{\mathrm{LLM}}=891$\,ms); values below $1$ indicate lower expected latency.
    Ranges span $k=100$--$1000$; the last column removes the cache-size constraint.
    Without a budget, CUCB-SC-MT's lookup over 3,168 net centers takes 132\,ms.}
    \setlength{\tabcolsep}{5pt}
    \resizebox{\textwidth}{!}{%
    \begin{tabular}{l*{10}{c}}
        \toprule
        \multicolumn{5}{c}{}
        & \multicolumn{6}{c}{Expected latency relative to no cache} \\
        \cmidrule(lr){6-11}
        Method
        & \makecell{Embeddings \\ (ms)}
        & \makecell{Lookup (ms) \\ $k=100$--$1000$}
        & \makecell{Hit latency \\ (ms)}
        & \makecell{Hit rate \\ $k=100$--$1000$}
        & $k=100$ & $k=200$ & $k=300$ & $k=500$ & $k=1000$ & no budget \\
        \midrule
        CUCB-SC-MT            & 72.3 & 9.76--30.46 & 82--103 & 25.5--30.6\% & 0.84 & 0.82 & 0.81 & 0.81 & 0.81 & 0.90 \\
        \midrule
        ElastiCache-MT        & 37.2 & 0.02--0.11  & 37      & 6.1--20.9\%  & 0.98 & 0.94 & 0.91 & 0.84 & 0.83 & 0.76 \\
        \midrule
        ContextCache          & 34.4 & 0.48--0.67  & 35      & 6.0--22.7\%  & 0.98 & 0.96 & 0.93 & 0.87 & 0.81 & 0.68 \\
        \midrule
        vCache                & 37.2 & 2.02--2.32  & 39--40  & 0.2--13.4\%  & 1.04 & 1.04 & 1.03 & 1.00 & 0.91 & 0.87 \\
        \midrule
        Joint-state threshold & 72.3 & 0.02--0.12  & 72      & 0.3--6.1\%   & 1.08 & 1.06 & 1.06 & 1.03 & 1.02 & 0.96 \\
        \midrule
        Single-turn           & 34.4 & 0.82--1.09  & 35      & 0.1--9.1\%   & 1.02 & 0.95 & 1.01 & 1.04 & 1.01 & 1.01 \\
        \bottomrule
    \end{tabular}%
    }
    
    \label{tab:cache-latency-taskmaster3-offline}
\end{table*}

\begin{table*}[t]
    \centering
    \renewcommand{\arraystretch}{1.25}
    \caption{Offline latency estimates on SocraTeach. Budget columns report the
    estimated mean per-request latency relative to the no-cache LLM-call latency
    ($t_{\mathrm{LLM}}=1{,}471$\,ms); values below $1$ indicate lower expected latency.
    Ranges span $k=100$--$1000$; the last column removes the cache-size constraint.}
    \setlength{\tabcolsep}{5pt}
    \resizebox{\textwidth}{!}{%
    \begin{tabular}{l*{10}{c}}
        \toprule
        \multicolumn{5}{c}{}
        & \multicolumn{6}{c}{Expected latency relative to no cache} \\
        \cmidrule(lr){6-11}
        Method
        & \makecell{Embeddings \\ (ms)}
        & \makecell{Lookup (ms) \\ $k=100$--$1000$}
        & \makecell{Hit latency \\ (ms)}
        & \makecell{Hit rate \\ $k=100$--$1000$}
        & $k=100$ & $k=200$ & $k=300$ & $k=500$ & $k=1000$ & no budget \\
        \midrule
        CUCB-SC-MT            & 35.7 & 10.80--76.87 & 46--113 & 6.8--47.0\% & 0.99 & 0.89 & 0.84 & 0.65 & 0.61 & 0.58 \\
        \midrule
        ElastiCache-MT        & 16.5 & 0.03--0.21   & 17      & 0.4--13.2\% & 1.01 & 1.01 & 1.01 & 1.01 & 0.88 & 0.02 \\
        \midrule
        ContextCache          & 16.2 & 0.45--0.58   & 17      & 1.1--20.6\% & 1.00 & 1.00 & 1.00 & 0.93 & 0.81 & 0.47 \\
        \midrule
        vCache                & 16.5 & 1.65--2.71   & 18--19  & 0.0--0.2\%  & 1.01 & 1.01 & 1.01 & 1.01 & 1.01 & 0.69 \\
        \midrule
        Joint-state threshold & 35.7 & 0.01--0.11   & 36      & 0.4--17.9\% & 1.02 & 1.02 & 1.01 & 0.97 & 0.85 & 0.22 \\
        \midrule
        Single-turn           & 16.2 & 0.84--1.03   & 17      & 0.6--12.2\% & 1.01 & 0.98 & 0.97 & 0.92 & 0.89 & 0.62 \\
        \bottomrule
    \end{tabular}%
    }
    
    \label{tab:cache_latency_comparison}
\end{table*}

\begin{table*}[t]
    \centering
    \renewcommand{\arraystretch}{1.25}
    \caption{Online latency estimates on Taskmaster-3. Budget columns report the
    estimated mean per-request latency, including the amortized background work,
    relative to the no-cache LLM-call latency ($t_{\mathrm{LLM}}=858$\,ms); lower is better. Decisions are timed at the final
    caches; background work runs after the response, off the request path. These
    are component-based estimates rather than measured end-to-end latencies.}
    \setlength{\tabcolsep}{5pt}
    \resizebox{\textwidth}{!}{%
    \begin{tabular}{l*{11}{c}}
        \toprule
        \multicolumn{6}{c}{}
        & \multicolumn{6}{c}{Estimated latency relative to no cache} \\
        \cmidrule(lr){7-12}
        Method
        & \makecell{Embeddings \\ (ms)}
        & \makecell{Decision \\ (ms)}
        & \makecell{Hit latency \\ (ms)}
        & \makecell{Background \\ (ms/request)}
        & \makecell{Hit rate \\ $k=100$--$1000$}
        & $k=100$ & $k=200$ & $k=300$ & $k=500$ & $k=1000$ & \makecell{No \\ budget} \\
        \midrule
        CLCB-LS-MT
        & 70.4 & 12.0--12.4 & 82--83 & 27.8--27.9 & 28.1--32.7\% & 0.85 & 0.82 & 0.81 & 0.80 & 0.81 & 0.81 \\
        \midrule
        ContextCache
        & 33.7 & 1.3--1.5 & 35 & 0.006--0.009 & 16.7--27.8\% & 0.76 & 0.87 & 0.85 & 0.81 & 0.79 & 0.75 \\
        \midrule
        ElastiCache-MT
        & 36.3 & 0.1--0.3 & 36--37 & 0.005 & 10.3--21.3\% & 0.94 & 0.91 & 0.89 & 0.83 & 0.84 & 0.77 \\
        \midrule
        vCache
        & 36.3 & 1.7--4.4 & 38--41 & 0.010--0.013 & 0.3--14.8\% & 1.04 & 1.04 & 1.03 & 0.98 & 0.90 & 0.92 \\
        \midrule
        Joint-state threshold
        & 70.4 & 0.1--0.3 & 70--71 & 0.005--0.006 & 0.7--3.5\% & 1.08 & 1.06 & 1.06 & 1.05 & 1.05 & 0.99 \\
        \midrule
        \makecell[l]{Single-turn \\ (CUCB-SC-Cont)}
        & 33.7 & 4.6--5.3 & 38--39 & 8.4--14.6 & 0.1--5.5\% & 1.04 & 1.01 & 1.04 & 1.05 & 1.03 & 1.03 \\
        \bottomrule
    \end{tabular}%
    }
    
    \label{tab:online-latency-taskmaster3}
\end{table*}

\begin{table*}[t]
    \centering
    \renewcommand{\arraystretch}{1.25}
    \caption{Online latency estimates on SocraTeach. Budget columns report the
    estimated mean per-request latency, including the amortized background work,
    relative to the no-cache LLM-call latency ($t_{\mathrm{LLM}}=1{,}100$\,ms); lower is better. Decisions are timed at the final
    caches; background work runs after the response, off the request path. These
    are component-based estimates rather than measured end-to-end latencies.}
    \setlength{\tabcolsep}{5pt}
    \resizebox{\textwidth}{!}{%
    \begin{tabular}{l*{11}{c}}
        \toprule
        \multicolumn{6}{c}{}
        & \multicolumn{6}{c}{Estimated latency relative to no cache} \\
        \cmidrule(lr){7-12}
        Method
        & \makecell{Embeddings \\ (ms)}
        & \makecell{Decision \\ (ms)}
        & \makecell{Hit latency \\ (ms)}
        & \makecell{Background \\ (ms/request)}
        & \makecell{Hit rate \\ $k=100$--$1000$}
        & $k=100$ & $k=200$ & $k=300$ & $k=500$ & $k=1000$ & \makecell{No \\ budget} \\
        \midrule
        CLCB-LS-MT
        & 31.8 & 7.2--7.3 & 39 & 22.6--23.2 & 35.7--40.3\% & 0.70 & 0.69 & 0.69 & 0.67 & 0.65 & 0.59 \\
        \midrule
        ContextCache
        & 15.4 & 0.6 & 16 & 0.002 & 7.0--29.7\% & 0.82 & 0.80 & 0.94 & 0.77 & 0.72 & 0.53 \\
        \midrule
        ElastiCache-MT
        & 16.3 & 0.026--0.2 & 16--17 & 0.002 & 1.8--6.9\% & 0.99 & 1.00 & 0.99 & 0.97 & 0.95 & 0.10 \\
        \midrule
        vCache
        & 16.3 & 1.0--1.6 & 17--18 & 0.004--0.005 & 0.1--0.7\% & 1.01 & 1.01 & 1.01 & 1.01 & 1.01 & 0.65 \\
        \midrule
        Joint-state threshold
        & 31.8 & 0.035--0.2 & 32 & 0.002 & 1.1--12.2\% & 1.02 & 1.01 & 0.97 & 0.95 & 0.91 & 0.41 \\
        \midrule
        \makecell[l]{Single-turn \\ (CUCB-SC-Cont)}
        & 15.4 & 3.4--3.5 & 19 & 7.8--9.8 & 0.1--3.4\% & 1.02 & 1.02 & 1.02 & 1.01 & 0.99 & 0.89 \\
        \bottomrule
    \end{tabular}%
    }
    
    \label{tab:online-latency-socrateach}
\end{table*}

\subsection{Online reuse quality}
\label{app:reuse}
 
Tables~\ref{tab:online-reuse-quality-taskmaster3}
and~\ref{tab:online-reuse-quality-socrateach} extend Table~\ref{tab:reuse-key} to
every method and budget. They add the hit rate, the share of hits that reuse a
response of the same conversation, and the regret against the best fixed
$k$-entry cache in hindsight. ``Other conv.\ only'' variants never serve a
response of the current conversation; for CLCB-LS-MT, this removes the session
buffer.
 
On SocraTeach, 54\% to 88\% of CLCB-LS-MT's hits at $k=100$--$1000$ reuse an
earlier response of the same conversation, and removing the session buffer raises
its loss by 8.6 to 14.0\,pp. On Taskmaster-3, where same-conversation responses are
less often acceptable, these hits are 18\% to 23\% of its hits and the buffer is
worth 1.9 to 2.5\,pp. ContextCache also relies on the current conversation: its LRU
cache holds the conversation's recent turns, which supply 59\% to 94\%
(SocraTeach) and 39\% to 91\% (Taskmaster-3) of its hits at $k\le 1000$. On
Taskmaster-3, where these responses are rarely acceptable, this is why ContextCache
exceeds the no-cache loss at $k\le 300$ (up to 109.2\%). When both
methods may only reuse responses from other conversations, CLCB-LS-MT's loss is
lower than ContextCache's at every budget, by 2.9 to 13.3\,pp on SocraTeach and by
16.0 to 21.9\,pp on Taskmaster-3. At $k=1000$, CLCB-LS-MT's acceptable reuses fall
on requests that cost 1.45 (Taskmaster-3) and 1.51 (SocraTeach) times the average
request, against 1.32 (ElastiCache-MT) and 1.15 (ContextCache) on Taskmaster-3 and
1.10 (ContextCache) on SocraTeach.


\subsection{Loss tables, learning over the stream and the price of a reuse}
\label{app:diagnostics}
\begin{figure}[t]
    \centering
    \includegraphics[width=\columnwidth]{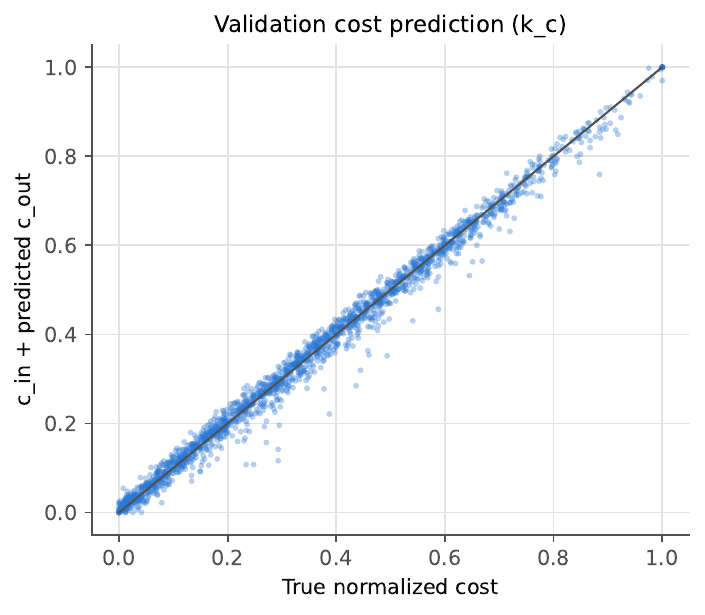}
    \caption{Validation cost prediction in offline setting for Taskmaster.}
    \label{fig:cost_pred_taskmaster_offline}
\end{figure}

\begin{figure}[t]
    \centering
    \includegraphics[width=\columnwidth]{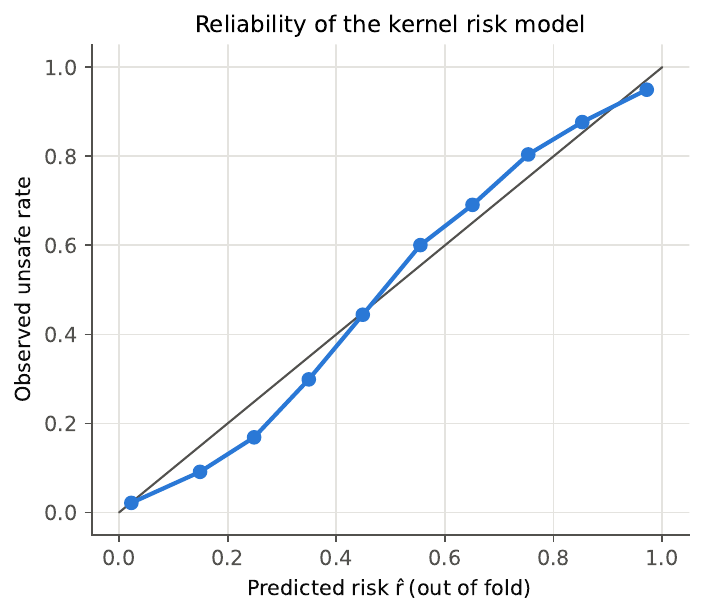}
    \caption{Validation risk function predictions in offline setting for Taskmaster.}
    \label{fig:risk_reliability}
\end{figure}

\begin{figure*}[t]
    \centering
    \includegraphics[width=\textwidth]{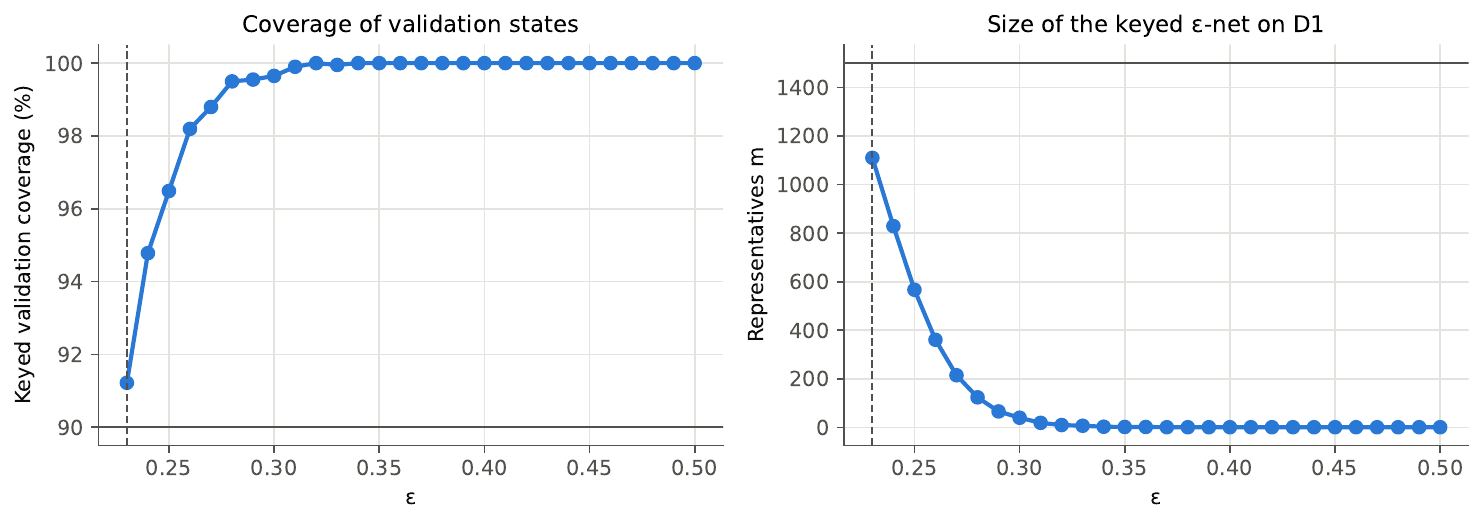}
    \caption{Validation epsilon selection in offline setting for Taskmaster.}
    \label{fig:eps_selection_taskmaster_offline}
\end{figure*}

Tables~\ref{tab:cache-comparison-offline_taskmaster}--\ref{tab:cache-comparison-socrateach-online}
give the losses of Figure~\ref{fig:loss_by_budget} with their CIs.
Figures~\ref{fig:cost_pred_taskmaster_offline}--\ref{fig:eps_selection_taskmaster_offline}
show validation diagnostics of CUCB-SC-MT on Taskmaster-3: the KRR cost
prediction, the reliability of the kernel risk model, and the coverage and
size of the keyed $\epsilon$-net as $\epsilon$ varies.
 
\paragraph{Learning over the stream.}
Tables~\ref{tab:quarter-taskmaster3} and~\ref{tab:quarter-socrateach} split
the online loss into the four quarters of the stream, each as a percentage of
that quarter's no-cache cost. On Taskmaster-3, CLCB-LS-MT's loss falls by
7.9--9.1\,pp from the first to the last quarter at every budget, most of it by
the second quarter (80.3\% to 72.0\% at $k=1000$), while the budgeted
baselines change by at most 2.6\,pp. On SocraTeach the decrease grows with
the budget, from 7.5\,pp at $k=100$ to 13.4\,pp at $k=1000$ and 15.8\,pp
without a budget, since a larger cache keeps more of the responses whose risk
has been learned. CLCB-LS-MT is at most 2.3\,pp above ContextCache in the
first quarter and below it in every later quarter at every budget, while
ContextCache changes by at most 4.1\,pp. Without a budget, vCache also
improves as its thresholds adapt (by 7.4\,pp on Taskmaster-3 and 40.5\,pp on
SocraTeach), but it ends at 84.8\% and 97.9\%, against 72.0\% and 68.5\% for
CLCB-LS-MT; on SocraTeach, unbounded ElastiCache-MT and the joint-state
threshold stay above no caching in every quarter.
 
\begin{figure*}[t]
    \centering
    \includegraphics[width=\textwidth]{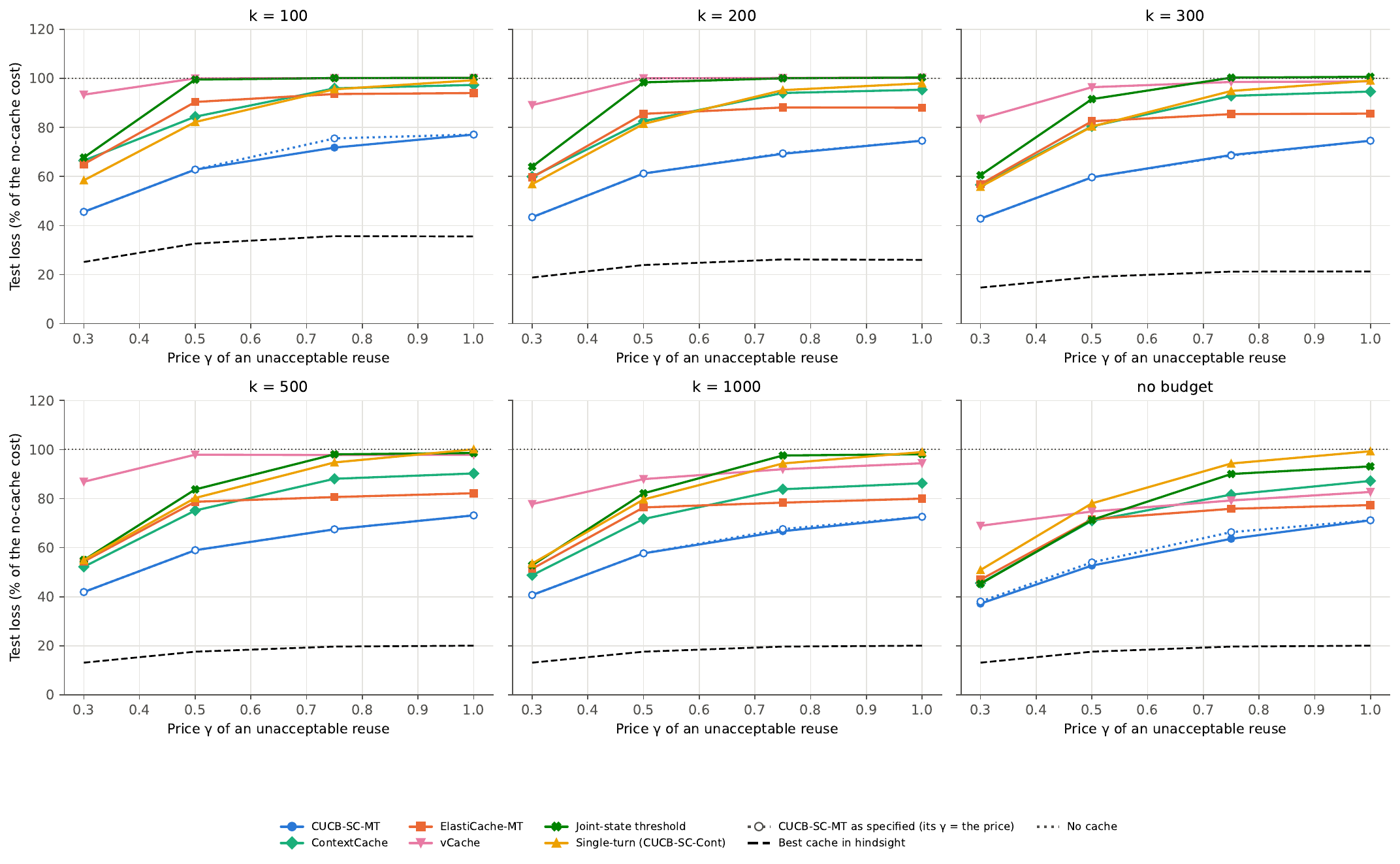}
    \caption{Taskmaster-3 offline: test loss versus the price $\gamma$ of an unacceptable reuse, with every method tuned on validation at each price. Dotted: CUCB-SC-MT as specified ($\hat\gamma=\gamma$); dashed: best $k$-entry cache in hindsight.}
    \label{fig:price_sweep_taskmaster_offline}
\end{figure*}

\begin{figure*}[t]
    \centering
    \includegraphics[width=\textwidth]{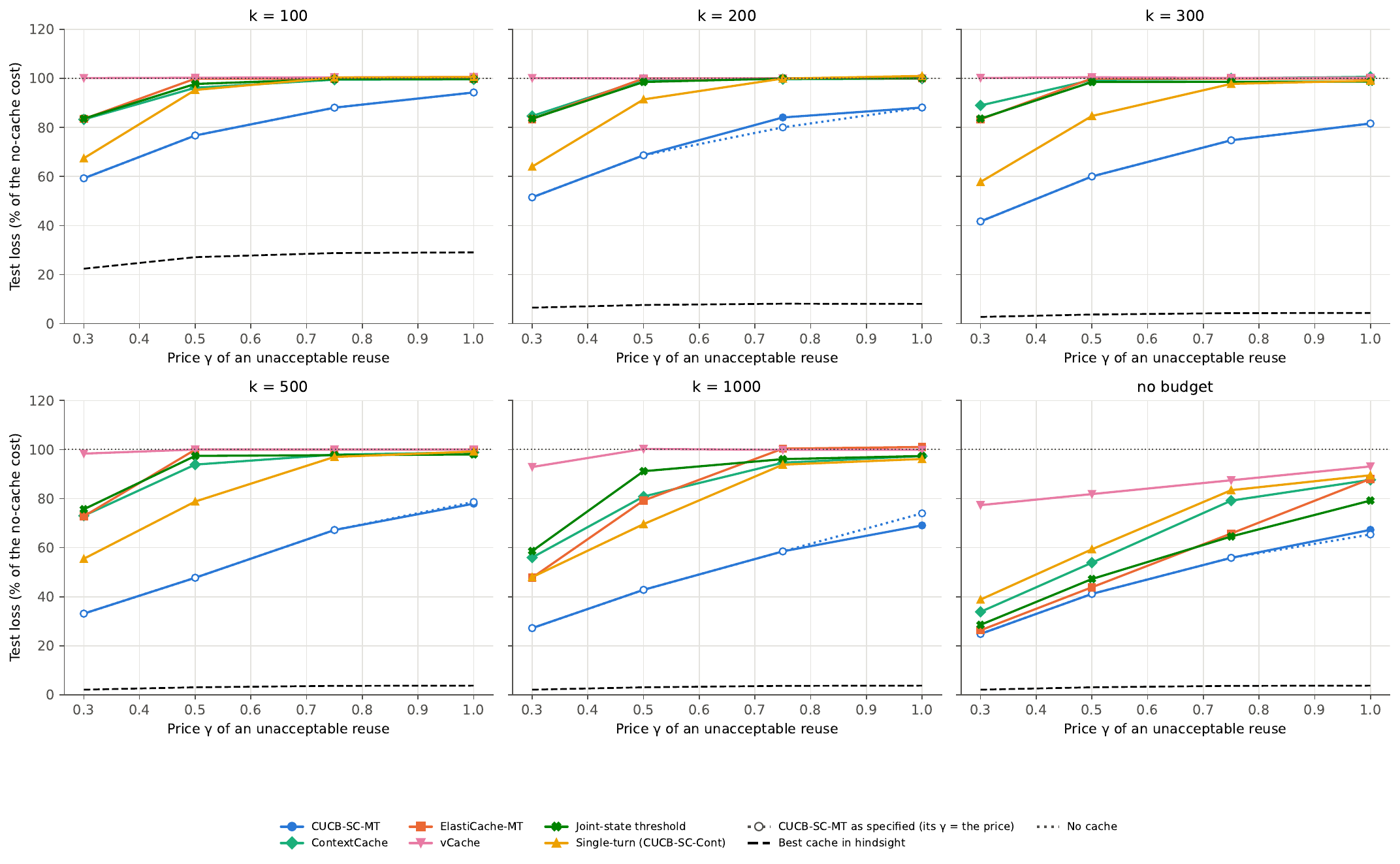}
    \caption{SocraTeach offline: test loss versus the price $\gamma$ of an unacceptable reuse, with every method tuned on validation at each price. Dotted: CUCB-SC-MT as specified ($\hat\gamma=\gamma$); dashed: best $k$-entry cache in hindsight.}
    \label{fig:price_sweep_socrateach_offline}
\end{figure*}

\paragraph{Price of an unacceptable reuse.}
Figures~\ref{fig:price_sweep_taskmaster_offline}
and~\ref{fig:price_sweep_socrateach_offline} repeat the offline comparison for
$\gamma\in\{0.3,0.5,0.75,1\}$, with every method tuned on validation at each
price. A lower price makes an unacceptable reuse cheaper, so every method
reuses more and its loss falls. CUCB-SC-MT has the lowest loss at every price
and budget on both workloads. The budgeted baselines approach no caching as
$\gamma$ grows, most clearly on SocraTeach, while CUCB-SC-MT stays below it.
Without a budget on SocraTeach at $\gamma=0.3$, caching every source state
becomes competitive: ElastiCache-MT and the joint-state threshold come within
a few points of CUCB-SC-MT. CUCB-SC-MT as specified ($\hat\gamma=\gamma$,
dotted) differs from the tuned version at only a few points, by at most
4.9\,pp (SocraTeach, $k=1000$, $\gamma=1$), so tuning $\hat\gamma$ matters
little. The best $k$-entry cache in hindsight (dashed) uses the test labels,
so it is a reference rather than an achievable target.

\begin{table*}[t]
\centering
\small
\caption{Taskmaster-3 online: reuse quality by method and budget. Precision is the judge's acceptance rate of the hits; false reuse and safe reuse are shares of all requests; safe cost avoided is a share of the no-cache cost. Regret is against the best fixed $k$-entry cache in hindsight (judged pool); for CLCB-LS-MT the session buffer is not part of the comparator.}
\label{tab:online-reuse-quality-taskmaster3}
\resizebox{\textwidth}{!}{%
\begin{tabular}{lrrrrrrrrr}
\toprule
Method & Budget & Hit rate & Precision & Safe reuse & False reuse & Safe cost avoided & Same-conv. hits & Loss (\%) & Regret (pp) \\
\midrule
CLCB-LS-MT & k = 100 & 28.1\% & 73.4\% & 20.6\% & 7.5\% & 32.4\% & 22.9\% & 75.0 & +19.6 \\
CLCB-LS-MT & k = 200 & 30.4\% & 73.4\% & 22.3\% & 8.1\% & 33.9\% & 20.5\% & 74.1 & +25.3 \\
CLCB-LS-MT & k = 300 & 32.3\% & 74.2\% & 23.9\% & 8.3\% & 35.0\% & 18.4\% & 73.6 & +28.4 \\
CLCB-LS-MT & k = 500 & 32.7\% & 75.1\% & 24.5\% & 8.1\% & 35.4\% & 17.9\% & 73.0 & +32.8 \\
CLCB-LS-MT & k = 1000 & 32.3\% & 74.6\% & 24.1\% & 8.2\% & 35.0\% & 18.7\% & 73.5 & +40.6 \\
CLCB-LS-MT & no budget & 31.7\% & 74.0\% & 23.4\% & 8.2\% & 34.3\% & 19.4\% & 74.1 & +53.1 \\
CLCB-LS-MT, other conv. only & k = 100 & 25.0\% & 76.9\% & 19.2\% & 5.8\% & 29.3\% & 0.0\% & 76.9 & +21.5 \\
CLCB-LS-MT, other conv. only & k = 200 & 28.2\% & 77.3\% & 21.8\% & 6.4\% & 30.6\% & 0.0\% & 76.6 & +27.7 \\
CLCB-LS-MT, other conv. only & k = 300 & 27.9\% & 78.1\% & 21.8\% & 6.1\% & 30.6\% & 0.0\% & 76.0 & +30.8 \\
CLCB-LS-MT, other conv. only & k = 500 & 28.5\% & 78.3\% & 22.4\% & 6.2\% & 31.4\% & 0.0\% & 75.3 & +35.1 \\
CLCB-LS-MT, other conv. only & k = 1000 & 27.8\% & 78.7\% & 21.9\% & 5.9\% & 30.7\% & 0.0\% & 75.7 & +42.8 \\
CLCB-LS-MT, other conv. only & no budget & 28.1\% & 78.5\% & 22.1\% & 6.1\% & 31.2\% & 0.0\% & 75.5 & +54.5 \\
ContextCache & k = 100 & 27.8\% & 39.9\% & 11.1\% & 16.7\% & 16.7\% & 91.3\% & 109.2 & +53.8 \\
ContextCache & k = 200 & 16.7\% & 50.2\% & 8.4\% & 8.3\% & 11.4\% & 74.4\% & 101.1 & +52.2 \\
ContextCache & k = 300 & 19.2\% & 53.2\% & 10.2\% & 9.0\% & 13.3\% & 65.6\% & 100.2 & +55.1 \\
ContextCache & k = 500 & 22.6\% & 56.3\% & 12.8\% & 9.9\% & 15.8\% & 54.0\% & 99.4 & +59.3 \\
ContextCache & k = 1000 & 25.1\% & 62.0\% & 15.5\% & 9.5\% & 17.9\% & 39.0\% & 97.2 & +64.2 \\
ContextCache & no budget & 29.4\% & 70.0\% & 20.6\% & 8.8\% & 21.2\% & 13.0\% & 94.6 & +73.6 \\
ContextCache, other conv. only & k = 100 & 3.2\% & 74.5\% & 2.4\% & 0.8\% & 2.6\% & 0.0\% & 98.8 & +43.4 \\
ContextCache, other conv. only & k = 200 & 5.3\% & 78.5\% & 4.1\% & 1.1\% & 4.6\% & 0.0\% & 97.4 & +48.5 \\
ContextCache, other conv. only & k = 300 & 7.8\% & 79.6\% & 6.2\% & 1.6\% & 6.7\% & 0.0\% & 95.9 & +50.7 \\
ContextCache, other conv. only & k = 500 & 11.1\% & 79.4\% & 8.8\% & 2.3\% & 9.6\% & 0.0\% & 94.5 & +54.3 \\
ContextCache, other conv. only & k = 1000 & 16.0\% & 80.9\% & 12.9\% & 3.1\% & 13.9\% & 0.0\% & 91.7 & +58.8 \\
ContextCache, other conv. only & no budget & 26.4\% & 76.0\% & 20.0\% & 6.3\% & 20.2\% & 0.0\% & 91.9 & +70.9 \\
ElastiCache-MT & k = 100 & 10.3\% & 60.4\% & 6.2\% & 4.1\% & 9.0\% & 60.9\% & 99.0 & +43.6 \\
ElastiCache-MT & k = 200 & 13.3\% & 68.1\% & 9.0\% & 4.2\% & 12.8\% & 44.9\% & 95.6 & +46.7 \\
ElastiCache-MT & k = 300 & 15.6\% & 72.2\% & 11.3\% & 4.3\% & 15.7\% & 36.8\% & 92.9 & +47.8 \\
ElastiCache-MT & k = 500 & 21.3\% & 74.0\% & 15.8\% & 5.6\% & 20.8\% & 30.6\% & 90.2 & +50.0 \\
ElastiCache-MT & k = 1000 & 20.5\% & 82.5\% & 16.9\% & 3.6\% & 22.3\% & 19.4\% & 85.0 & +52.0 \\
ElastiCache-MT & no budget & 27.0\% & 84.1\% & 22.7\% & 4.3\% & 27.3\% & 13.7\% & 81.9 & +60.9 \\
ElastiCache-MT, other conv. only & k = 100 & 4.0\% & 89.3\% & 3.6\% & 0.4\% & 4.9\% & 0.0\% & 96.1 & +40.7 \\
ElastiCache-MT, other conv. only & k = 200 & 7.3\% & 90.4\% & 6.6\% & 0.7\% & 8.9\% & 0.0\% & 92.6 & +43.8 \\
ElastiCache-MT, other conv. only & k = 300 & 9.6\% & 91.7\% & 8.8\% & 0.8\% & 11.8\% & 0.0\% & 90.0 & +44.8 \\
ElastiCache-MT, other conv. only & k = 500 & 14.6\% & 89.9\% & 13.1\% & 1.5\% & 16.9\% & 0.0\% & 86.4 & +46.2 \\
ElastiCache-MT, other conv. only & k = 1000 & 16.4\% & 92.7\% & 15.2\% & 1.2\% & 19.7\% & 0.0\% & 82.9 & +49.9 \\
ElastiCache-MT, other conv. only & no budget & 23.4\% & 91.5\% & 21.4\% & 2.0\% & 25.2\% & 0.0\% & 79.5 & +58.5 \\
vCache & k = 100 & 0.3\% & 70.3\% & 0.2\% & 0.1\% & 0.2\% & 0.0\% & 100.1 & +44.7 \\
vCache & k = 200 & 0.7\% & 81.2\% & 0.6\% & 0.1\% & 0.7\% & 0.0\% & 99.6 & +50.7 \\
vCache & k = 300 & 1.2\% & 89.2\% & 1.1\% & 0.1\% & 1.5\% & 0.0\% & 98.8 & +53.6 \\
vCache & k = 500 & 6.9\% & 86.5\% & 6.0\% & 0.9\% & 7.7\% & 0.0\% & 93.9 & +53.7 \\
vCache & k = 1000 & 14.8\% & 81.1\% & 12.0\% & 2.8\% & 14.7\% & 0.0\% & 90.4 & +57.5 \\
vCache & no budget & 12.9\% & 89.9\% & 11.6\% & 1.3\% & 15.1\% & 0.0\% & 87.3 & +66.3 \\
Joint-state threshold & k = 100 & 0.7\% & 61.9\% & 0.4\% & 0.3\% & 0.6\% & 80.3\% & 99.8 & +44.4 \\
Joint-state threshold & k = 200 & 2.0\% & 55.2\% & 1.1\% & 0.9\% & 1.1\% & 51.3\% & 100.4 & +51.5 \\
Joint-state threshold & k = 300 & 2.4\% & 61.7\% & 1.5\% & 0.9\% & 1.5\% & 41.1\% & 100.1 & +54.9 \\
Joint-state threshold & k = 500 & 3.5\% & 66.6\% & 2.3\% & 1.2\% & 2.1\% & 29.0\% & 100.1 & +59.9 \\
Joint-state threshold & k = 1000 & 3.4\% & 70.7\% & 2.4\% & 1.0\% & 2.3\% & 25.5\% & 99.6 & +66.7 \\
Joint-state threshold & no budget & 9.2\% & 78.6\% & 7.2\% & 2.0\% & 6.7\% & 8.2\% & 97.6 & +76.6 \\
Single-turn (CUCB-SC-Cont) & k = 100 & 2.2\% & 24.0\% & 0.5\% & 1.7\% & 0.8\% & 0.0\% & 101.9 & +46.5 \\
Single-turn (CUCB-SC-Cont) & k = 200 & 5.5\% & 30.0\% & 1.6\% & 3.8\% & 2.7\% & 0.0\% & 102.8 & +54.0 \\
Single-turn (CUCB-SC-Cont) & k = 300 & 2.6\% & 27.2\% & 0.7\% & 1.9\% & 1.0\% & 0.0\% & 102.1 & +57.0 \\
Single-turn (CUCB-SC-Cont) & k = 500 & 0.1\% & 18.2\% & 0.0\% & 0.1\% & 0.0\% & 0.0\% & 100.2 & +60.0 \\
Single-turn (CUCB-SC-Cont) & k = 1000 & 2.8\% & 29.1\% & 0.8\% & 2.0\% & 1.1\% & 0.0\% & 102.3 & +69.3 \\
Single-turn (CUCB-SC-Cont) & no budget & 3.2\% & 32.8\% & 1.0\% & 2.1\% & 1.2\% & 0.0\% & 102.4 & +81.4 \\
\bottomrule
\end{tabular}}
\end{table*}

\begin{table*}[t]
\centering
\small
\caption{SocraTeach online: reuse quality by method and budget. Precision is the judge's acceptance rate of the hits; false reuse and safe reuse are shares of all requests; safe cost avoided is a share of the no-cache cost. Regret is against the best fixed $k$-entry cache in hindsight (judged pool); for CLCB-LS-MT the session buffer is not part of the comparator.}
\label{tab:online-reuse-quality-socrateach}
\resizebox{\textwidth}{!}{%
\begin{tabular}{lrrrrrrrrr}
\toprule
Method & Budget & Hit rate & Precision & Safe reuse & False reuse & Safe cost avoided & Same-conv. hits & Loss (\%) & Regret (pp) \\
\midrule
CLCB-LS-MT & k = 100 & 35.7\% & 61.1\% & 21.8\% & 13.9\% & 34.3\% & 87.7\% & 82.7 & +5.4 \\
CLCB-LS-MT & k = 200 & 36.4\% & 63.6\% & 23.2\% & 13.3\% & 36.6\% & 77.6\% & 79.9 & +18.6 \\
CLCB-LS-MT & k = 300 & 37.1\% & 64.3\% & 23.8\% & 13.2\% & 37.6\% & 69.3\% & 78.9 & +29.6 \\
CLCB-LS-MT & k = 500 & 38.4\% & 66.2\% & 25.4\% & 13.0\% & 39.6\% & 60.4\% & 77.1 & +43.8 \\
CLCB-LS-MT & k = 1000 & 40.3\% & 67.5\% & 27.2\% & 13.1\% & 41.0\% & 54.4\% & 76.1 & +61.6 \\
CLCB-LS-MT & no budget & 46.9\% & 71.9\% & 33.7\% & 13.2\% & 43.4\% & 44.0\% & 74.4 & +72.1 \\
CLCB-LS-MT, other conv. only & k = 100 & 6.9\% & 55.3\% & 3.8\% & 3.1\% & 6.2\% & 0.0\% & 96.8 & +19.4 \\
CLCB-LS-MT, other conv. only & k = 200 & 12.0\% & 67.5\% & 8.1\% & 3.9\% & 12.5\% & 0.0\% & 92.3 & +31.0 \\
CLCB-LS-MT, other conv. only & k = 300 & 14.8\% & 68.6\% & 10.1\% & 4.6\% & 15.5\% & 0.0\% & 90.7 & +41.4 \\
CLCB-LS-MT, other conv. only & k = 500 & 18.3\% & 71.5\% & 13.1\% & 5.2\% & 19.8\% & 0.0\% & 87.5 & +54.2 \\
CLCB-LS-MT, other conv. only & k = 1000 & 22.0\% & 74.7\% & 16.5\% & 5.6\% & 23.5\% & 0.0\% & 84.7 & +70.2 \\
CLCB-LS-MT, other conv. only & no budget & 27.2\% & 78.8\% & 21.5\% & 5.8\% & 24.9\% & 0.0\% & 83.9 & +81.6 \\
ContextCache & k = 100 & 19.8\% & 71.4\% & 14.1\% & 5.7\% & 22.5\% & 93.5\% & 85.1 & +7.8 \\
ContextCache & k = 200 & 21.2\% & 72.5\% & 15.4\% & 5.8\% & 22.9\% & 86.6\% & 85.1 & +23.7 \\
ContextCache & k = 300 & 7.0\% & 77.3\% & 5.4\% & 1.6\% & 5.2\% & 58.9\% & 97.1 & +47.9 \\
ContextCache & k = 500 & 24.7\% & 74.7\% & 18.5\% & 6.3\% & 23.8\% & 73.3\% & 85.2 & +51.9 \\
ContextCache & k = 1000 & 29.7\% & 76.9\% & 22.8\% & 6.9\% & 25.1\% & 59.5\% & 85.3 & +70.7 \\
ContextCache & no budget & 48.5\% & 78.4\% & 38.0\% & 10.5\% & 35.2\% & 38.4\% & 81.9 & +79.6 \\
ContextCache, other conv. only & k = 100 & 1.2\% & 87.4\% & 1.1\% & 0.2\% & 0.6\% & 0.0\% & 99.7 & +22.3 \\
ContextCache, other conv. only & k = 200 & 2.8\% & 87.1\% & 2.4\% & 0.4\% & 1.5\% & 0.0\% & 99.2 & +37.8 \\
ContextCache, other conv. only & k = 300 & 3.2\% & 91.3\% & 2.9\% & 0.3\% & 1.0\% & 0.0\% & 99.6 & +50.4 \\
ContextCache, other conv. only & k = 500 & 6.7\% & 85.2\% & 5.7\% & 1.0\% & 3.2\% & 0.0\% & 98.8 & +65.6 \\
ContextCache, other conv. only & k = 1000 & 12.5\% & 83.8\% & 10.4\% & 2.0\% & 6.1\% & 0.0\% & 98.0 & +83.5 \\
ContextCache, other conv. only & no budget & 36.6\% & 81.2\% & 29.7\% & 6.9\% & 21.1\% & 0.0\% & 91.8 & +89.5 \\
ElastiCache-MT & k = 100 & 2.7\% & 67.9\% & 1.8\% & 0.9\% & 1.5\% & 66.9\% & 100.3 & +22.9 \\
ElastiCache-MT & k = 200 & 1.8\% & 93.0\% & 1.7\% & 0.1\% & 0.3\% & 6.5\% & 100.0 & +38.7 \\
ElastiCache-MT & k = 300 & 2.7\% & 94.2\% & 2.5\% & 0.2\% & 0.4\% & 4.5\% & 100.0 & +50.8 \\
ElastiCache-MT & k = 500 & 4.2\% & 93.4\% & 3.9\% & 0.3\% & 0.6\% & 2.9\% & 100.2 & +67.0 \\
ElastiCache-MT & k = 1000 & 6.9\% & 95.6\% & 6.6\% & 0.3\% & 0.8\% & 0.0\% & 100.1 & +85.5 \\
ElastiCache-MT & no budget & 91.8\% & 48.2\% & 44.2\% & 47.6\% & 47.6\% & 12.3\% & 154.7 & +152.3 \\
ElastiCache-MT, other conv. only & k = 100 & 0.9\% & 94.8\% & 0.8\% & 0.0\% & 0.1\% & 0.0\% & 100.0 & +22.6 \\
ElastiCache-MT, other conv. only & k = 200 & 1.7\% & 96.2\% & 1.7\% & 0.1\% & 0.2\% & 0.0\% & 99.9 & +38.6 \\
ElastiCache-MT, other conv. only & k = 300 & 2.6\% & 96.4\% & 2.5\% & 0.1\% & 0.3\% & 0.0\% & 99.9 & +50.7 \\
ElastiCache-MT, other conv. only & k = 500 & 4.1\% & 94.8\% & 3.9\% & 0.2\% & 0.5\% & 0.0\% & 100.1 & +66.9 \\
ElastiCache-MT, other conv. only & k = 1000 & 6.9\% & 95.6\% & 6.6\% & 0.3\% & 0.8\% & 0.0\% & 100.1 & +85.5 \\
ElastiCache-MT, other conv. only & no budget & 84.6\% & 49.3\% & 41.7\% & 43.0\% & 44.3\% & 0.0\% & 151.8 & +149.5 \\
vCache & k = 100 & 0.1\% & 25.0\% & 0.0\% & 0.1\% & 0.0\% & 0.0\% & 100.1 & +22.7 \\
vCache & k = 200 & 0.2\% & 39.1\% & 0.1\% & 0.1\% & 0.1\% & 0.0\% & 100.2 & +38.9 \\
vCache & k = 300 & 0.2\% & 40.0\% & 0.1\% & 0.1\% & 0.1\% & 0.0\% & 100.2 & +50.9 \\
vCache & k = 500 & 0.3\% & 54.5\% & 0.2\% & 0.1\% & 0.2\% & 0.0\% & 100.1 & +66.9 \\
vCache & k = 1000 & 0.7\% & 78.4\% & 0.5\% & 0.1\% & 0.7\% & 0.0\% & 99.6 & +85.1 \\
vCache & no budget & 37.1\% & 59.5\% & 22.1\% & 15.0\% & 20.8\% & 0.0\% & 114.3 & +111.9 \\
Joint-state threshold & k = 100 & 1.1\% & 85.8\% & 1.0\% & 0.2\% & 0.4\% & 24.2\% & 99.8 & +22.5 \\
Joint-state threshold & k = 200 & 2.0\% & 90.4\% & 1.8\% & 0.2\% & 0.6\% & 13.2\% & 99.8 & +38.4 \\
Joint-state threshold & k = 300 & 5.6\% & 77.7\% & 4.4\% & 1.3\% & 3.3\% & 35.4\% & 98.7 & +49.5 \\
Joint-state threshold & k = 500 & 7.8\% & 81.3\% & 6.3\% & 1.5\% & 4.2\% & 25.5\% & 98.4 & +65.1 \\
Joint-state threshold & k = 1000 & 12.2\% & 83.5\% & 10.2\% & 2.0\% & 6.0\% & 15.4\% & 97.7 & +83.2 \\
Joint-state threshold & no budget & 61.8\% & 52.8\% & 32.6\% & 29.2\% & 36.2\% & 10.4\% & 129.8 & +127.5 \\
Single-turn (CUCB-SC-Cont) & k = 100 & 0.1\% & 100.0\% & 0.1\% & 0.0\% & 0.1\% & 0.0\% & 99.9 & +22.6 \\
Single-turn (CUCB-SC-Cont) & k = 200 & 0.6\% & 82.1\% & 0.5\% & 0.1\% & 0.8\% & 0.0\% & 99.4 & +38.0 \\
Single-turn (CUCB-SC-Cont) & k = 300 & 0.7\% & 84.6\% & 0.6\% & 0.1\% & 0.9\% & 0.0\% & 99.3 & +50.0 \\
Single-turn (CUCB-SC-Cont) & k = 500 & 1.8\% & 79.3\% & 1.5\% & 0.4\% & 1.4\% & 0.0\% & 99.3 & +66.1 \\
Single-turn (CUCB-SC-Cont) & k = 1000 & 3.4\% & 84.7\% & 2.9\% & 0.5\% & 2.0\% & 0.0\% & 99.0 & +84.5 \\
Single-turn (CUCB-SC-Cont) & no budget & 13.1\% & 89.1\% & 11.7\% & 1.4\% & 5.6\% & 0.0\% & 97.2 & +94.9 \\
\bottomrule
\end{tabular}}
\end{table*}

\begin{table*}[t]
\centering
\small

\caption{Taskmaster-3 online: loss in each quarter of the stream (\% of that quarter's no-cache cost) and its change from the first to the last quarter.}
\label{tab:quarter-taskmaster3}
\resizebox{\textwidth}{!}{%
\begin{tabular}{lrrrrrr}
\toprule
Method & Budget & Q1 & Q2 & Q3 & Q4 & Q4 $-$ Q1 (pp) \\
\midrule
CLCB-LS-MT & k = 100 & 81.6 & 73.5 & 72.5 & 72.5 & $-$9.1 \\
ContextCache & k = 100 & 108.4 & 109.5 & 109.1 & 109.8 & +1.4 \\
ElastiCache-MT & k = 100 & 99.1 & 99.5 & 98.2 & 99.2 & +0.1 \\
vCache & k = 100 & 100.1 & 100.4 & 100.0 & 99.8 & $-$0.2 \\
Joint-state threshold & k = 100 & 99.9 & 99.7 & 99.9 & 99.6 & $-$0.3 \\
Single-turn (CUCB-SC-Cont) & k = 100 & 102.0 & 102.2 & 102.0 & 101.5 & $-$0.5 \\
CLCB-LS-MT & k = 200 & 80.3 & 72.8 & 71.4 & 72.0 & $-$8.2 \\
ContextCache & k = 200 & 100.0 & 102.7 & 100.6 & 100.9 & +0.9 \\
ElastiCache-MT & k = 200 & 95.6 & 95.9 & 94.9 & 95.8 & +0.2 \\
vCache & k = 200 & 99.7 & 99.6 & 99.6 & 99.3 & $-$0.4 \\
Joint-state threshold & k = 200 & 100.3 & 100.5 & 100.4 & 100.2 & $-$0.2 \\
Single-turn (CUCB-SC-Cont) & k = 200 & 102.1 & 104.0 & 103.4 & 101.8 & $-$0.3 \\
CLCB-LS-MT & k = 300 & 79.4 & 72.9 & 71.1 & 71.2 & $-$8.2 \\
ContextCache & k = 300 & 100.3 & 101.2 & 99.3 & 100.0 & $-$0.3 \\
ElastiCache-MT & k = 300 & 93.0 & 92.7 & 91.6 & 94.4 & +1.4 \\
vCache & k = 300 & 100.0 & 99.2 & 96.9 & 99.1 & $-$0.9 \\
Joint-state threshold & k = 300 & 100.0 & 100.2 & 100.3 & 100.1 & +0.1 \\
Single-turn (CUCB-SC-Cont) & k = 300 & 102.3 & 102.3 & 102.2 & 101.7 & $-$0.6 \\
CLCB-LS-MT & k = 500 & 78.7 & 71.9 & 70.7 & 70.8 & $-$7.9 \\
ContextCache & k = 500 & 99.9 & 99.9 & 99.5 & 98.5 & $-$1.3 \\
ElastiCache-MT & k = 500 & 90.6 & 89.9 & 89.0 & 91.2 & +0.7 \\
vCache & k = 500 & 95.7 & 92.8 & 93.6 & 93.7 & $-$2.0 \\
Joint-state threshold & k = 500 & 100.1 & 100.1 & 100.1 & 100.1 & +0.0 \\
Single-turn (CUCB-SC-Cont) & k = 500 & 100.2 & 100.1 & 100.2 & 100.1 & $-$0.1 \\
CLCB-LS-MT & k = 1000 & 80.3 & 72.0 & 70.2 & 71.7 & $-$8.6 \\
ContextCache & k = 1000 & 97.6 & 96.2 & 97.0 & 98.0 & +0.3 \\
ElastiCache-MT & k = 1000 & 86.0 & 84.0 & 84.0 & 85.9 & $-$0.0 \\
vCache & k = 1000 & 93.0 & 89.1 & 89.1 & 90.4 & $-$2.6 \\
Joint-state threshold & k = 1000 & 99.2 & 99.9 & 99.6 & 99.7 & +0.5 \\
Single-turn (CUCB-SC-Cont) & k = 1000 & 102.5 & 102.3 & 102.3 & 102.0 & $-$0.4 \\
CLCB-LS-MT & no budget & 80.8 & 73.1 & 70.5 & 72.0 & $-$8.8 \\
ContextCache & no budget & 95.7 & 94.3 & 93.4 & 95.0 & $-$0.8 \\
ElastiCache-MT & no budget & 84.1 & 80.9 & 80.8 & 81.8 & $-$2.3 \\
vCache & no budget & 92.3 & 86.3 & 85.8 & 84.8 & $-$7.4 \\
Joint-state threshold & no budget & 99.0 & 97.2 & 98.1 & 96.2 & $-$2.8 \\
Single-turn (CUCB-SC-Cont) & no budget & 102.5 & 102.6 & 102.4 & 102.1 & $-$0.4 \\
\bottomrule
\end{tabular}}
\end{table*}

\begin{table*}[t]
\centering
\small
\caption{SocraTeach online: loss in each quarter of the stream (\% of that quarter's no-cache cost) and its change from the first to the last quarter.}
\label{tab:quarter-socrateach}
\resizebox{\textwidth}{!}{%
\begin{tabular}{lrrrrrr}
\toprule
Method & Budget & Q1 & Q2 & Q3 & Q4 & Q4 $-$ Q1 (pp) \\
\midrule
CLCB-LS-MT & k = 100 & 87.4 & 80.3 & 83.4 & 79.9 & $-$7.5 \\
ContextCache & k = 100 & 86.8 & 83.8 & 87.0 & 82.9 & $-$3.9 \\
ElastiCache-MT & k = 100 & 99.7 & 100.7 & 100.5 & 100.2 & +0.4 \\
vCache & k = 100 & 100.2 & 100.1 & 100.1 & 99.9 & $-$0.3 \\
Joint-state threshold & k = 100 & 100.0 & 99.6 & 99.7 & 99.9 & $-$0.1 \\
Single-turn (CUCB-SC-Cont) & k = 100 & 99.8 & 100.0 & 100.0 & 99.9 & +0.1 \\
CLCB-LS-MT & k = 200 & 86.6 & 76.8 & 78.0 & 78.4 & $-$8.2 \\
ContextCache & k = 200 & 86.7 & 83.4 & 87.1 & 83.3 & $-$3.4 \\
ElastiCache-MT & k = 200 & 99.8 & 100.1 & 100.2 & 100.0 & +0.2 \\
vCache & k = 200 & 100.3 & 100.3 & 100.4 & 100.0 & $-$0.3 \\
Joint-state threshold & k = 200 & 100.0 & 99.3 & 99.6 & 100.1 & +0.1 \\
Single-turn (CUCB-SC-Cont) & k = 200 & 99.5 & 99.6 & 99.3 & 99.1 & $-$0.4 \\
CLCB-LS-MT & k = 300 & 85.6 & 76.7 & 77.8 & 75.5 & $-$10.1 \\
ContextCache & k = 300 & 97.3 & 96.5 & 97.5 & 97.2 & $-$0.1 \\
ElastiCache-MT & k = 300 & 99.8 & 100.1 & 100.3 & 99.9 & +0.1 \\
vCache & k = 300 & 100.1 & 100.3 & 100.3 & 99.9 & $-$0.2 \\
Joint-state threshold & k = 300 & 99.4 & 98.3 & 98.6 & 98.8 & $-$0.6 \\
Single-turn (CUCB-SC-Cont) & k = 300 & 99.5 & 99.5 & 99.1 & 99.0 & $-$0.5 \\
CLCB-LS-MT & k = 500 & 85.6 & 75.2 & 74.8 & 72.6 & $-$13.1 \\
ContextCache & k = 500 & 87.0 & 83.7 & 87.1 & 83.0 & $-$4.1 \\
ElastiCache-MT & k = 500 & 99.7 & 100.3 & 100.7 & 100.1 & +0.4 \\
vCache & k = 500 & 100.0 & 100.3 & 100.4 & 99.7 & $-$0.3 \\
Joint-state threshold & k = 500 & 99.5 & 98.0 & 97.8 & 98.0 & $-$1.5 \\
Single-turn (CUCB-SC-Cont) & k = 500 & 99.8 & 99.2 & 99.4 & 99.0 & $-$0.8 \\
CLCB-LS-MT & k = 1000 & 84.6 & 75.8 & 72.7 & 71.2 & $-$13.4 \\
ContextCache & k = 1000 & 86.2 & 83.5 & 87.4 & 84.0 & $-$2.1 \\
ElastiCache-MT & k = 1000 & 99.8 & 100.0 & 100.7 & 99.8 & +0.0 \\
vCache & k = 1000 & 99.9 & 99.3 & 100.0 & 99.3 & $-$0.6 \\
Joint-state threshold & k = 1000 & 98.6 & 96.7 & 97.9 & 97.8 & $-$0.8 \\
Single-turn (CUCB-SC-Cont) & k = 1000 & 99.8 & 98.7 & 98.8 & 98.7 & $-$1.2 \\
CLCB-LS-MT & no budget & 84.3 & 73.3 & 71.6 & 68.5 & $-$15.8 \\
ContextCache & no budget & 82.0 & 80.5 & 82.6 & 82.6 & +0.6 \\
ElastiCache-MT & no budget & 165.8 & 150.6 & 156.8 & 145.8 & $-$20.0 \\
vCache & no budget & 138.3 & 113.4 & 107.4 & 97.9 & $-$40.5 \\
Joint-state threshold & no budget & 128.3 & 124.2 & 135.5 & 131.6 & +3.2 \\
Single-turn (CUCB-SC-Cont) & no budget & 99.6 & 97.5 & 96.1 & 95.6 & $-$4.0 \\
\bottomrule
\end{tabular}}
\end{table*}

\begin{table*}[t]
    \centering
    \renewcommand{\arraystretch}{1.25}
    \setlength{\tabcolsep}{10pt}
    \caption{Taskmaster-3 offline: loss at $\gamma=1$ by cache budget (\% of the no-cache cost; lower is better), with 95\% bootstrap confidence intervals in brackets. These are the values of Figure~\ref{fig:loss_by_budget}.}
    \resizebox{\textwidth}{!}{%
    \begin{tabular}{lcccccc}
        \toprule
        Method
        & $k=100$
        & $k=200$
        & $k=300$
        & $k=500$
        & $k=1000$
        & no budget \\
        \midrule

        CUCB-SC-MT
        & \makecell{77.1 \\ $[72.6,\,81.6]$}
        & \makecell{74.6 \\ $[69.8,\,79.5]$}
        & \makecell{74.6 \\ $[69.4,\,79.6]$}
        & \makecell{73.2 \\ $[68.1,\,78.2]$}
        & \makecell{72.6 \\ $[67.5,\,77.7]$}
        & \makecell{71.2 \\ $[65.7,\,76.7]$} \\
        \midrule

        ElastiCache-MT
        & \makecell{94.0 \\ $[91.9,\,96.0]$}
        & \makecell{88.1 \\ $[84.9,\,91.2]$}
        & \makecell{85.6 \\ $[81.7,\,89.6]$}
        & \makecell{82.2 \\ $[76.8,\,87.3]$}
        & \makecell{80.0 \\ $[74.7,\,85.2]$}
        & \makecell{77.3 \\ $[71.3,\,83.4]$} \\
        \midrule

        ContextCache
        & \makecell{97.3 \\ $[95.3,\,99.2]$}
        & \makecell{95.4 \\ $[93.0,\,97.5]$}
        & \makecell{94.6 \\ $[91.7,\,97.3]$}
        & \makecell{90.3 \\ $[86.4,\,94.0]$}
        & \makecell{86.3 \\ $[81.4,\,90.7]$}
        & \makecell{87.2 \\ $[81.2,\,93.0]$} \\
        \midrule

        vCache
        & \makecell{100.2 \\ $[99.9,\,100.8]$}
        & \makecell{100.3 \\ $[99.7,\,101.0]$}
        & \makecell{98.7 \\ $[97.8,\,99.6]$}
        & \makecell{97.9 \\ $[96.4,\,99.2]$}
        & \makecell{94.4 \\ $[90.9,\,97.8]$}
        & \makecell{82.7 \\ $[78.0,\,87.4]$} \\
        \midrule

        Joint-state threshold
        & \makecell{100.2 \\ $[100.0,\,100.6]$}
        & \makecell{100.3 \\ $[99.7,\,101.0]$}
        & \makecell{100.6 \\ $[99.7,\,101.6]$}
        & \makecell{98.6 \\ $[97.1,\,100.0]$}
        & \makecell{98.1 \\ $[96.0,\,100.0]$}
        & \makecell{93.1 \\ $[90.0,\,96.1]$} \\
        \midrule

        Single-turn
        & \makecell{99.3 \\ $[98.0,\,100.5]$}
        & \makecell{98.0 \\ $[95.8,\,100.4]$}
        & \makecell{99.1 \\ $[97.8,\,100.5]$}
        & \makecell{100.1 \\ $[99.9,\,100.4]$}
        & \makecell{99.0 \\ $[97.6,\,100.4]$}
        & \makecell{99.3 \\ $[97.8,\,100.9]$} \\
        \bottomrule
    \end{tabular}%
    }

    \label{tab:cache-comparison-offline_taskmaster}
\end{table*}

\begin{table*}[t]
    \centering
    \renewcommand{\arraystretch}{1.25}
    \setlength{\tabcolsep}{10pt}
    \caption{Taskmaster-3 online: loss at $\gamma=1$ by cache budget (\% of the no-cache cost; lower is better), with 95\% bootstrap confidence intervals in brackets. These are the values of Figure~\ref{fig:loss_by_budget}.}
    \resizebox{\textwidth}{!}{%
    \begin{tabular}{lcccccc}
        \toprule
        Method
        & $k=100$
        & $k=200$
        & $k=300$
        & $k=500$
        & $k=1000$
        & no budget \\
        \midrule

        CLCB-LS-MT
        & \makecell{75.0 \\ $[73.5,\,76.4]$}
        & \makecell{74.1 \\ $[72.6,\,75.6]$}
        & \makecell{73.6 \\ $[72.2,\,75.1]$}
        & \makecell{73.0 \\ $[71.5,\,74.5]$}
        & \makecell{73.5 \\ $[72.0,\,75.0]$}
        & \makecell{74.1 \\ $[72.6,\,75.5]$} \\
        \midrule

        ContextCache
        & \makecell{109.2 \\ $[107.9,\,110.6]$}
        & \makecell{101.1 \\ $[100.1,\,102.0]$}
        & \makecell{100.2 \\ $[99.1,\,101.4]$}
        & \makecell{99.4 \\ $[98.1,\,100.8]$}
        & \makecell{97.2 \\ $[95.7,\,98.7]$}
        & \makecell{94.6 \\ $[93.0,\,96.2]$} \\
        \midrule

        ElastiCache-MT
        & \makecell{99.0 \\ $[98.0,\,99.9]$}
        & \makecell{95.6 \\ $[94.3,\,96.7]$}
        & \makecell{92.9 \\ $[91.6,\,94.3]$}
        & \makecell{90.2 \\ $[88.6,\,91.7]$}
        & \makecell{85.0 \\ $[83.4,\,86.6]$}
        & \makecell{81.9 \\ $[80.0,\,83.7]$} \\
        \midrule

        vCache
        & \makecell{100.1 \\ $[99.9,\,100.2]$}
        & \makecell{99.6 \\ $[99.4,\,99.8]$}
        & \makecell{98.8 \\ $[98.5,\,99.1]$}
        & \makecell{93.9 \\ $[93.2,\,94.7]$}
        & \makecell{90.4 \\ $[89.2,\,91.6]$}
        & \makecell{87.3 \\ $[86.1,\,88.4]$} \\
        \midrule

        Joint-state threshold
        & \makecell{99.8 \\ $[99.6,\,100.0]$}
        & \makecell{100.4 \\ $[100.0,\,100.7]$}
        & \makecell{100.1 \\ $[99.7,\,100.5]$}
        & \makecell{100.1 \\ $[99.6,\,100.6]$}
        & \makecell{99.6 \\ $[99.1,\,100.1]$}
        & \makecell{97.6 \\ $[96.8,\,98.4]$} \\
        \midrule

        \makecell[l]{Single-turn \\ (CUCB-SC-Cont)}
        & \makecell{101.9 \\ $[101.5,\,102.3]$}
        & \makecell{102.8 \\ $[102.3,\,103.4]$}
        & \makecell{102.1 \\ $[101.7,\,102.6]$}
        & \makecell{100.2 \\ $[100.1,\,100.3]$}
        & \makecell{102.3 \\ $[101.8,\,102.7]$}
        & \makecell{102.4 \\ $[102.0,\,102.9]$} \\
        \bottomrule
    \end{tabular}%
    }

    \label{tab:cache-comparison-taskmaster3-online}
\end{table*}

\begin{table*}[t]
    \centering
    \renewcommand{\arraystretch}{1.25}
    \setlength{\tabcolsep}{10pt}
    \caption{SocraTeach offline: loss at $\gamma=1$ by cache budget (\% of the no-cache cost; lower is better), with 95\% bootstrap confidence intervals in brackets. These are the values of Figure~\ref{fig:loss_by_budget}.}
    \resizebox{\textwidth}{!}{%
    \begin{tabular}{lcccccc}
        \toprule
        Method
        & $k=100$
        & $k=200$
        & $k=300$
        & $k=500$
        & $k=1000$
        & no budget \\
        \midrule

        CUCB-SC-MT
        & \makecell{94.3 \\ $[90.5,\,97.7]$}
        & \makecell{88.1 \\ $[83.5,\,92.8]$}
        & \makecell{81.6 \\ $[76.4,\,86.6]$}
        & \makecell{77.9 \\ $[70.6,\,85.4]$}
        & \makecell{69.1 \\ $[61.4,\,76.8]$}
        & \makecell{67.3 \\ $[59.8,\,74.9]$} \\
        \midrule

        ElastiCache-MT
        & \makecell{99.9 \\ $[99.7,\,100.0]$} 
        & \makecell{100.0 \\ $[99.9,\,100.0]$} 
        & \makecell{100.0 \\ $[99.9,\,100.0]$} 
        & \makecell{100.0 \\ $[99.9,\,100.0]$} 
        & \makecell{101.0 \\ $[99.5,\,103.0]$} 
        & \makecell{88.0 \\ $[78.5,\,97.7]$}
         
         \\
         \midrule

        ContextCache
        & \makecell{99.6 \\ $[98.6,\,100.6]$}
        & \makecell{99.9 \\ $[98.7,\,101.1]$}
        & \makecell{100.6 \\ $[99.6,\,102.0]$}
        & \makecell{98.8 \\ $[96.7,\,101.1]$}
        & \makecell{97.4 \\ $[93.9,\,100.9]$}
        & \makecell{87.7 \\ $[81.4,\,93.4]$}
        \\
        \midrule

        vCache
        & \makecell{100.6 \\ $[100.0,\,101.4]$}
        & \makecell{100.0 \\ $[100.0,\,100.0]$}
        & \makecell{100.3 \\ $[100.0,\,101.0]$}
        & \makecell{100.0 \\ $[100.0,\,100.0]$}
        & \makecell{99.9 \\ $[99.6,\,100.0]$}
        & \makecell{93.1 \\ $[87.7,\,98.3]$}
        \\
        \midrule

        Joint-state threshold
        & \makecell{99.7 \\ $[99.2,\,100.0]$}
        & \makecell{100.0 \\ $[99.3,\,100.9]$}
        & \makecell{98.7 \\ $[97.2,\,100.0]$}
        & \makecell{98.0 \\ $[96.4,\,99.5]$}
        & \makecell{97.4 \\ $[94.8,\,100.0]$}
        & \makecell{79.2 \\ $[70.9,\,86.9]$}
        \\
        \midrule

        Single-turn
        & \makecell{100.6 \\ $[99.9,\,101.5]$}
        & \makecell{101.0 \\ $[99.2,\,102.9]$}
        & \makecell{99.1 \\ $[96.6,\,101.6]$}
        & \makecell{99.2 \\ $[96.2,\,102.0]$}
        & \makecell{96.2 \\ $[92.8,\,99.4]$}

        & \makecell{89.5 \\ $[83.8,\,95.0]$} \\
        \midrule

    \end{tabular}%
    }

    \label{tab:cache-comparison-socrateach-offline}
\end{table*}

\begin{table*}[t]
    \centering
    \renewcommand{\arraystretch}{1.25}
    \setlength{\tabcolsep}{10pt}
    \caption{SocraTeach online: loss at $\gamma=1$ by cache budget (\% of the no-cache cost; lower is better), with 95\% bootstrap confidence intervals in brackets. These are the values of Figure~\ref{fig:loss_by_budget}.}
    \resizebox{\textwidth}{!}{%
    \begin{tabular}{lcccccc}
        \toprule
        Method
        & $k=100$
        & $k=200$
        & $k=300$
        & $k=500$
        & $k=1000$
        & No budget \\
        \midrule

        CLCB-LS-MT
        & \makecell{82.7 \\ $[81.0,\,84.5]$}
        & \makecell{79.9 \\ $[78.2,\,81.8]$}
        & \makecell{78.9 \\ $[77.1,\,80.7]$}
        & \makecell{77.1 \\ $[75.2,\,78.9]$}
        & \makecell{76.1 \\ $[74.1,\,77.9]$}
        & \makecell{74.4 \\ $[72.6,\,76.3]$} \\
        \midrule

        ContextCache
        & \makecell{85.1 \\ $[83.8,\,86.4]$}
        & \makecell{85.1 \\ $[83.8,\,86.4]$}
        & \makecell{97.1 \\ $[96.4,\,97.8]$}
        & \makecell{85.2 \\ $[83.8,\,86.5]$}
        & \makecell{85.3 \\ $[83.9,\,86.7]$}
        & \makecell{81.9 \\ $[80.3,\,83.7]$} \\
        \midrule

        ElastiCache-MT
        & \makecell{100.3 \\ $[99.8,\,100.7]$}
        & \makecell{100.0 \\ $[99.9,\,100.2]$}
        & \makecell{100.0 \\ $[99.8,\,100.2]$}
        & \makecell{100.2 \\ $[99.9,\,100.5]$}
        & \makecell{100.1 \\ $[99.8,\,100.4]$}
        & \makecell{154.7 \\ $[150.7,\,159.0]$} \\
        \midrule

        vCache
        & \makecell{100.1 \\ $[100.0,\,100.2]$}
        & \makecell{100.2 \\ $[100.0,\,100.5]$}
        & \makecell{100.2 \\ $[100.0,\,100.4]$}
        & \makecell{100.1 \\ $[99.9,\,100.3]$}
        & \makecell{99.6 \\ $[99.3,\,99.9]$}
        & \makecell{114.3 \\ $[111.8,\,116.9]$} \\
        \midrule

        Joint-state threshold
        & \makecell{99.8 \\ $[99.6,\,100.0]$}
        & \makecell{99.8 \\ $[99.5,\,100.0]$}
        & \makecell{98.7 \\ $[98.1,\,99.3]$}
        & \makecell{98.4 \\ $[97.7,\,99.0]$}
        & \makecell{97.7 \\ $[96.9,\,98.5]$}
        & \makecell{129.8 \\ $[126.8,\,133.0]$} \\
        \midrule

        \makecell[l]{Single-turn \\ (CUCB-SC-Cont)}
        & \makecell{99.9 \\ $[99.8,\,100.0]$}
        & \makecell{99.4 \\ $[99.1,\,99.6]$}
        & \makecell{99.3 \\ $[99.0,\,99.5]$}
        & \makecell{99.3 \\ $[98.9,\,99.7]$}
        & \makecell{99.0 \\ $[98.6,\,99.5]$}
        & \makecell{97.2 \\ $[96.5,\,98.0]$} \\
        \midrule

    \end{tabular}%
    }

    \label{tab:cache-comparison-socrateach-online}
\end{table*}

\end{document}